\documentclass[sn-mathphys,pdflatex,iicol]{sn-jnl}

\usepackage{cite}
\usepackage{amsmath}
\usepackage{mathtools}
\usepackage{algorithm}
\usepackage{algpseudocode}
\usepackage{subfig}
\usepackage{soul}
\usepackage{tabularx}
\usepackage{graphicx}
\usepackage{textcomp}
\usepackage{booktabs}
\usepackage{colortbl}
\usepackage{caption}
\usepackage{bm}
\usepackage{url}
\usepackage[leftcaption]{sidecap}
\usepackage{ulem}
\definecolor{gray}{rgb}{0.86,0.86,0.86}

\jyear{2022}%

\theoremstyle{thmstyleone}%
\theoremstyle{thmstyletwo}%

\theoremstyle{thmstylethree}%

\begin{document}

\title[A Frequency-based Parent Selection]{A Frequency-based Parent Selection for Reducing the Effect of Evaluation Time Bias in Asynchronous Parallel Multi-objective Evolutionary Algorithms}


\author*[1]{\fnm{Tomohiro} \sur{Harada}}\email{harada@tmu.ac.jp}

\affil*[1]{\orgdiv{Faculty of System Design}, \orgname{Tokyo Metropolitan University}, \orgaddress{\street{2-503, 6-6 Asahigaoka}, \city{Hino}, \postcode{1910065}, \state{Tokyo}, \country{Japan}}}


\abstract{Parallel evolutionary algorithms (PEAs) have been studied for reducing the execution time of evolutionary algorithms by utilizing parallel computing.
An asynchronous PEA (APEA) is a scheme of PEAs that increases computational efficiency by generating a new solution immediately after a solution evaluation completes without the idling time of computing nodes. However, because APEA gives more search opportunities to solutions with shorter evaluation times, the evaluation time bias of solutions negatively affects the search performance. To overcome this drawback, this paper proposes a new parent selection method to reduce the effect of evaluation time bias in APEAs. The proposed method considers the search frequency of solutions and selects the parent solutions so that the search progress in the population is uniform regardless of the evaluation time bias. This paper conducts experiments on multi-objective optimization problems that simulate the evaluation time bias. The experiments use NSGA-III, a well-known multi-objective evolutionary algorithm, and compare the proposed method with the conventional synchronous/asynchronous parallelization. The experimental results reveal that the proposed method can reduce the effect of the evaluation time bias while reducing the computing time of the parallel NSGA-III.
}




\maketitle

\section{Introduction}
Evolutionary algorithms (EAs) have been applied to a wide range of real-world optimization problems owing to their high search capability without any problem-specific knowledge.
When applying EAs to real-world applications, solution evaluations may take much computational time, such as due to physical simulation or complex numerical calculations. 

Parallel EAs (PEAs)~\citep{Alba2013,HaradaPGA2020,Raghul2022} are a promising technique to speed up the optimization process for computationally expensive problems.
A master-worker parallelization~\citep{Durillo2008} is one of the typical approaches of PEAs, where a single master computing node executes the main procedure of EA, e.g., initialization, parent selection, genetic operations, and survival selection. In contrast, many worker nodes evaluate each solution in parallel.

Master-worker PEAs (MW-PEAs) can be mainly classified into two approaches, a \textit{synchronous} PEA (SPEA) and an \textit{asynchronous} PEA (APEA)~\citep{Depolli2013}.
SPEA generates a population in the next generation after evaluating all solutions.
On the other hand, in APEA, a new solution to be evaluated is generated immediately after completing one solution evaluation.
Since SPEA needs to wait for a solution with the longest evaluation time before generating the next population, it increases the idling time of worker nodes and decreases the computational efficiency.
On the other hand, APEA can overcome this drawback because it can continuously evolve solutions without the idling time of worker nodes.
However, since APEA generates a new solution whenever a solution evaluation completes, it could lead to local optima with a short evaluation time~\citep{SAEA_GECCO,SAEA_FOGA}.

This paper proposes a new parent selection method to reduce the effect of evaluation time bias in APEA.
Concretely, the proposed method considers the search frequency of each search region and selects parents so that the search frequency of all solutions becomes uniform. 
The proposed method introduces a new parameter that stores the search frequency of solutions and selects parents from solutions with fewer search frequencies.

This paper is an extended version of the author's work~\citep{HaradaSSCI2020}.
This paper improves the behavior analysis of the proposed method in more detail, mainly:
the previous work analyzed the behavior of the proposed method using only one multi-objective benchmark problem (DTLZ1). 
On the other hand, to achieve a deeper analysis, this study utilizes multimodal multi-objective test functions (MMFs)~\citep{MMF2018} and designs benchmarks with two Pareto-optimal solution sets where each Pareto-optimal solution has a different evaluation time. Comparing the proposed method with synchronous/asynchronous parallel MOEA shows that the proposed method can equally obtain Pareto-optimal solution sets with different evaluation times (see Section~\ref{sec:test_problem}).
In addition, this paper provides further analysis of a proposed method parameter, mainly $r_s$ in the proposed method (see Section~\ref{sec:algorithm}).

The rest of this paper is organized as follows.
The following section briefly introduces APEAs and mentions their problems in the evaluation time bias.
Section~\ref{sec:proposed} proposes the parent selection strategy and shows its concrete example on NSGA-III.
Section~\ref{sec:test_problem} defines the test problems used in this work, and  Section~\ref{sec:setting} describes the experimental settings.
Then, the parameter setting of the proposed method is discussed in Section~\ref{sec:exp2}, and Section~\ref{sec:result} compares the proposed method with SPEA and APEA.
Finally, Section~\ref{sec:conclusion} concludes this paper and addresses future works.

\section{Background}\label{sec:APEA}
Parallel evolutionary algorithms (PEAs)~\citep{Alba2013,HaradaPGA2020,Raghul2022} have been studied to reduce the computing time of EA methods by executing a single EA run on multiple computing nodes and have been applied to several real-world applications, such as education~\citep{NGUYEN2021104439}, data mining~\citep{Soufan2015}, nanoscience~\citep{Shayeghi2015}, and routing~\citep{abbasi2020efficient}.

A master-worker model (known as a global model) is a straightforward approach to implementing PEAs~\citep{Durillo2008,Luna2016}, and is widely used in many recent works~\citep{P2022101536,Chitty2021}.
On a master-worker PEA (MW-PEA), a master computing node executes the main procedure of EAs, such as initialization, selection, genetic operators, and replacement.
On the other hand, many worker nodes evaluate newly generated solutions in parallel and return their results to the master node.

MW-PEAs can be classified into synchronous PEAs (SPEAs) and asynchronous PEAs (APEAs).
SPEA waits for all evaluations of solutions executed by worker nodes and generates a new population using all newly evaluated solutions.
Since SPEA needs to wait for the longest evaluation for each generation, computational efficiency decreases if the evaluation times differ.
On the other hand, APEA continuously generates a new solution without waiting for evaluations of other solutions.
This enables the efficient use of the computing resource even if the evaluation times of solutions differ.
Since previous research has demonstrated the effectiveness of APEAs, for example, on  continuous optimizations~\citep{ade2013}, genetic programmings~\citep{Harada2014}, and multi-objective optimizations~\citep{SMSEMOA2016}, this paper focuses on APEAs.

Although APEA is a practical approach of MW-PEAs, the previous studies demonstrated that APEA is \textit{biased} toward the search region having a short evaluation time if the evaluation time differs depending on the search region~\citep{SAEA_FOGA,SAEA_GECCO}.
This happens because APEA gives many opportunities to search for solutions with a short evaluation time.
SPEA, on the other hand, is an option to avoid the effect of the evaluation time bias because it is not affected by the evaluation time bias.
However, SPEA still wastes waiting time because the evaluation times of solutions differ.

From the above, it can be seen that SPEA has poor computational efficiency regardless of the evaluation time bias. On the other hand, APEA is computationally efficient but is affected by the evaluation time bias. Therefore, this research proposes a method to reduce the effect of the evaluation time bias in APEA while maintaining its computational efficiency.

\section{Proposed method}\label{sec:proposed}
This section proposes a new parent selection method for reducing the effect of evaluation time bias in APEAs.
The following subsection first explains the basic concept of the proposed method, and then Section~\ref{sec:algorithm} introduces the proposed parent selection.
Finally, Section~\ref{sec:example} shows an example of applying the proposed method to the asynchronous parallel NSGA-III.

\subsection{Basic concept}
APEAs are negatively affected by the evaluation time bias because the search frequency for solutions with a short evaluation time increases. 
On the other hand, SPEAs are not affected by the evaluation time bias because the search frequency is almost the same in all search regions regardless of the evaluation time bias.
This fact suggests that adjusting the search progress of all solutions to be uniform can effectively reduce the effect of evaluation time bias in APEAs.

From this viewpoint, this paper proposes a new parent selection method that introduces a new parameter, a search frequency parameter, that stores how many offspring are generated from each solution.
The proposed selection method selects parents to preserve the uniformity of the search frequency of solutions according to the additional frequency parameter.
This contributes to preventing excessive parent selection of solutions in the regions with a short evaluation time.

\begin{figure}[tb]
\centering
\subfloat[Conventional APEAs]{\includegraphics[scale=0.2]{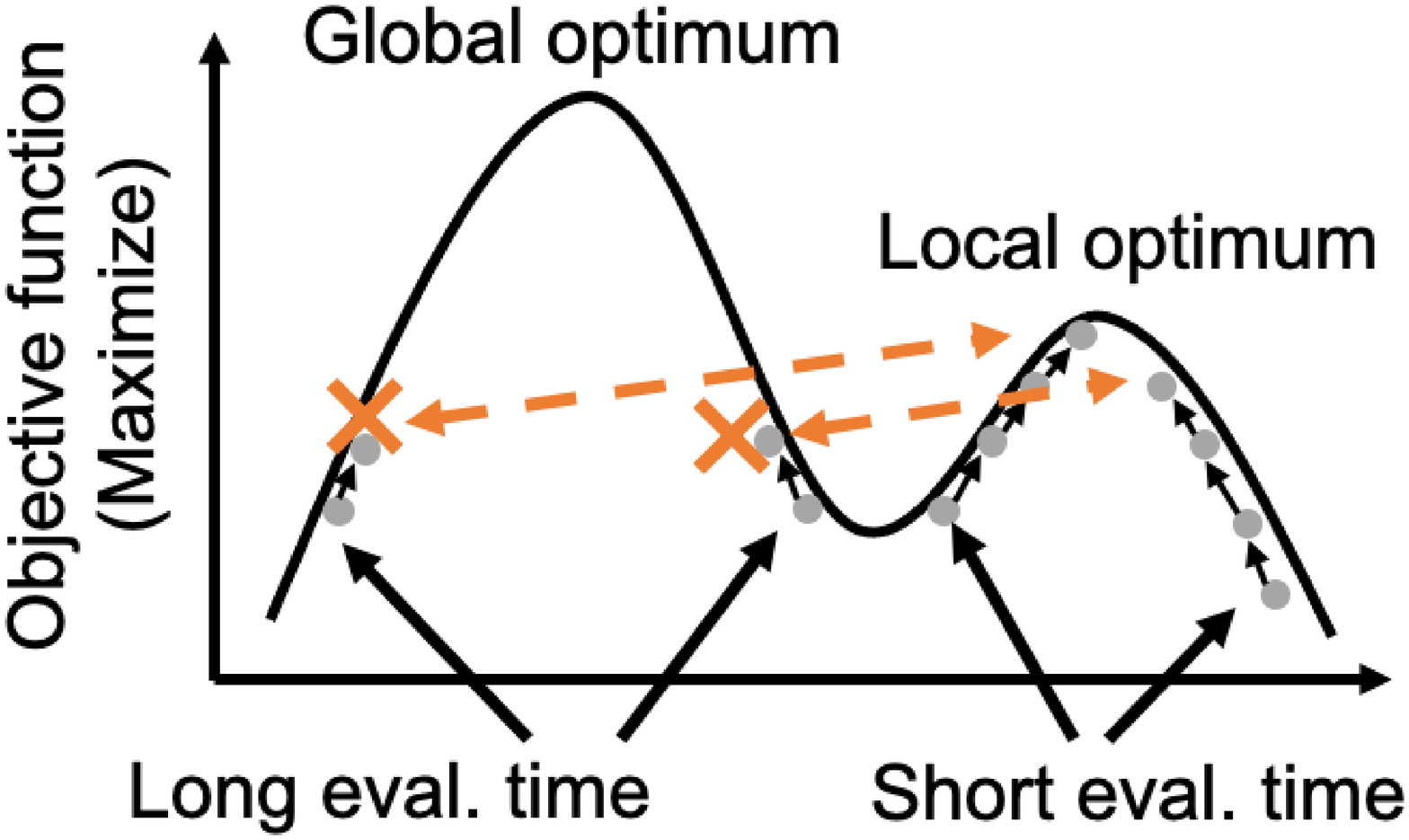}\label{fig:concept_apea}}\\
\subfloat[The proposed parent selection]{\includegraphics[scale=0.2]{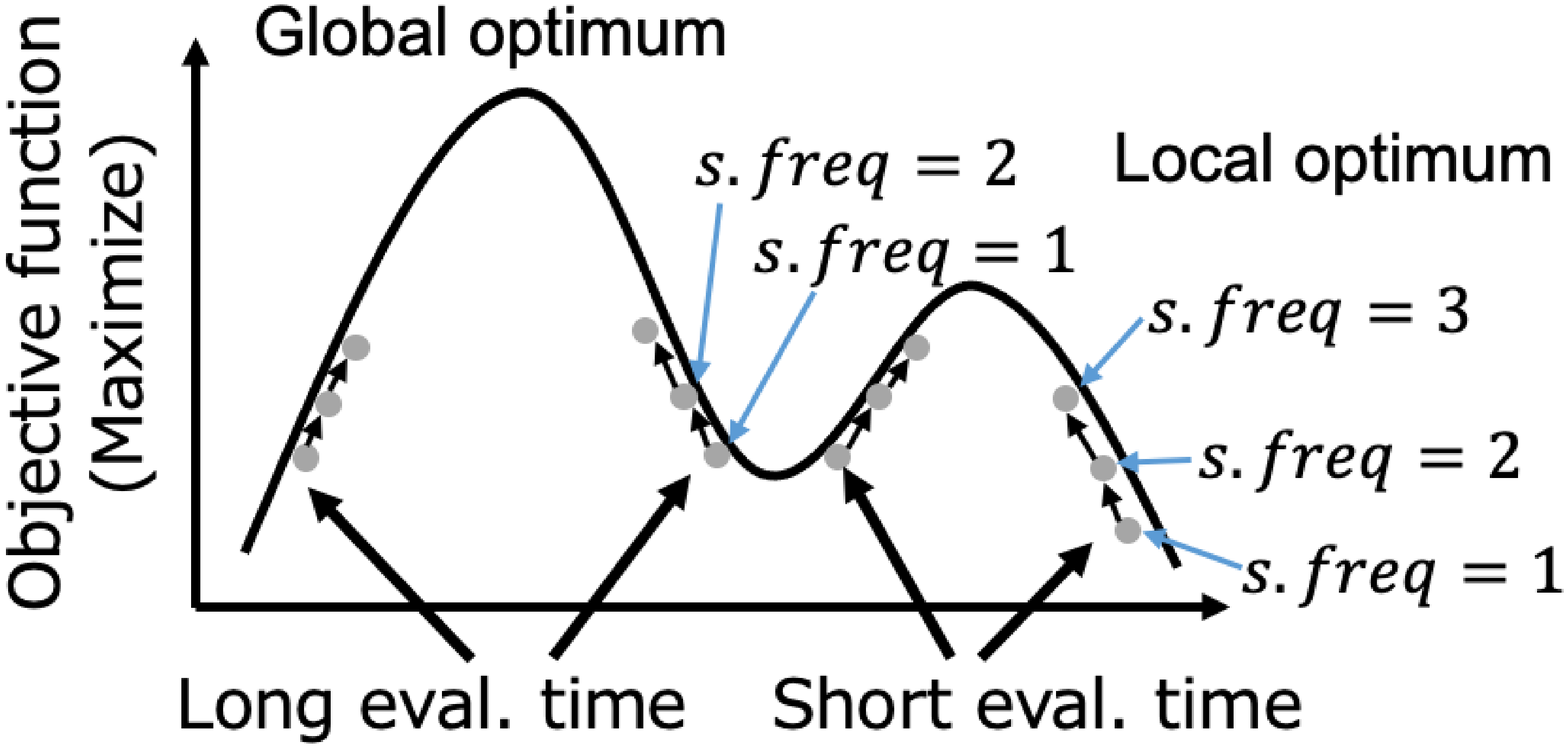}\label{fig:concept_proposed}}
\caption{Illustrations of the conventional APEA and the proposed method}
\label{fig:concept}
\end{figure}
Fig.~\ref{fig:concept} illustrates the conventional and proposed APEAs on a one-dimensional maximization problem with the evaluation time bias.
This example has a global optimum with a long evaluation time and a local optimum with a short evaluation time.
A conventional APEA depicted in Fig.~\ref{fig:concept_apea} increases the search frequency of solutions with a short evaluation time (the right area). This eliminates solutions close to the global optimum (a long evaluation time) by comparing them with the more frequently searched solutions near the local optimum.
To overcome this problem, the proposed method depicted in Fig.~\ref{fig:concept_proposed} stores the search frequency of solutions ($s.freq$) and attempts to preserve the uniformity of these frequencies.
This may avoid eliminating solutions with a long evaluation time due to its slow search progress.

\subsection{Algorithm}\label{sec:algorithm}
The proposed method introduces a new parameter to store the search frequency of each solution.
Let the search frequency of a solution $s$ be $s.freq$.
First, the frequency parameter $s.freq$ for all solutions is initialized to 1.
After that, when a solution is selected as a parent, the frequency parameter is incremented as $s.freq\leftarrow s.freq + 1$.
In addition, generated offspring $s_{new}$ inherits the frequency parameter as the mean of its parents.
In particular, when an offspring solution $s_{new}$ is generated from two parents, $p_1$ and $p_2$, the search frequency parameter of $s_{new}$ is calculated as $s_{new}.freq\leftarrow (p_1.freq+p_2.freq)/2$.
This allows us to store the search progress of each search region as additional information for each solution.

Unlike the conventional parent selection, the proposed method selects only solutions with fewer search frequencies.
In particular, when the parent selection, the proposed method preliminary selects a candidate pool from the current population according to the search frequency parameter.
This can reduce the search opportunities for solutions with a short evaluation time and lead solutions to be uniformly selected as the parents.

\begin{algorithm}[!tb]
\def\NoNumber#1{{\def\alglinenumber##1{}\State #1}\addtocounter{ALG@line}{-1}}
\caption{A pseudo-code of a simple APEA with the proposed method. The \uline{underlined} texts are specific to the proposed method.}
\label{alg:apea_with_proposed}
\begin{algorithmic}[1]
\State Generate $S$ random solutions
\Statex\Comment{$S$ is the number of worker nodes}
\State Send solutions to worker nodes
\State $P_0\leftarrow\emptyset$
\While{$\lvert P_0 \rvert<N$}\Comment{$N$ is the population size}
    \State $s\leftarrow$wait for a solution from worker nodes
    \State $P_0\leftarrow P_0\cup \{s\}$
    \State Generate a random solution $s$
    \State \uline{$s.freq\leftarrow 1$}
    \State Send $s$ to an idling worker node
\EndWhile
\State $t\leftarrow 0$
\While{Terminal conditions}
    \State $s_c\leftarrow$wait for a solution from worker nodes
    \State $P_{t+1}\leftarrow$ select $N$ solutions from $P_{t}\cup \{s_c\}$
    \Statex\Comment{Any replacement strategy}
    \State \uline{Sort $P_{t+1}$ in ascending order according to $s.freq\:(s\in P_{t+1})$}
    \State \uline{$P_{t+1}^\prime\leftarrow$ the top $r_s \lvert P_{t+1}\rvert$ solutions in $P_{t+1}$}
    \State Select two parents, $p_1$ and $p_2$, from $P_{t+1}^\prime$
    \State $s_{new}\leftarrow$a new solution generated from $p_1$ and $p_2$
    \State \uline{$p_i.freq\leftarrow p_i.freq+1\:(i=\{1, 2\})$}
    \State \uline{$s_{new}.freq\leftarrow (p_1.freq+p_2.freq)/2$}
    \State Send $s_{new}$ to the idling worker node
    \State $t\leftarrow t+1$
\EndWhile
\end{algorithmic}
\end{algorithm}
Algorithm~\ref{alg:apea_with_proposed} shows a pseudo-code of an APEA with the proposed parent selection.
Additional procedures from a simple APEA are underlined.
The search frequency parameters of the initial solutions are set to 1 (Step 8).
The population is sorted in ascending order of the frequency parameters when selecting parent solutions (Step 15).
Then, the top $r_s \lvert P_{t+1} \rvert$ solutions in $P_{t+1}$ are extracted as a parent candidate pool $P_{t+1}^\prime$ in Step 16.
Here, $r_s\:(0\le r_s\le 1)$ is a selection ratio parameter that determines how the uniformity of search frequency is prioritized.
Since this may affect the search capability of the proposed method, its effect will be discussed in Section~\ref{sec:exp2}.
Then, parents are selected from $P_{t+1}^\prime$ according to the algorithm-specific selection method (e.g., tournament selection, roulette-wheel selection).
After generating an offspring, the frequency parameters of the parents are incremented by one (Step 19), and a newly generated solution inherits the frequencies of the parents (Step 20).

The difference between the conventional APEA and the proposed method is that: the conventional one selects parents from the entire population regardless of the search frequency. This induces that solutions having short evaluation times frequently get opportunities to be selected as parents, and the search direction is biased.
On the other hand, the proposed method considers the search frequency of solutions and selects parents from less selected solutions for the offspring generation.
This mechanism can allow selection as parents for all solutions and prevent the asynchronous evolution from being affected by the evaluation time bias.

\subsection{An example of the proposed method with NSGA-III}\label{sec:example}
NSGA-III~\citep{NSGAIII} is one of the most well-known and successful MOEA methods combining dominance and decomposition strategies. This section shows an example of applying the proposed method to the asynchronous parallel NSGA-III, which will be used in experiments in Sections~\ref{sec:exp2} and \ref{sec:result}.
Please see the detailed algorithm in \citep{NSGAIII}.

\begin{algorithm}[tb]
\caption{An algorithm of the asynchronous parallel NSGA-III with the proposed frequency-based parent selection}
\label{alg:proposed}
\begin{algorithmic}[1]
\State $t\leftarrow 0$.
\State $P_0\leftarrow \texttt{Initialization}()$.
\State $s.freq=1 \:(s\in P_0)$
\State Send all solutions to worker nodes.
\State Wait for evaluations of all solutions.
\While{Termination conditions}
\State $P_t^\prime\leftarrow$ Sort $P_t$ in ascending order of $s.freq\: (s\in P_t)$
\State $P_t^{\prime\prime}\leftarrow$ The first $r_s\lvert P_t\rvert$ solutions of $P^\prime$
\State $p_1, p_2\leftarrow \texttt{RandomSelection}(P_t^{\prime\prime})$
\State $s_{new}\leftarrow \texttt{GeneticOperators}(p_1, p_2)$ 
\State $p_{i}.freq\leftarrow p_{i}.freq + 1\:(i={1, 2})$
\State $s_{new}.freq\leftarrow (p_1.freq + p_2.freq)/2$
\State Send $s_{new}$ to an idling worker node.
\State $s\leftarrow$ wait for the next evaluation.
\State $R_t\leftarrow P_t\cup \{s\}$.
\State $P_{t+1}\leftarrow \texttt{Selection}(R_t)$.
\State $t\leftarrow t+1$.
\EndWhile
\end{algorithmic}
\end{algorithm}
Algorithm~\ref{alg:proposed} describes the brief flow of the asynchronous parallel NSGA-III assisted by the frequency-based selection (FS-NSGA-III).
The master node initializes the population (Step 2) and assigns the search frequency parameter $s.freq=1$ for all solutions in the initial population $P_0$ (Step 3).
After the initialization, the master node sends all solutions to worker nodes (Step 4), and the main procedure is repeated until satisfying the termination condition.
When generating offspring, the proposed method sorts the population in ascending order of the search frequency parameter (Step 7) and selects the top $r_s\lvert P_t\rvert$ solutions (candidate pool $P_t^{\prime\prime}$) from the current population $P_t^\prime$ (Step 8).
In contrast with NSGA-III randomly selecting two parent solutions from the entire population, FS-NSGA-III selects parents from the candidate pool $P_t^{\prime\prime}$ (Step 9).
After generating offspring, the proposed method increments the frequency parameters of the parents as $p_i.freq\leftarrow p_i.freq+1$ (Step 11) and inherits the frequency parameter of the generated offspring as the mean of its parents as $s_{new}.freq\leftarrow(p_1.freq+p_2.freq)/2$ (Step 12).
Then, the master node sends the generated offspring to an idling worker node (Step 13) and waits for the subsequent evaluation of a solution (Step 14).
When receiving an evaluation, NSGA-III selects the next population from the current population and a newly evaluated solution (Step 16).
NSGA-III uses the front ranking and the niche-preservation operation based on the reference point in the selection procedure.

\section{Test problems with evaluation time bias}\label{sec:test_problem}
\begin{figure}[tb]
\centering
\begin{tabular}{cc}
\begin{minipage}[b]{0.46\columnwidth}
\centering
\subfloat[MMF2]{
\includegraphics[width=\linewidth]{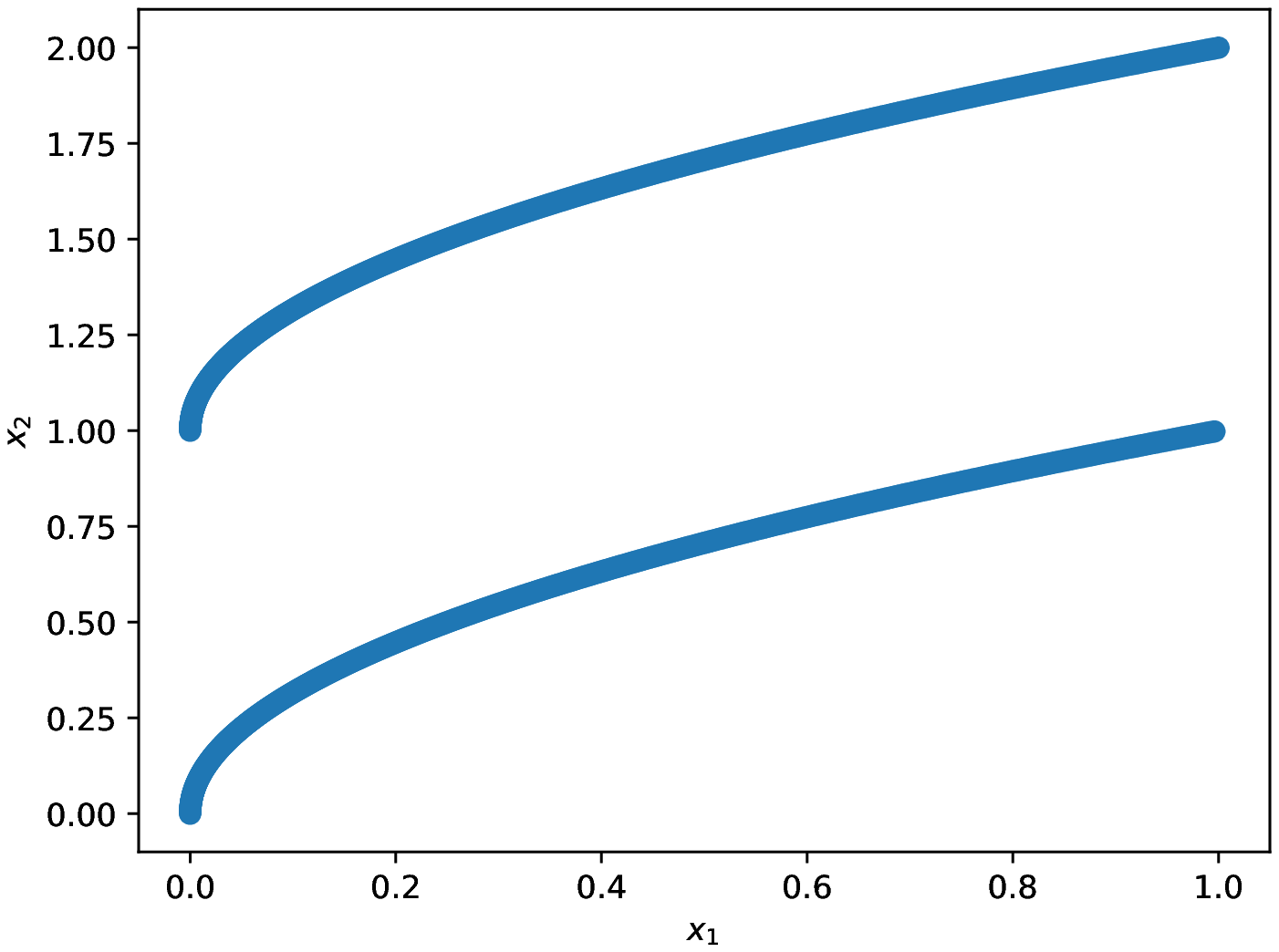}
}
\end{minipage}&
\begin{minipage}[b]{0.46\columnwidth}
\centering
\subfloat[MMF3]{
\includegraphics[width=\linewidth]{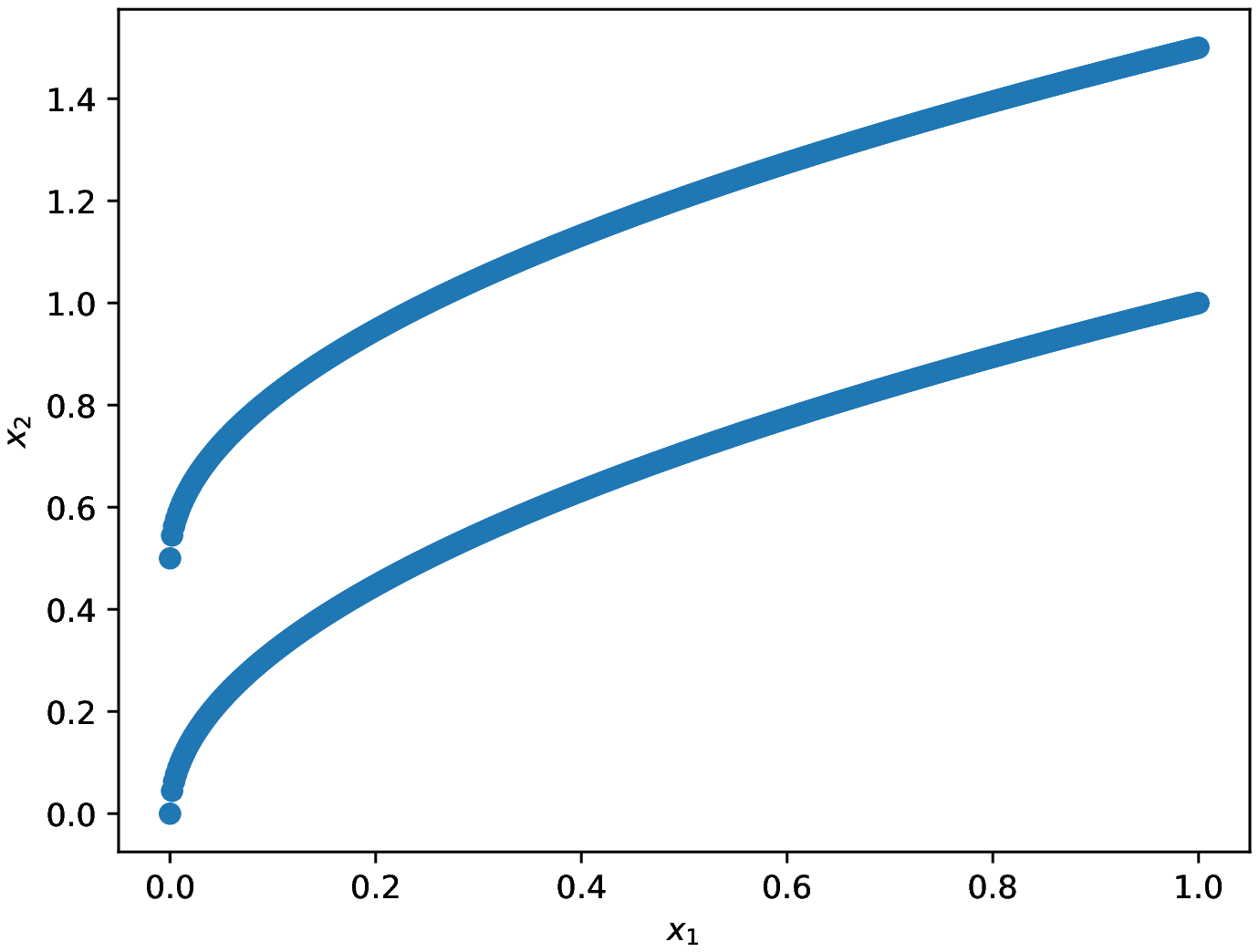}
}
\end{minipage}\\
\begin{minipage}[b]{0.46\columnwidth}
\centering
\subfloat[MMF4]{
\includegraphics[width=\linewidth]{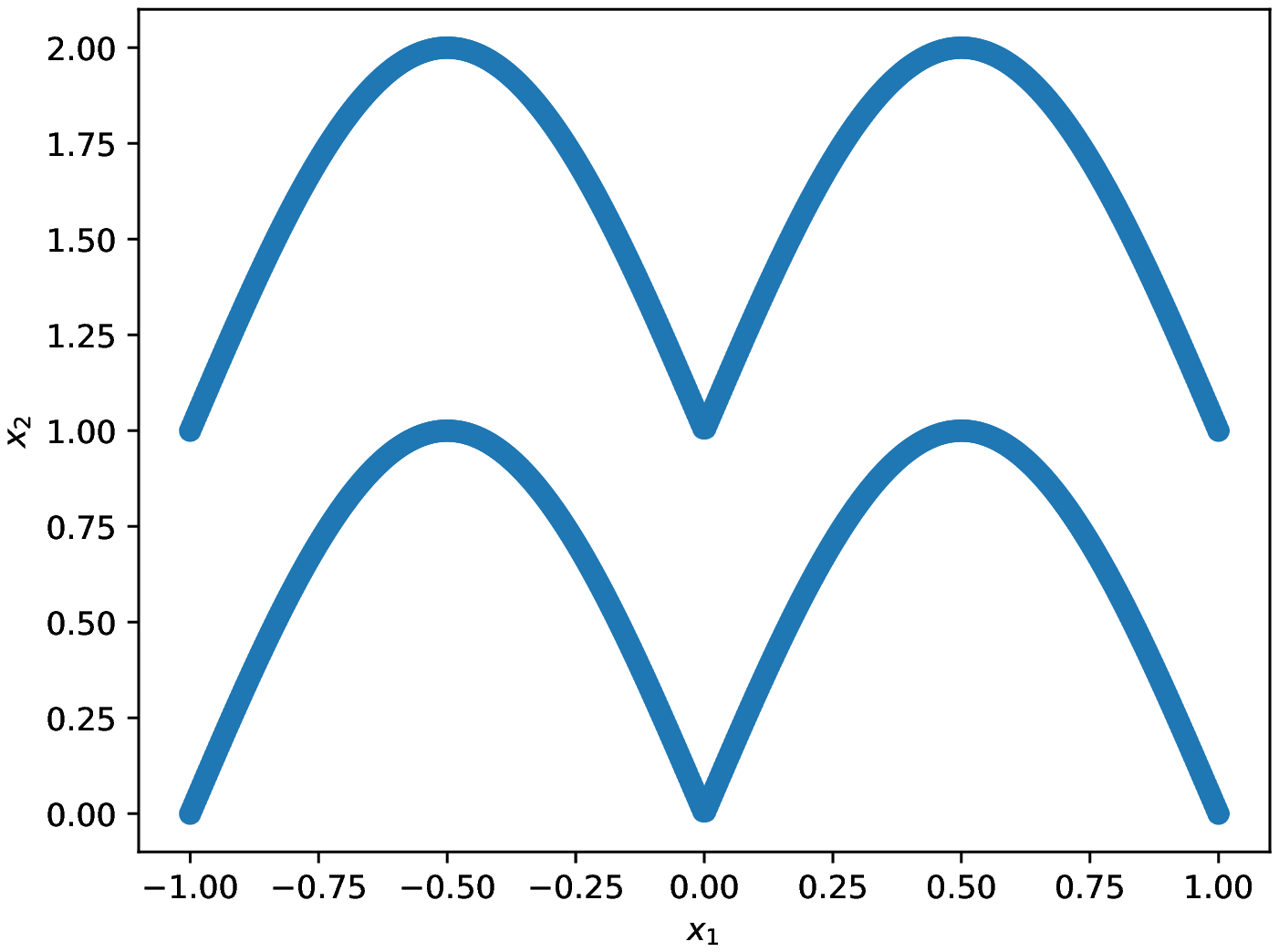}
}
\end{minipage}&
\begin{minipage}[b]{0.46\columnwidth}
\centering
\subfloat[MMF5]{
\includegraphics[width=\linewidth]{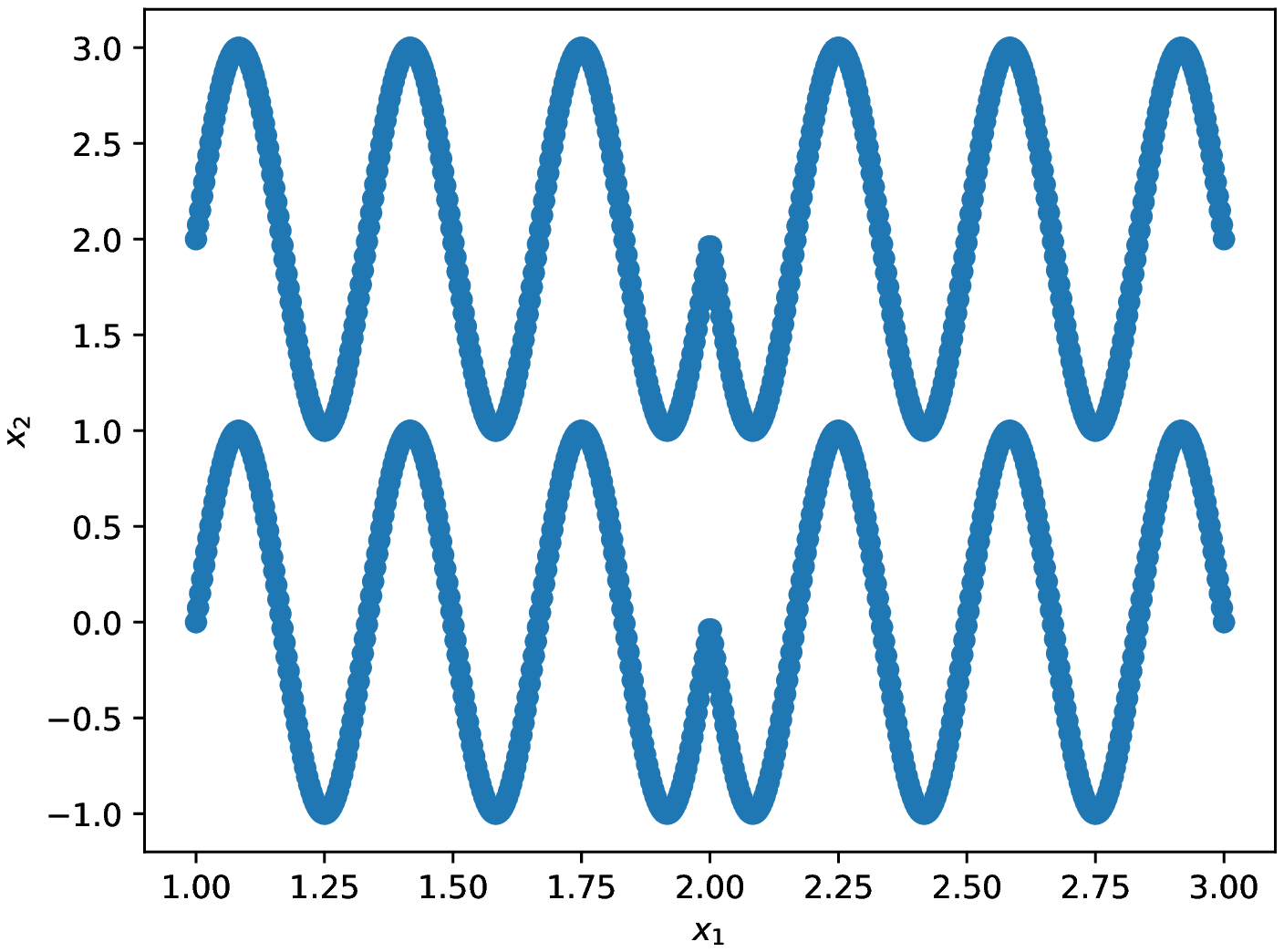}
}
\end{minipage}\\
\begin{minipage}[b]{0.46\columnwidth}
\centering
\subfloat[MMF6]{
\includegraphics[width=\linewidth]{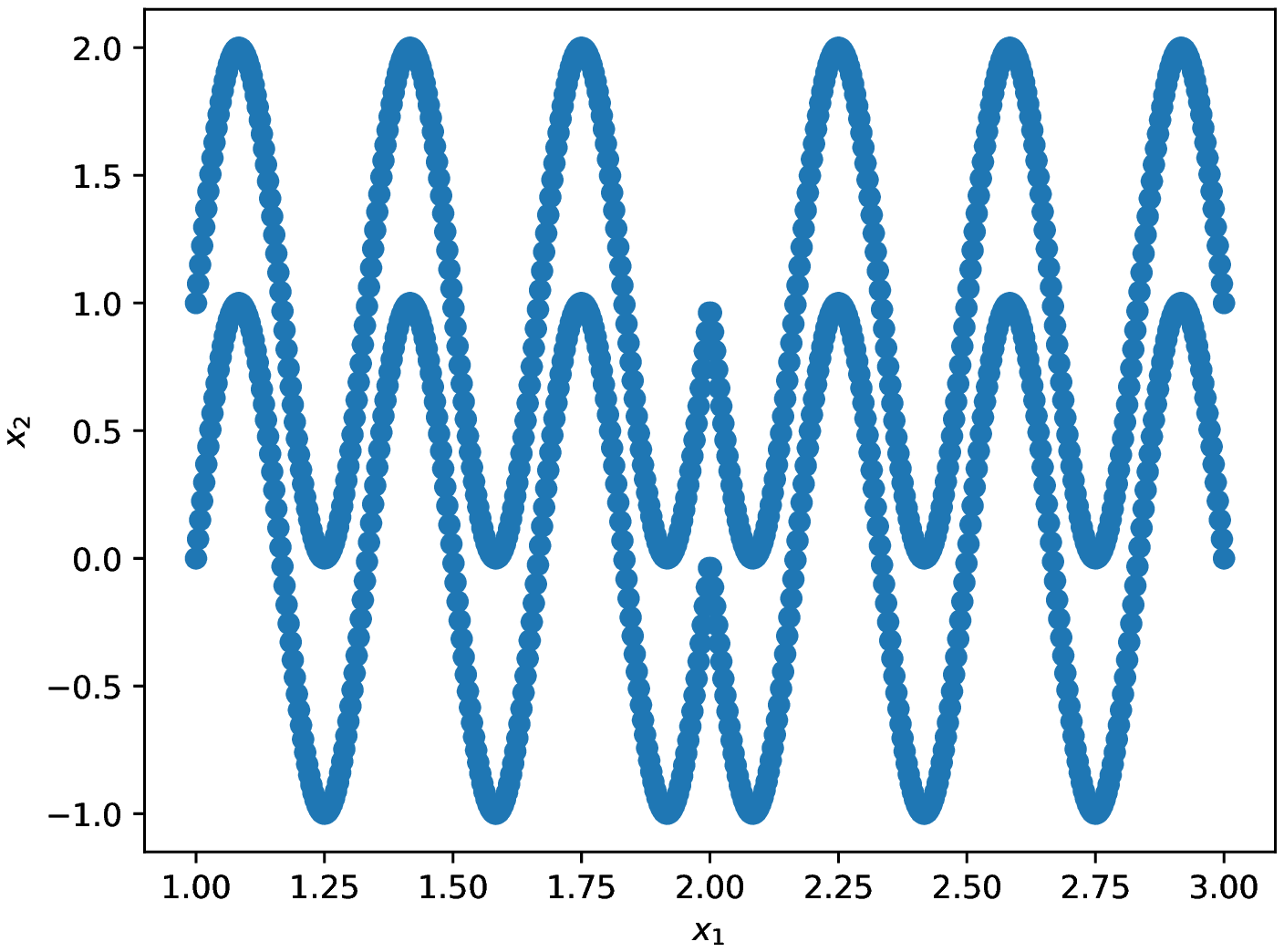}
}
\end{minipage}&
\begin{minipage}[b]{0.46\columnwidth}
\centering
\subfloat[MMF8]{
\includegraphics[width=\linewidth]{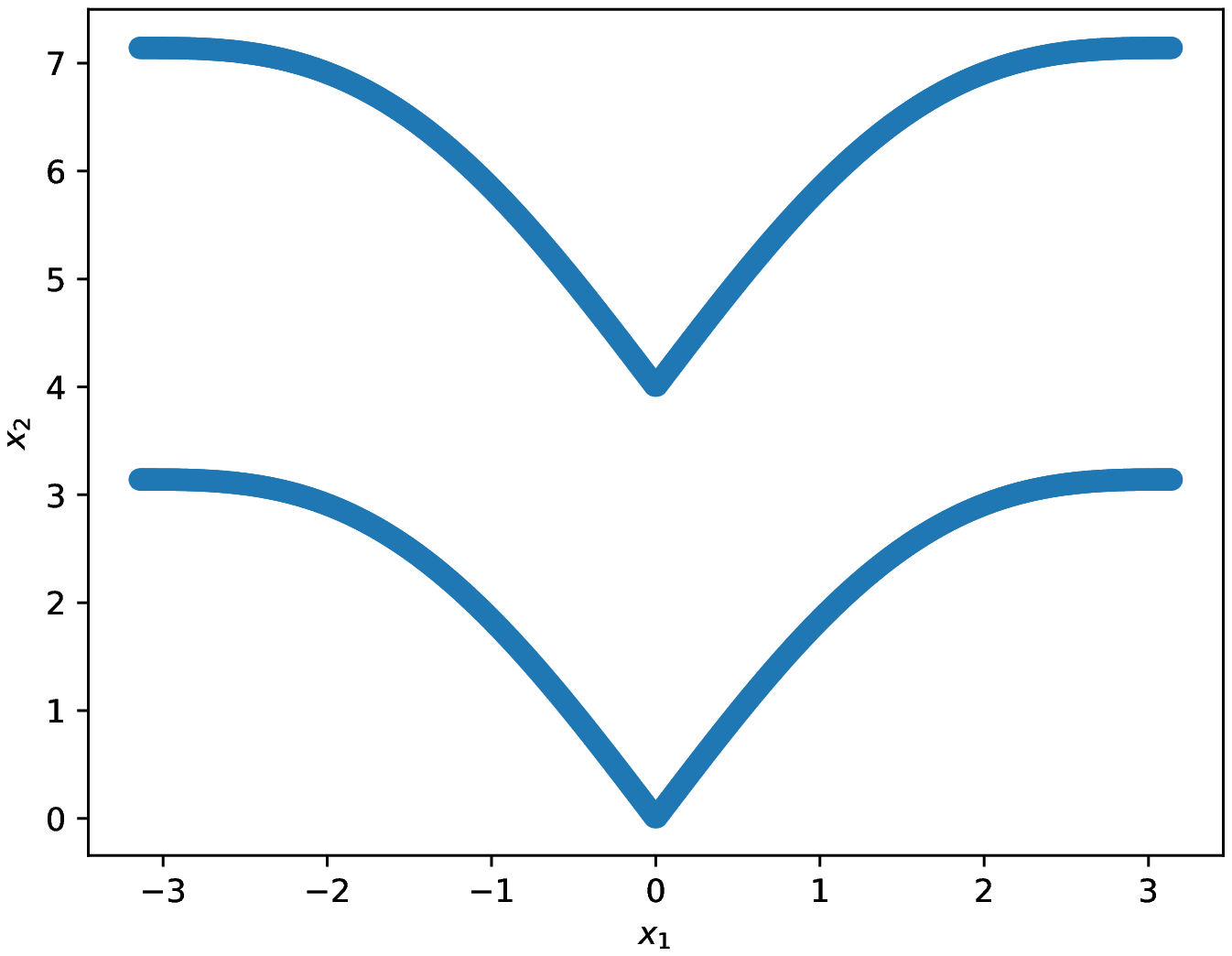}
}
\end{minipage}\\
\end{tabular}
\caption{The Pareto set of MMFs used in this work}
\label{fig:PS}
\end{figure}

This work designs multi-objective optimization test problems with the evaluation time bias to deeply analyze the behavior of the proposed method. In particular, this work uses multimodal multi-objective test functions (MMFs)~\citep{MMF2018}, which are bi-objective optimization problems with multiple separate Pareto sets (PS) in different regions of the decision space. One of the notable features of MMFs is that the Pareto front in the objective function space can be entirely approximated by only acquiring one of the separated Pareto sets. This study uses two-dimensional MMF2--6 and MMF8\footnote{This work does not use MMF1 and MMF7 because they have a continuous, non-separate Pareto set.}, which have two separated Pareto sets.  Fig.~\ref{fig:PS} depicts the Pareto set for each problem (see~\citep{MMF2018} for more detailed problem definitions). 
The Pareto set with a smaller $x_2$ value for each $x_1$ is defined as \textit{PS1}, while another one is defined as \textit{PS2} as follows:
\begin{align}
    PS1&=\{\bm{x}\mid \bm{x}=\{x_1, x_2\}\in P^*\land x_2<\theta(x_1)\},\nonumber\\
    PS2&=\{\bm{x}\mid \bm{x}=\{x_1, x_2\}\in P^*\land x_2\ge\theta(x_1)\},\nonumber\\
    \theta(x_1)&=0.5 \times \left(c_l(x_1)+c_s(x_1)\right),\nonumber
\end{align}
where $x_1$ and $x_2$ are the first and second design variables of $\bm{x}$, while $\theta(x_1)$ is a problem-dependent function that returns a boundary plane calculated by $x_1$. The functions $c_s(x)$ and $c_l(x)$ are defined for each problem as shown in Table~\ref{tb:c_func}, which is determined from the definition of the Pareto set. 
\begin{table}[tb]
\centering
\newcolumntype{C}{>{\centering\arraybackslash}X}
\caption{The definition of functions $c_s$ and $c_l$, and a parameter $\sigma$}
\label{tb:c_func}
\begin{tabularx}{\columnwidth}{cCc}
\toprule
Problem&Functions&$\sigma$\\
\midrule
MMF2&$
    \begin{array}{rcl}
c_s(x)&=&\sqrt{x}\\
c_l(x)&=&1+\sqrt{x}
    \end{array}\nonumber
    $&
0.25
\\
\midrule
MMF3&$
    \begin{array}{rcl}
c_s(x)&=&\sqrt{x}\\
c_l(x)&=&0.5+\sqrt{x}
\end{array}
$&
0.175
\\
\midrule
MMF4&$
    \begin{array}{rcl}
c_s(x)&=&\sin(\pi\lvert x\rvert)\\
c_l(x)&=&1+\sin(\pi\lvert x\rvert)
\end{array}
$&
0.25
\\
\midrule
MMF5&$
    \begin{array}{rcl}
c_s(x)&=&\sin(6\pi\lvert x-2\rvert+\pi)\\
c_l(x)&=&\sin(6\pi\lvert x-2\rvert+\pi)+2
\end{array}
$&
0.25
\\
\midrule
MMF6&$
    \begin{array}{rcl}
c_s(x)&=&\sin(6\pi\lvert x-2\rvert+\pi)\\
c_l(x)&=&\sin(6\pi\lvert x-2\rvert+\pi)+1
\end{array}
$&
0.375
\\
\midrule
MMF8&$
    \begin{array}{rcl}
c_s(x)&=&\sin(\lvert x\rvert)+\lvert x\rvert\\
c_l(x)&=&\sin(\lvert x\rvert)+\lvert x\rvert+4
\end{array}
$&
1.125
\\
\bottomrule
\end{tabularx}
\end{table}

Based on the Pareto set in MMFs, this work designs the evaluation time function so that the optimal solutions in PS2 require a longer evaluation time than those in PS1. Specifically, the biased evaluation time is defined as:
\begin{multline}
t_{bias}(\bm{x})=t_{mean}\left(1-\exp\left(-\frac{(x_2-c_s(x_1))^2}{2\sigma^2}\right)\right.\\
+\left.\exp\left(-\frac{(x_2-c_l(x_1))^2}{2\sigma^2}\right)\right).
\label{eq:et_bias}
\end{multline}
This work names $t_{bias}$ \textbf{Bias}. For each problem, $\sigma$ determines the variance of the Gaussian function, and the value of $\sigma$ is shown in Table~\ref{tb:c_func}.

\begin{figure}[tb]
\centering
\includegraphics[width=0.9\columnwidth]{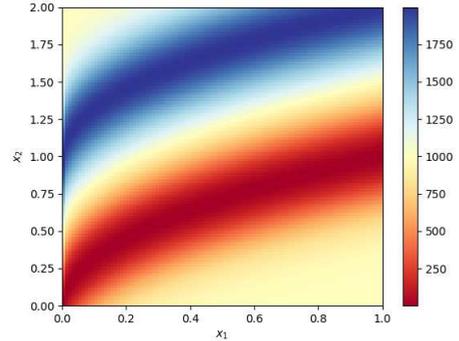}
\caption{A biased evaluation time in MMF2 calculated by Eq.~\eqref{eq:et_bias} where PS1 has a shorter evaluation time than PS2}
\label{fig:et_MMF2_bias1}
\end{figure}

An example of the evaluation time distribution of MMF2 is shown in Fig.~\ref{fig:et_MMF2_bias1}. The horizontal axis represents $x_1$, while the vertical axis represents $x_2$. The color bar indicates the evaluation time. In this setting, solutions in the PS1 have a shorter evaluation time than those in the PS2. It is expected that APEA will converge more quickly to the PS1 in such a situation.

In addition to \textbf{Bias}, this experiment uses a non-biased evaluation time function that returns a random value sampled from the normal distribution as $t_{norm}(\bm{x})\sim N(t_p, c_vt_p)$. The variable $t_p$ denotes the mean evaluation time, while $c_v$ determines the variance of the evaluation time.
Since $t_{norm}(\bm{x})$ is independent of the decision variable, there is no bias in the evaluation time --- name this \textbf{No-bias}.

\section{Experimental setting}\label{sec:setting}
This paper conducts experiments on the simulated parallel computational environment to investigate the effectiveness of the proposed method. The proposed method is applied to NSGA-III as a concrete algorithm shown in Section~\ref{sec:example}. The experiments compare three parallel NSGA-IIIs, synchronous parallelization (SP-NSGA-III),  asynchronous parallelization (AP-NSGA-III), and the proposed parallelization (FS-NSGA-III).
Note that this paper does not aim to solve multimodal multi-objective optimization problems efficiently but uses them to analyze the influence of evaluation time bias. Thus, this study does not uses specific techniques for finding niches.

This section first explains the simulation environment used in the experiment, and then the parameter settings used in the experiment are shown.
The final subsection provides evaluation criteria for assessing the competitive methods.

\subsection{Simulated parallel computational environment}
The experiments use a simulated parallel computational environment based on the computational time model proposed in the work of \citep{SSAMSPS}. 
This model consists of a single master node and $\lambda$ worker nodes.
The master node computes the main procedure of the EA algorithm in $t_s=1$ simulation time. 
In contrast, the worker nodes evaluate one solution and return their evaluation results. This experiment simulates $\lambda=100$ worker nodes where 100 solutions are simultaneously evaluated. The evaluation times on the worker nodes depend on the \textbf{Bias} and \textbf{No-bias} functions. In \textbf{Bias}, the value of $t_{mean}$ is set to $1000$. In such a setting, the maximum evaluation time is almost $2000$, while the minimum one is almost one, so the longest evaluation time is 2000 times longer than the shortest one.
On the other hand, in \textbf{No-bias}, the mean evaluation time $t_p=1000$, while the variance parameter $c_v=0.2$.

\subsection{Parameters}
The experiments were conducted for 31 independent runs for each parallelization method.
The population size is 100, which means all solutions in the population can be evaluated simultaneously in SP-NSGA-III. The maximum number of evaluations is $8.0\times 10^4$, corresponding to 800 generations in SP-NSGA-III. 
As the genetic operator, the simulated binary crossover (SBX) with the probability of $p_c=1.0$ and the distribution index of $\eta_c=30.0$ is used, and the polynomial mutation (PM) with the probability of $p_m=1/D$ and the distribution index of $\eta_m=20.0$.

\subsection{Evaluation criteria}
This experiment uses the inverted generational distance ($IGD$) indicator~\citep{IGD} to assess the quality of the obtained solutions in the objective space. The $IGD$ value is calculated as:
\begin{equation}
    IGD(P^*, P)=\frac{1}{\lvert P^*\rvert}\sum_{\bm{a}\in P^*}\min_{\bm{p} \in P}d(\bm{f}(\bm{a}), \bm{f}(\bm{p}))
    \label{eq:IGD}
\end{equation}
where $P^*$ denotes a reference point set (the true Pareto solution set), while $P$ denotes the non-dominated solutions obtained by the algorithm. The function $d(\bm{x}, \bm{y})$ calculates the Euclidean distance between $\bm{x}$ and $\bm{y}$.
The solutions obtained by the algorithm are worthful if the $IGD$ value is small.

In addition, the $IGDX$ indicator~\citep{IGDX} is used to evaluate the quality of solutions in the design variable space. The $IGDX$ value is calculated as:
\begin{equation}
    IGDX(P^*, P)=\frac{1}{\lvert P^* \rvert}\sum_{\bm{a}\in P^*}\min_{\bm{p} \in P}d(\bm{a}, \bm{p})
    \label{eq:IGDX}
\end{equation}
where $P^*$ and $P$ denote the true Pareto solution set and the obtained one by the algorithm.
When calculating $IGD$, the distance is calculated on the objective space. On the other hand, when calculating $IGDX$, the distance on the design variable is calculated.

This work independently calculates the $IGDX$ values for two separate Pareto sets to confirm whether both Pareto sets are obtained simultaneously.
The $IGDX$ values for PS1 and PS2 are denoted as $IGDX_1$ and $IGDX_2$, respectively.
To assess if both Pareto sets are equally obtained, the difference between two $IGDX$ values is defined as:
\begin{multline}
    \Delta IGDX(P^*, P)\\
    =IGDX(PS1, P)-IGDX(PS2, P).\label{eq:dIGDX}
\end{multline}
If $\Delta IGDX=0$, both Pareto sets are equally obtained. On the other hand, if $\Delta IGDX<0$, since $IGDX_1<IGDX_2$, the algorithm is biased to PS1, and vice versa. In the experiment using \textbf{Bias}, the $\Delta IGDX$ value of SP-NSGA-III is expected to be 0, while that of AP-NSGA-III may be less than 0 because its search direction is biased to PS1.
It can be expected that the proposed method shows similar behavior to the synchronous method by reducing the effect of the evaluation time bias.

The Kruskal-Wallis test will be performed to confirm a statistical difference between the three parallelization methods for each criterion. If a significant difference is found with the Kruskal-Wallis test, we perform the post-hoc pairwise comparisons using the Wilcoxon rank-sum test  with the Bonferroni adjustments.

\section{Comparison of Selection Ratios}
\label{sec:exp2}
\begin{figure*}[tb]
\begin{tabular}{cccccc}
\begin{minipage}[b]{0.14\textwidth}
\centering
\subfloat[MMF2]{\includegraphics[scale=0.3]{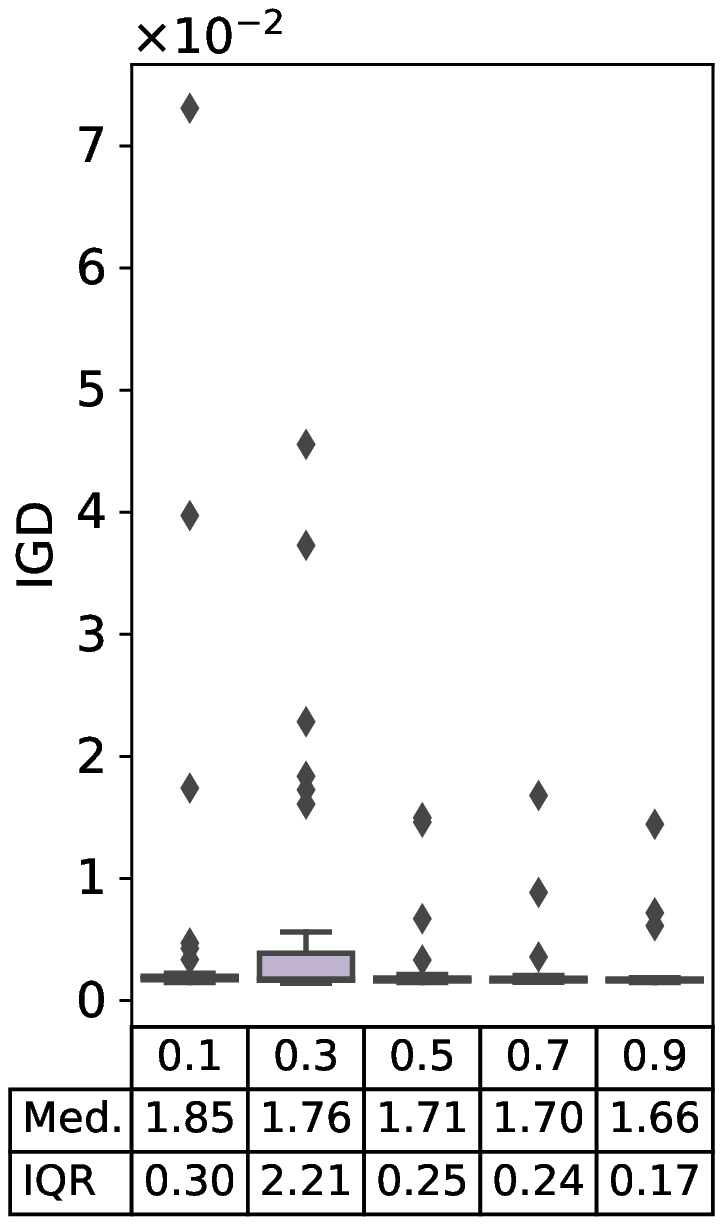}}
\end{minipage}
&
\begin{minipage}[b]{0.14\textwidth}
\centering
\subfloat[MMF3]{\includegraphics[scale=0.3]{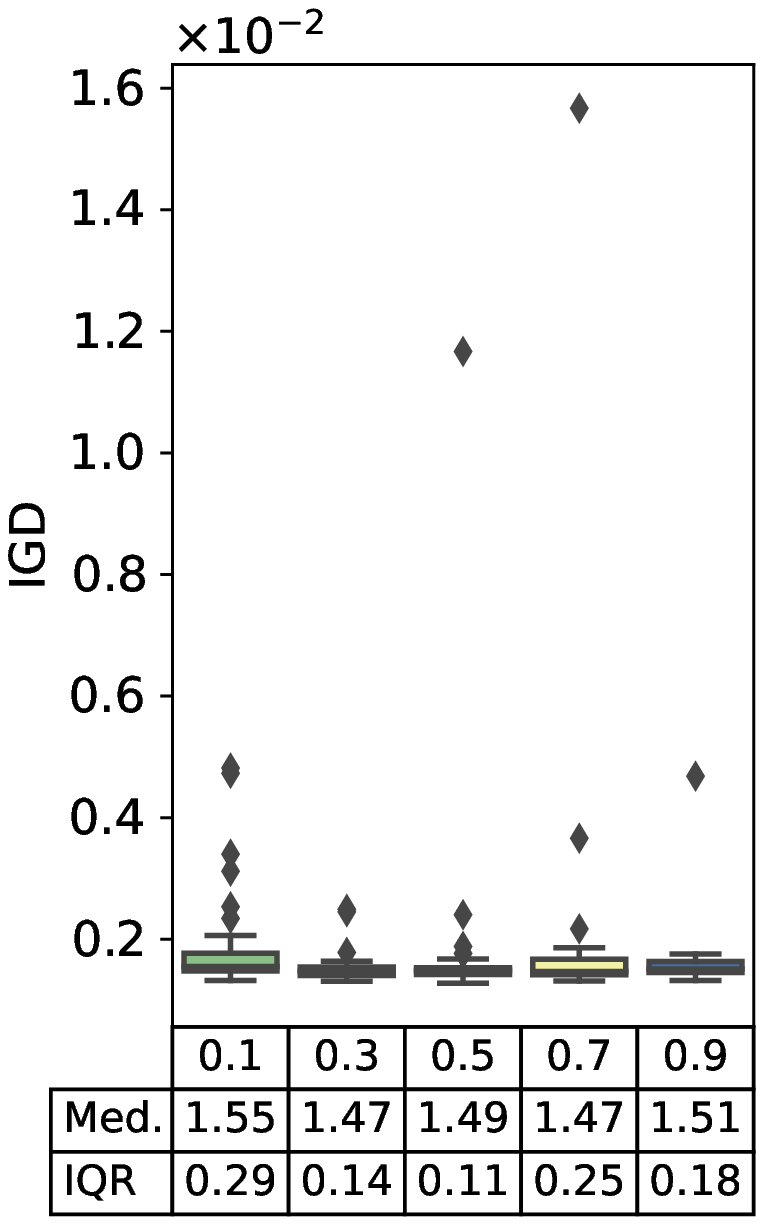}}
\end{minipage}
&
\begin{minipage}[b]{0.14\textwidth}
\centering
\subfloat[MMF4]{\includegraphics[scale=0.3]{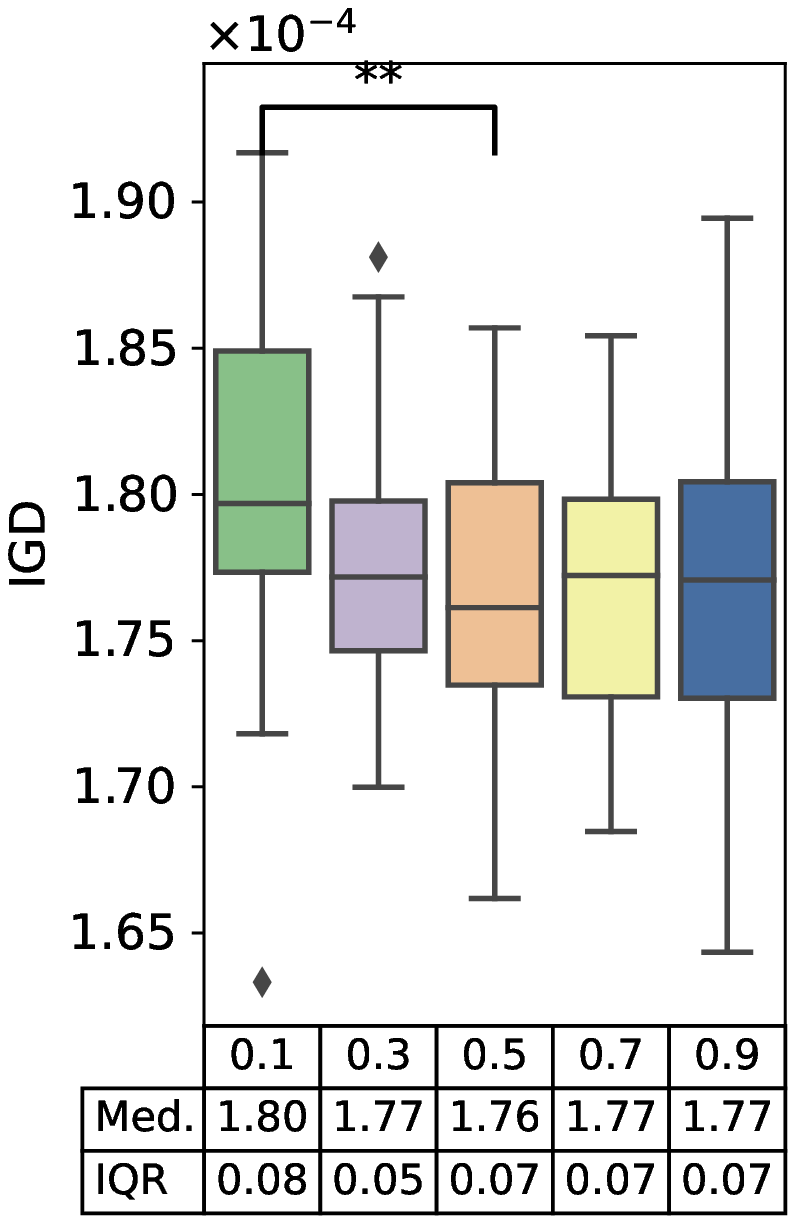}}
\end{minipage}
&
\begin{minipage}[b]{0.14\textwidth}
\centering
\subfloat[MMF5]{\includegraphics[scale=0.3]{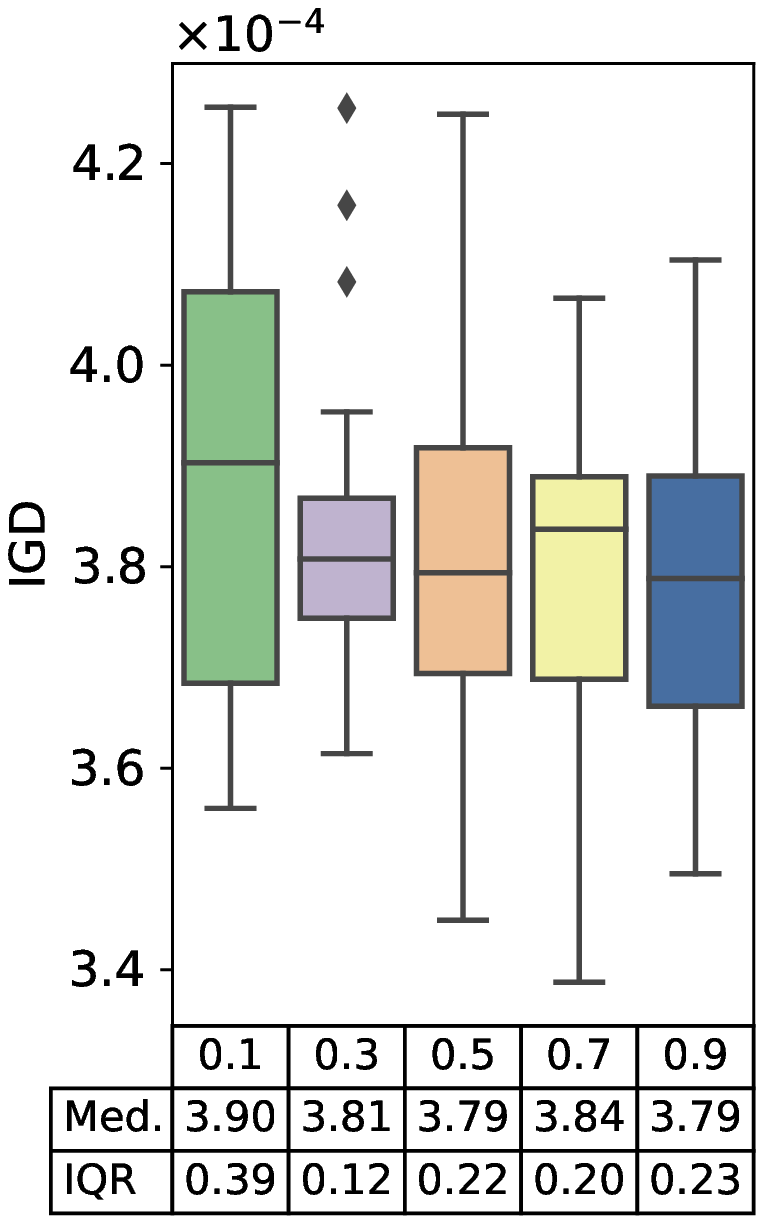}}
\end{minipage}
&
\begin{minipage}[b]{0.14\textwidth}
\centering
\subfloat[MMF6]{\includegraphics[scale=0.3]{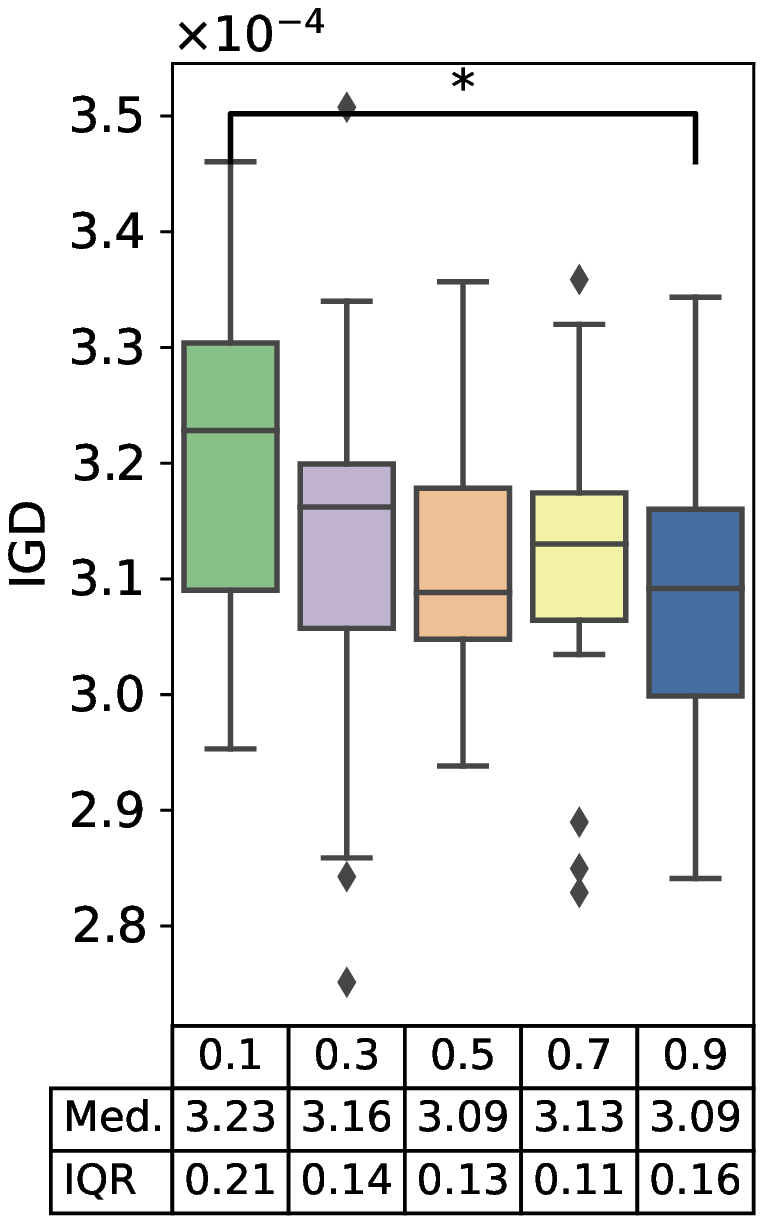}}
\end{minipage}
&
\begin{minipage}[b]{0.14\textwidth}
\centering
\subfloat[MMF8]{\includegraphics[scale=0.3]{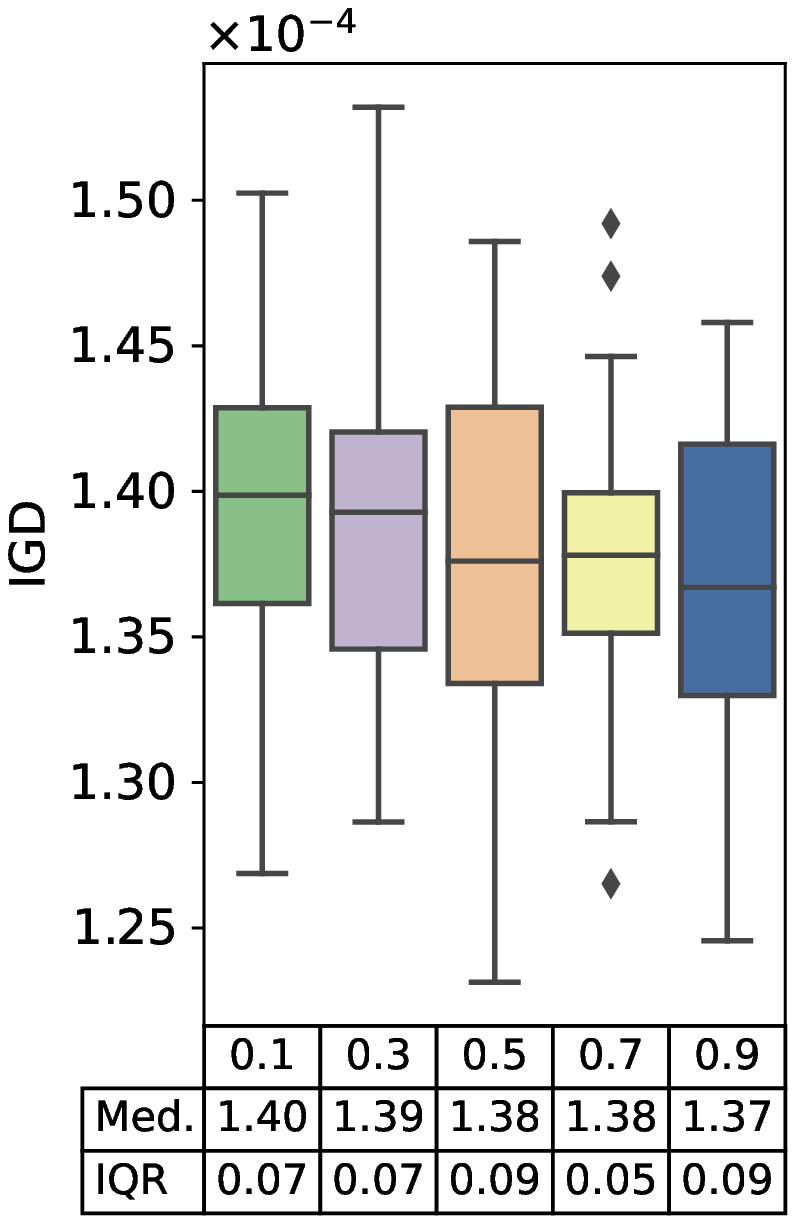}}
\end{minipage}
\end{tabular}
\caption{$IGD$ with \textbf{No-bias} after the maximum fitness evaluations (different $r_s$)}
\label{fig:igd_bp_nobias_exp2}
\begin{tabular}{cccccc}
\begin{minipage}[b]{0.14\textwidth}
\centering
\subfloat[MMF2]{\includegraphics[scale=0.3]{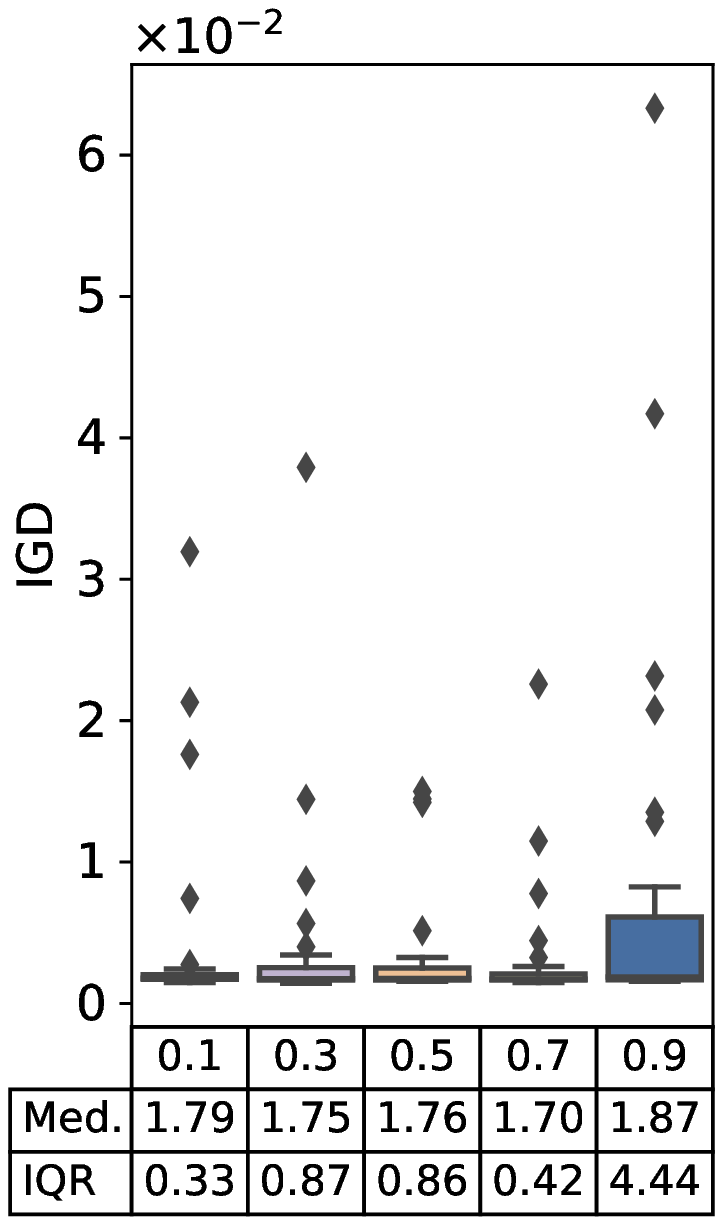}}
\end{minipage}
&
\begin{minipage}[b]{0.14\textwidth}
\centering
\subfloat[MMF3]{\includegraphics[scale=0.3]{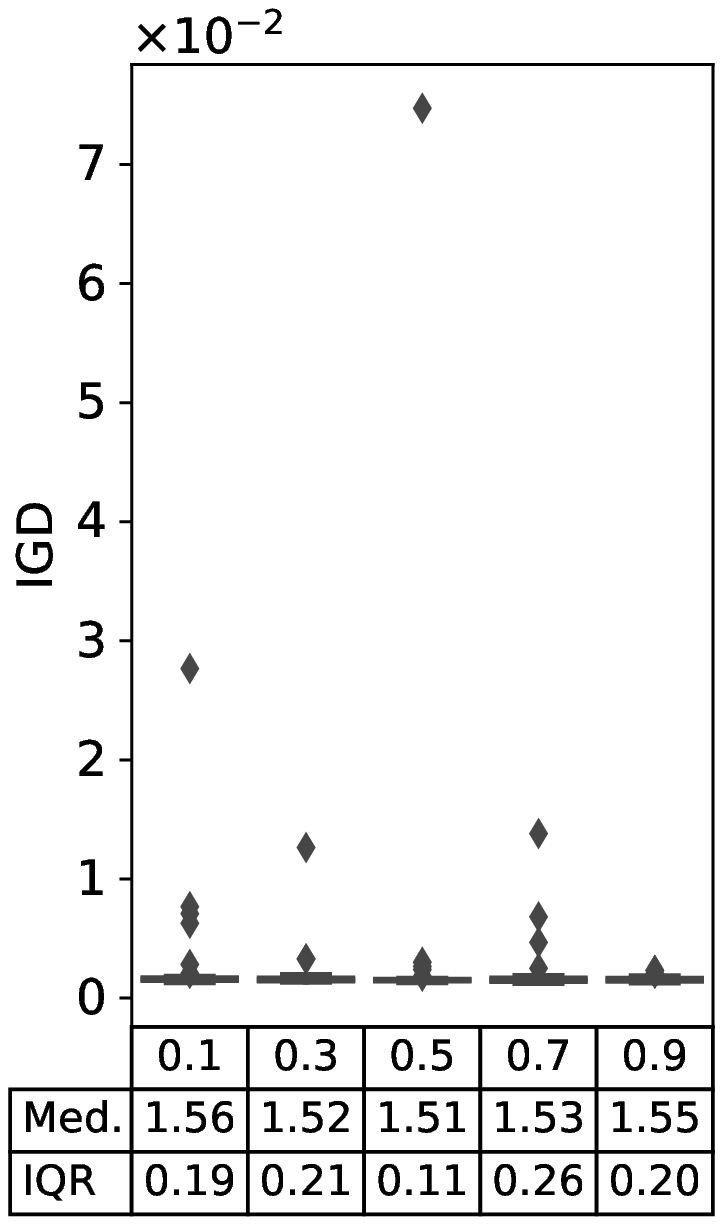}}
\end{minipage}
&
\begin{minipage}[b]{0.14\textwidth}
\centering
\subfloat[MMF4]{\includegraphics[scale=0.3]{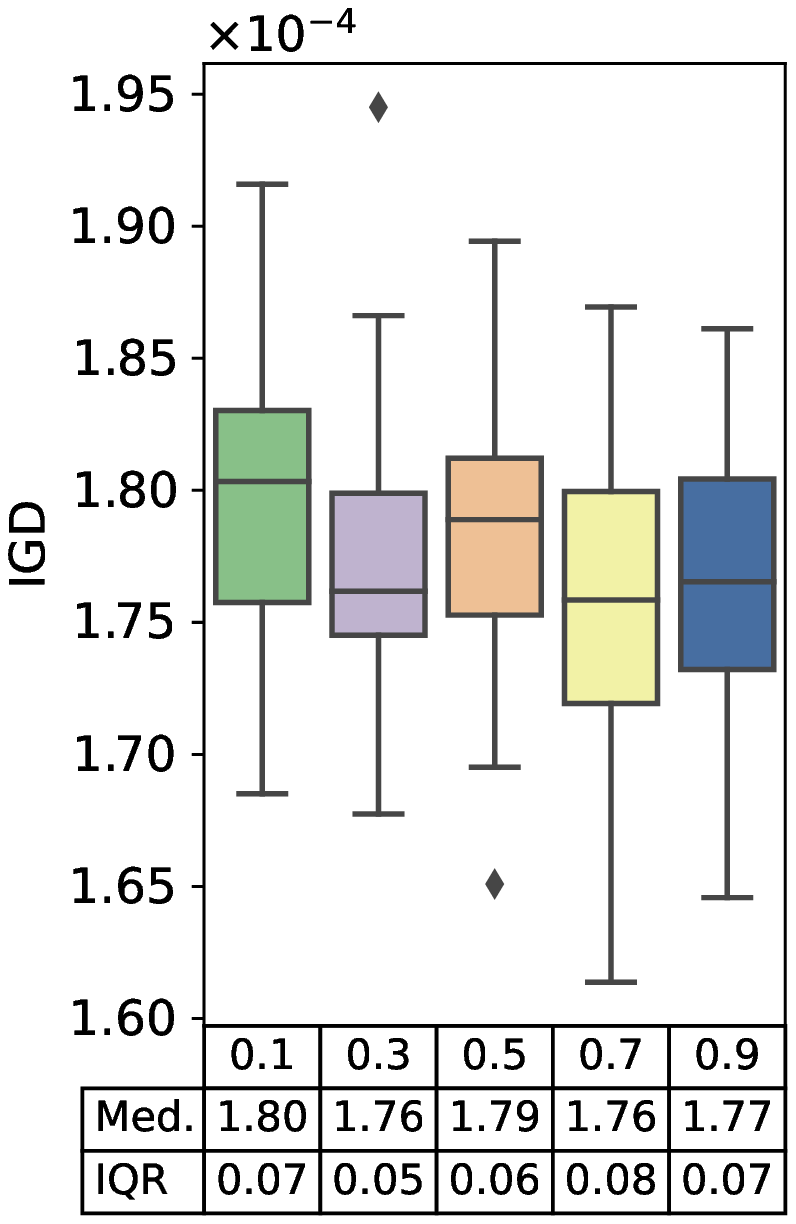}}
\end{minipage}
&
\begin{minipage}[b]{0.14\textwidth}
\centering
\subfloat[MMF5]{\includegraphics[scale=0.3]{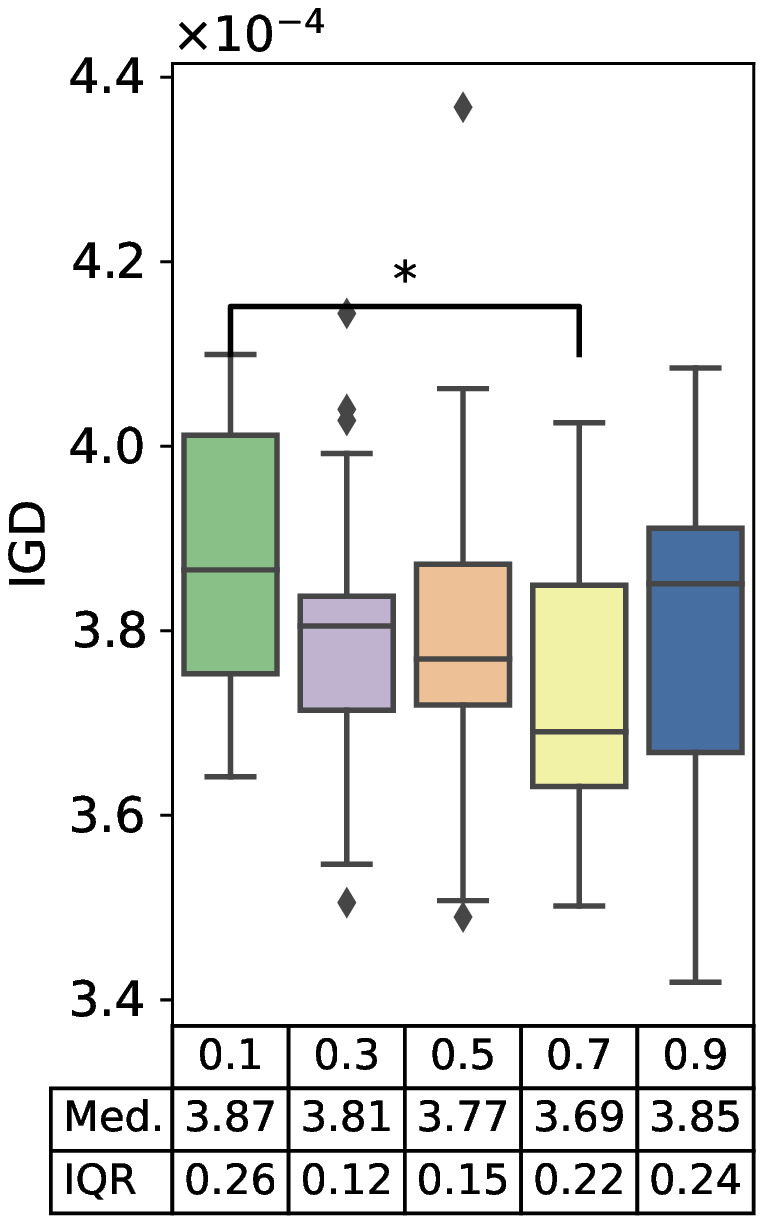}}
\end{minipage}
&
\begin{minipage}[b]{0.14\textwidth}
\centering
\subfloat[MMF6]{\includegraphics[scale=0.3]{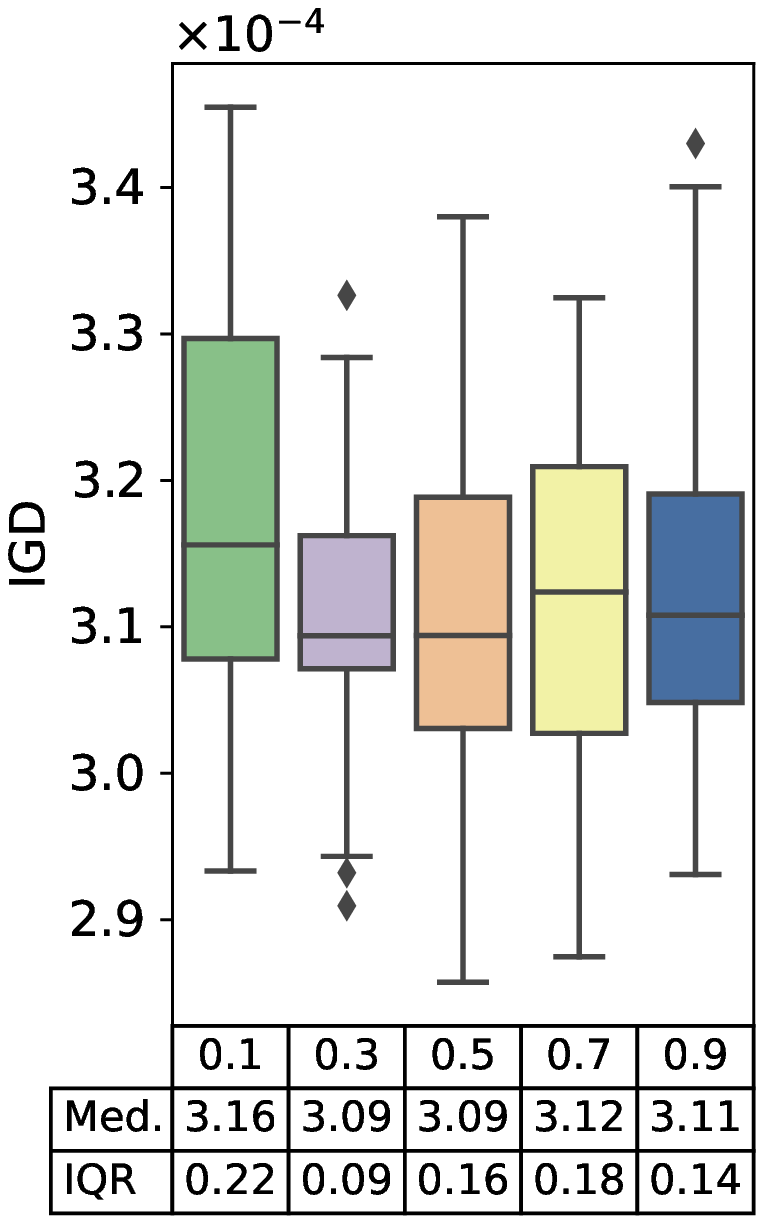}}
\end{minipage}
&
\begin{minipage}[b]{0.14\textwidth}
\centering
\subfloat[MMF8]{\includegraphics[scale=0.3]{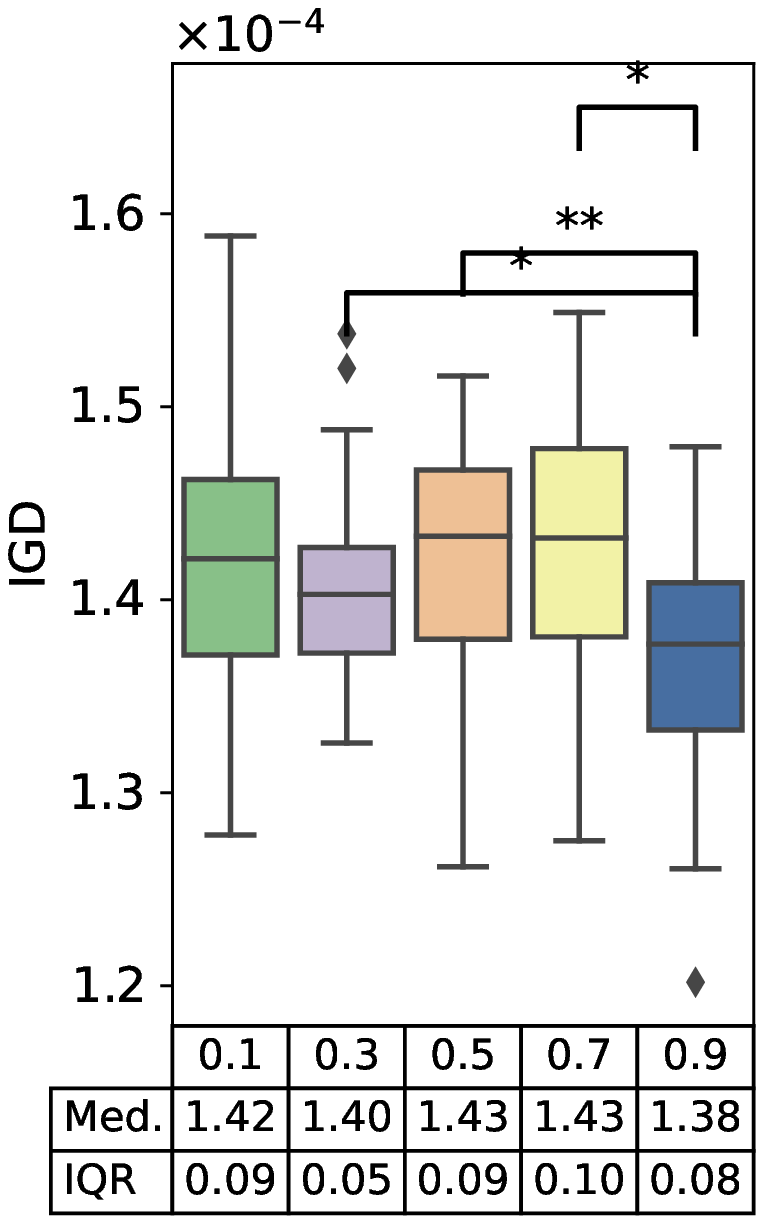}}
\end{minipage}
\end{tabular}
\caption{$IGD$ with \textbf{Bias} after the maximum fitness evaluations (different $r_s$)}
\label{fig:igd_bp_bias1_exp2}
\end{figure*}
This section analyzes how the selection ratio of the proposed method ($r_s$ in Algorithm~\ref{alg:apea_with_proposed}) affects the search capability and computational efficiency.
The experiments compare five selection ratios $r_s=\{0.1, 0.3, 0.5, 0.7, 0.9\}$.
The following subsections first show the results from the search capability ($IGD$) viewpoints. Then, the effect of the evaluation time bias is analyzed using the $\Delta IGDX$ value. Finally, the computational efficiency of different selection ratios is evaluated by comparing the simulation execution time until a particular quality of solutions is obtained.

\subsection{Search capability}
\label{sec:exp2_capability}
Figures~\ref{fig:igd_bp_nobias_exp2} and \ref{fig:igd_bp_bias1_exp2} show the boxplot of the $IGD$ value after the maximum number of evaluations for \textbf{No-bias} and \textbf{Bias}, and  the bottom table summarizes the median and IQR values (the difference between the third and the first quartiles).
The horizontal axis shows the selection ratio $r_s$, while the vertical axis shows the $IGD$ value.
The two boxes are connected with the ``*'' symbol if a significant difference with a significance level of 5\% is found, and the ``**'' symbol if the significance level is 1\%.

\begin{figure*}[tb]
\begin{tabular}{cccccc}
\begin{minipage}[b]{0.14\textwidth}
\subfloat[MMF2]{\includegraphics[scale=0.3]{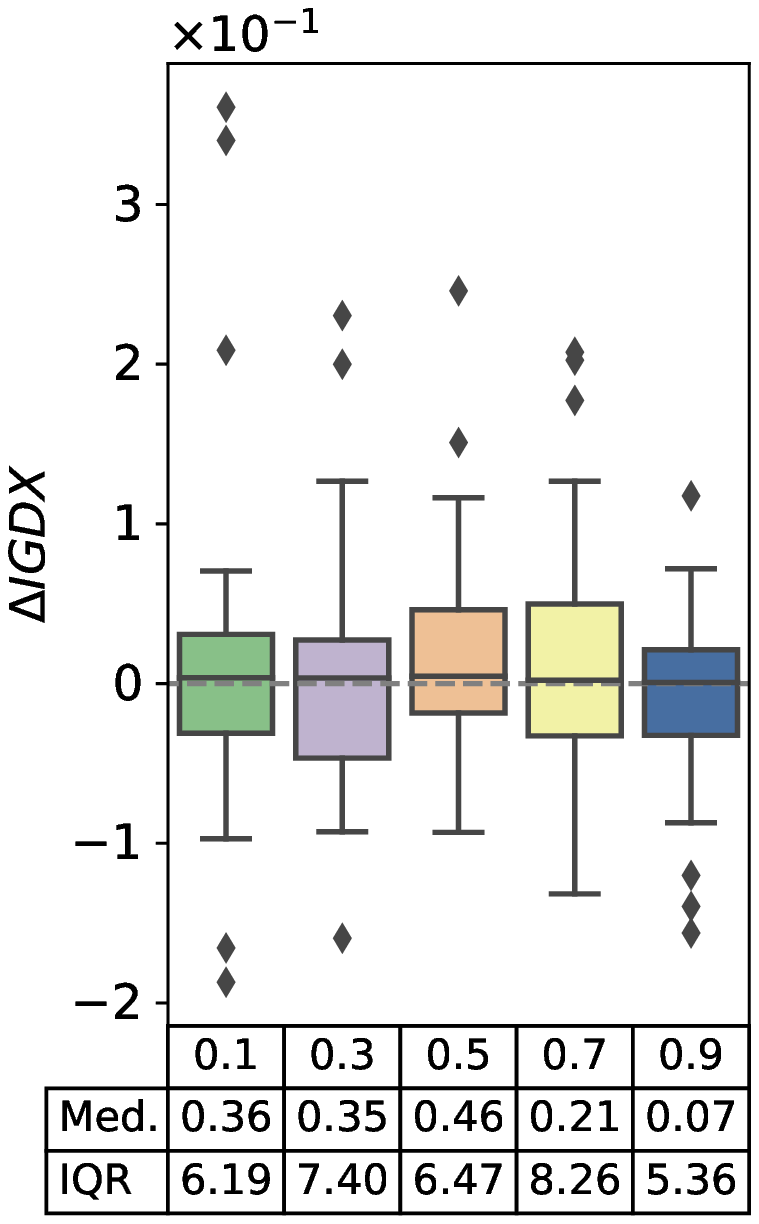}}
\end{minipage}
&
\begin{minipage}[b]{0.14\textwidth}
\subfloat[MMF3]{\includegraphics[scale=0.3]{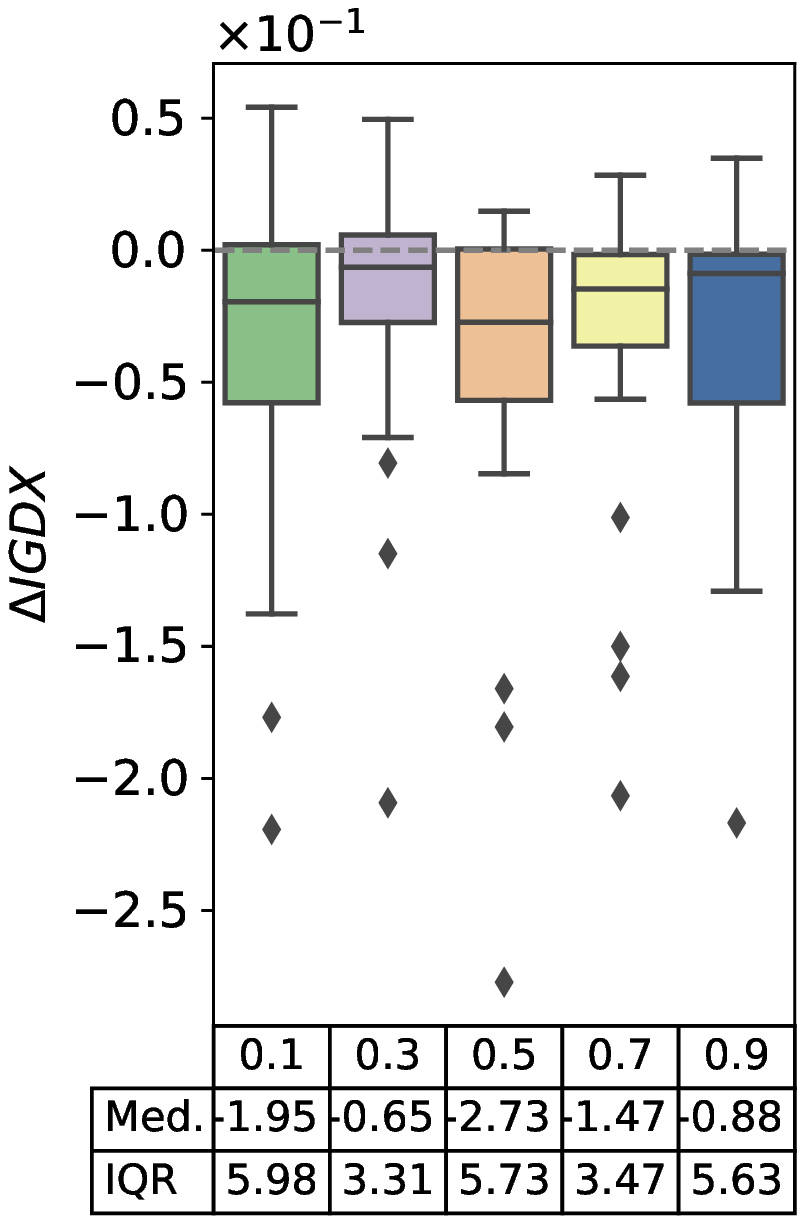}}
\end{minipage}
&
\begin{minipage}[b]{0.14\textwidth}
\subfloat[MMF4]{\includegraphics[scale=0.3]{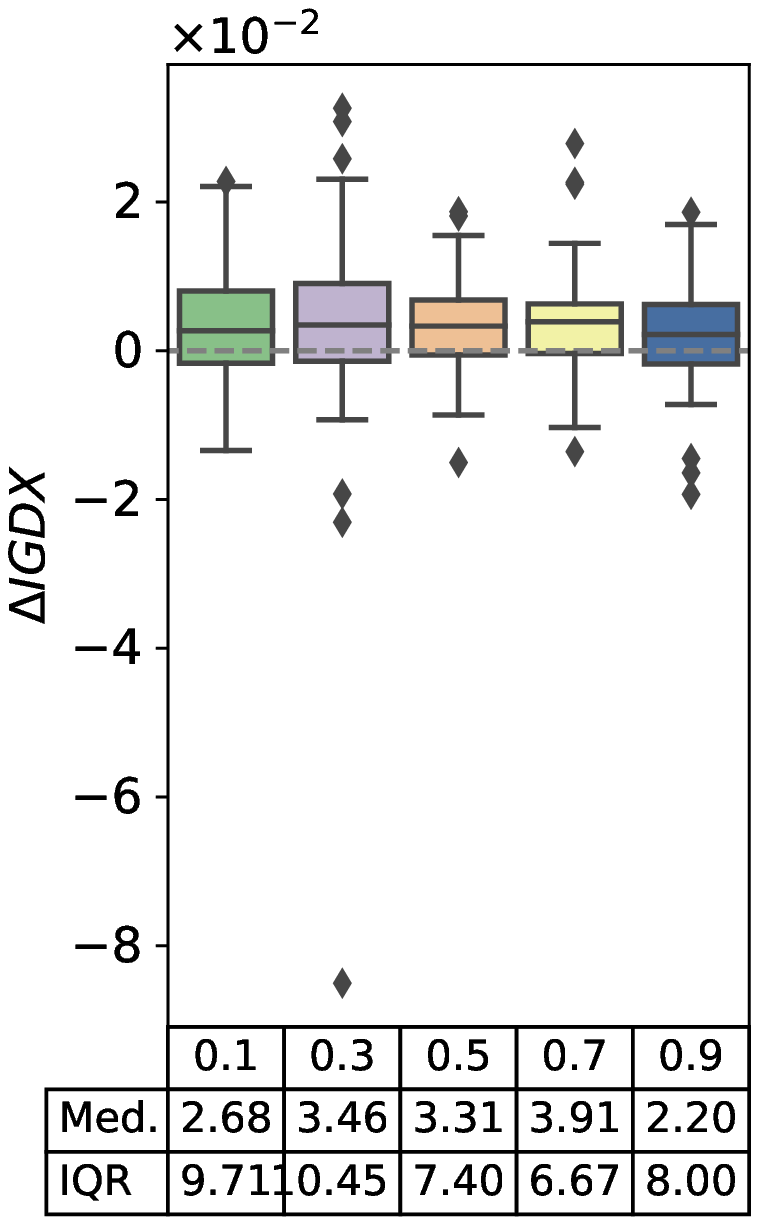}}
\end{minipage}
&
\begin{minipage}[b]{0.14\textwidth}
\subfloat[MMF5]{\includegraphics[scale=0.3]{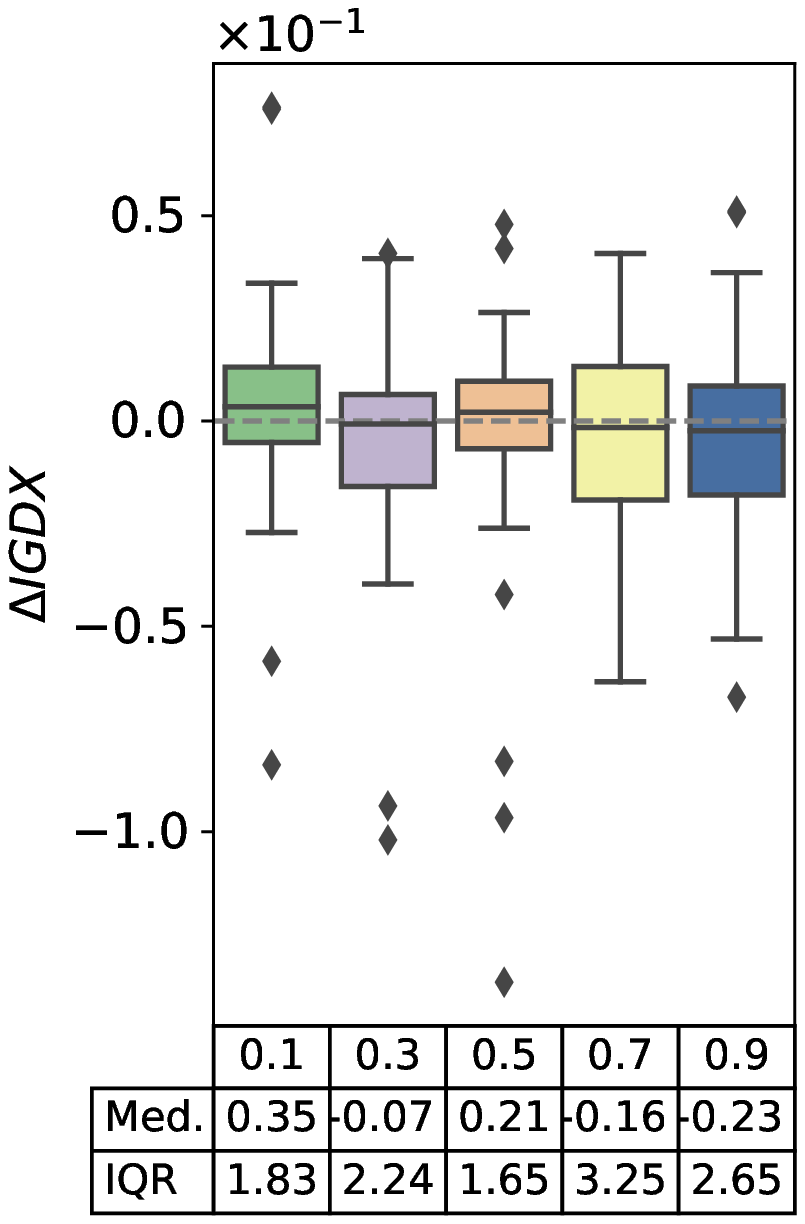}}
\end{minipage}
&
\begin{minipage}[b]{0.14\textwidth}
\subfloat[MMF6]{\includegraphics[scale=0.3]{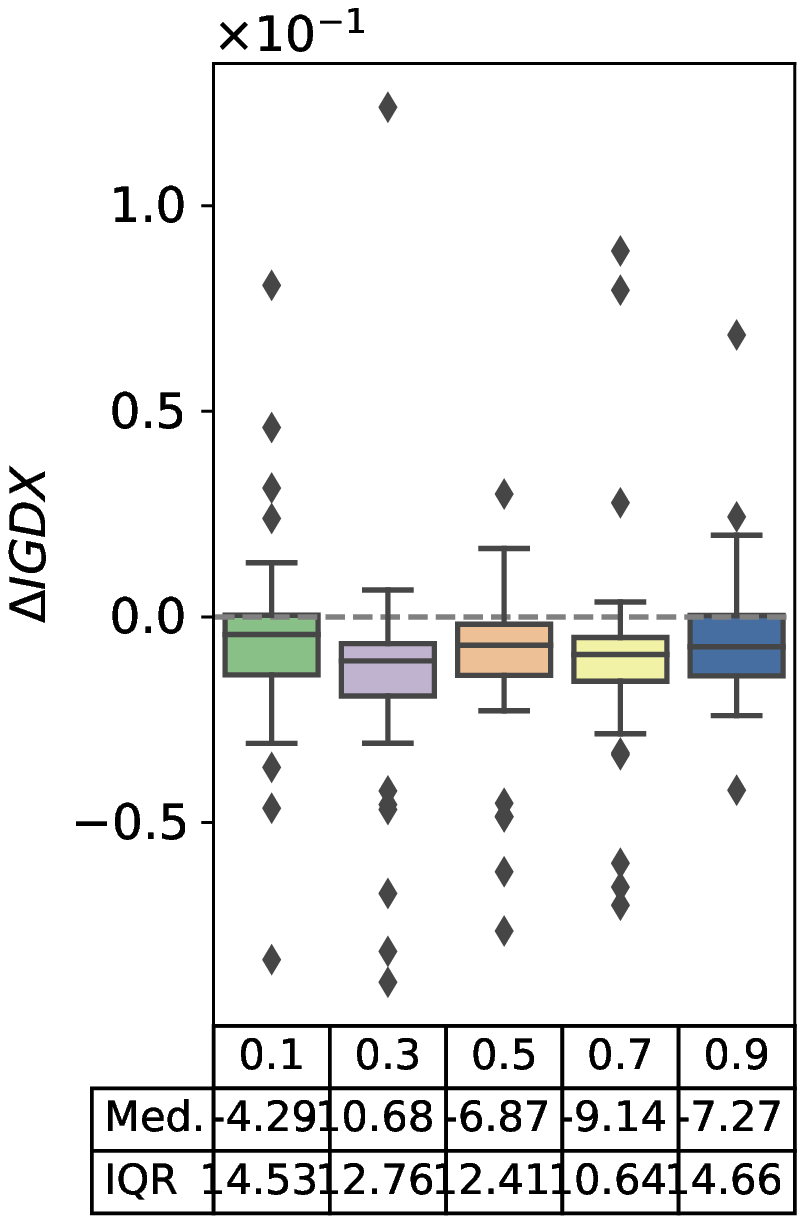}}
\end{minipage}
&
\begin{minipage}[b]{0.14\textwidth}
\subfloat[MMF8]{\includegraphics[scale=0.3]{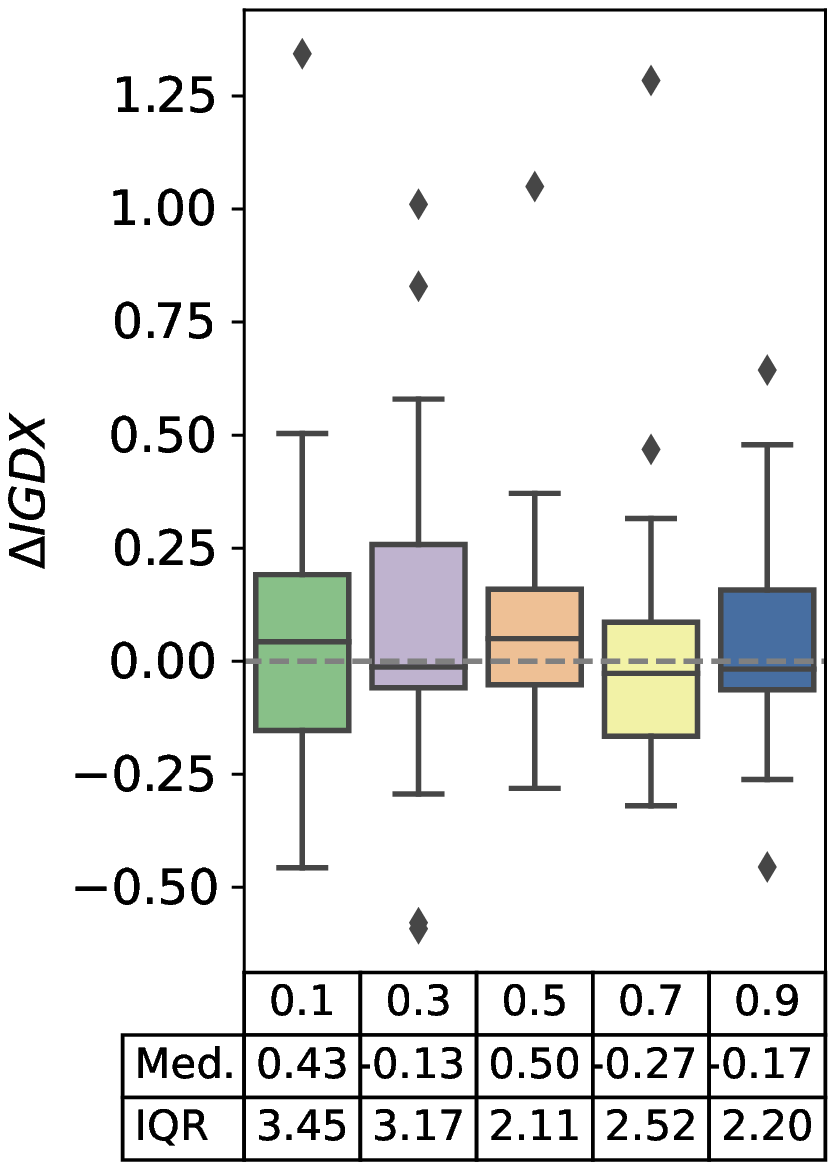}}
\end{minipage}
\end{tabular}
\caption{$\Delta IGDX$ with \textbf{No-bias} after the maximum fitness evaluations (different $r_s$)}
\label{fig:delta_igdx_nobias_exp2}
\begin{tabular}{cccccc}
\begin{minipage}[b]{0.14\textwidth}
\subfloat[MMF2]{\includegraphics[scale=0.3]{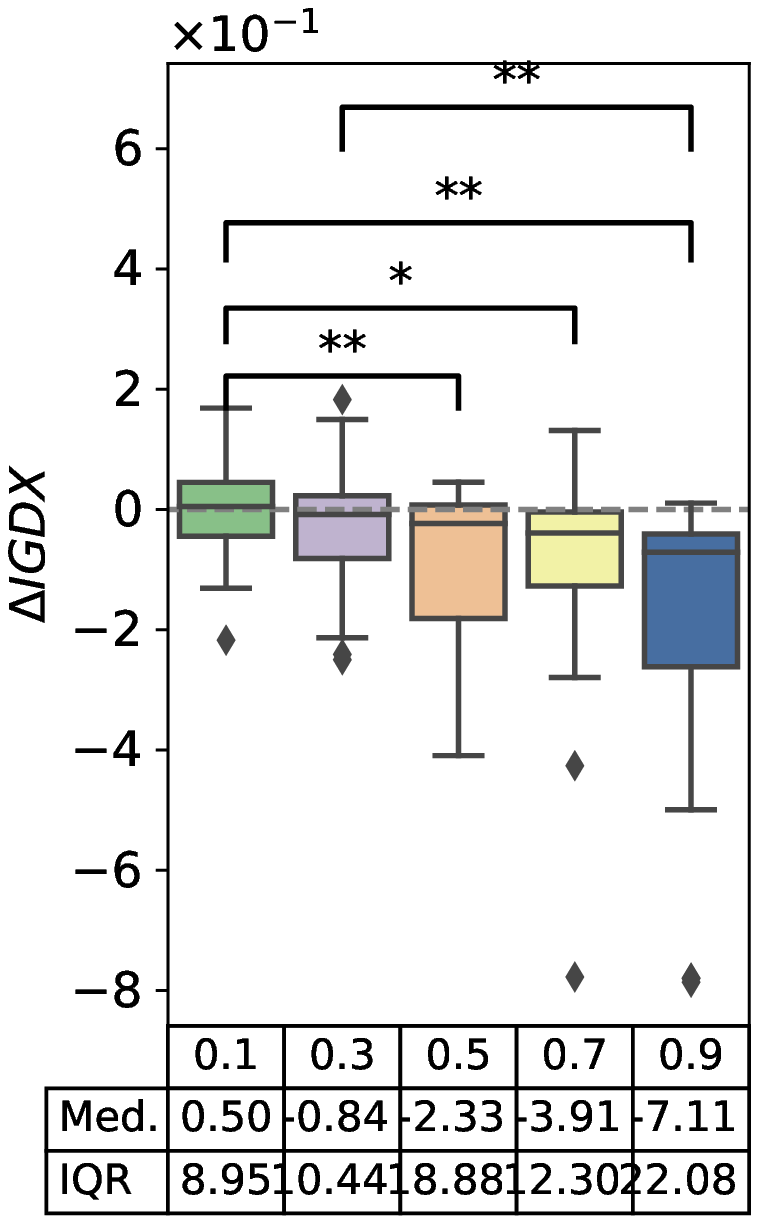}}
\end{minipage}
&
\begin{minipage}[b]{0.14\textwidth}
\subfloat[MMF3]{\includegraphics[scale=0.3]{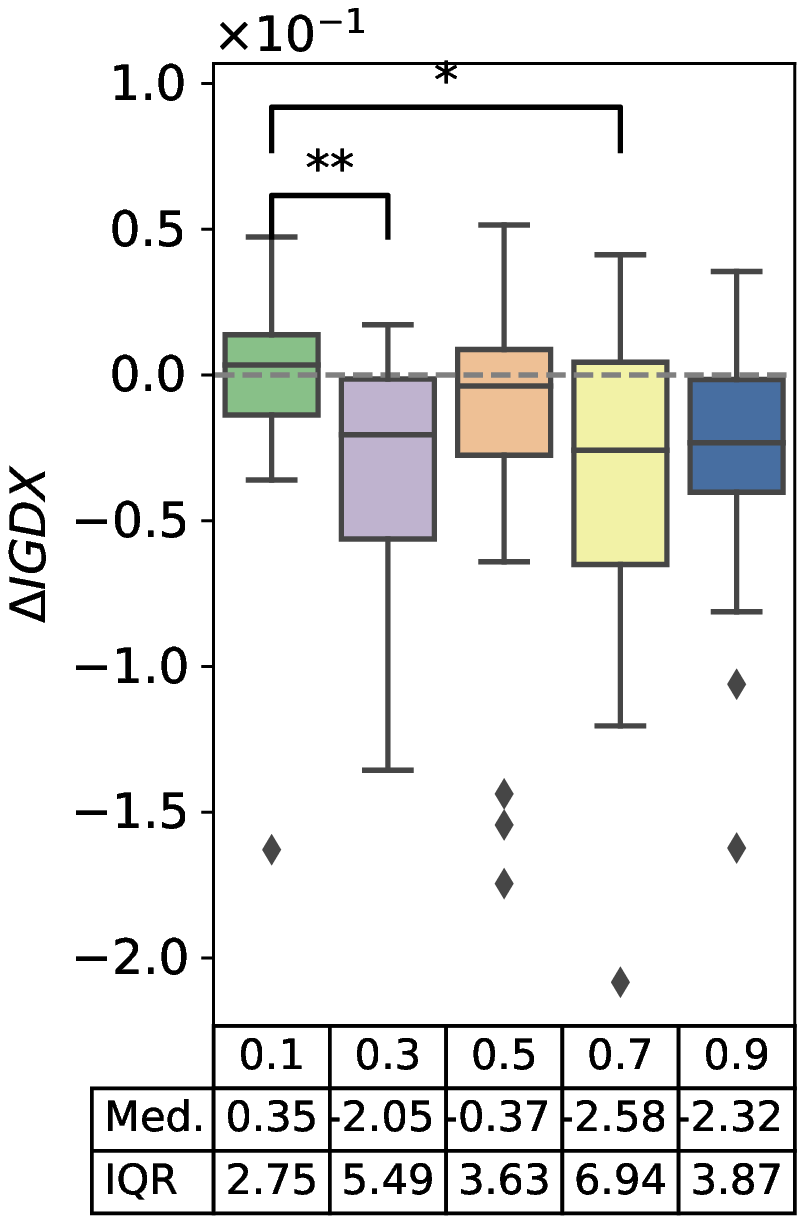}}
\end{minipage}
&
\begin{minipage}[b]{0.14\textwidth}
\subfloat[MMF4]{\includegraphics[scale=0.3]{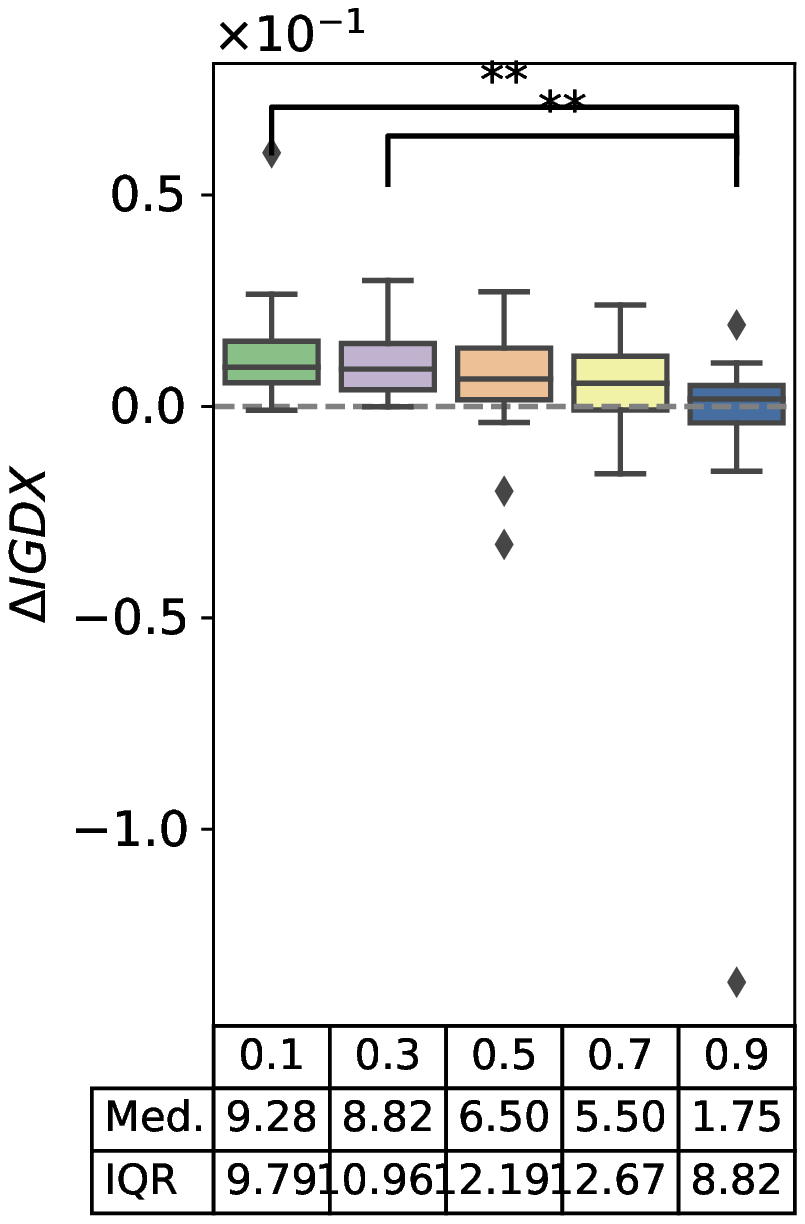}}
\end{minipage}
&
\begin{minipage}[b]{0.14\textwidth}
\subfloat[MMF5]{\includegraphics[scale=0.3]{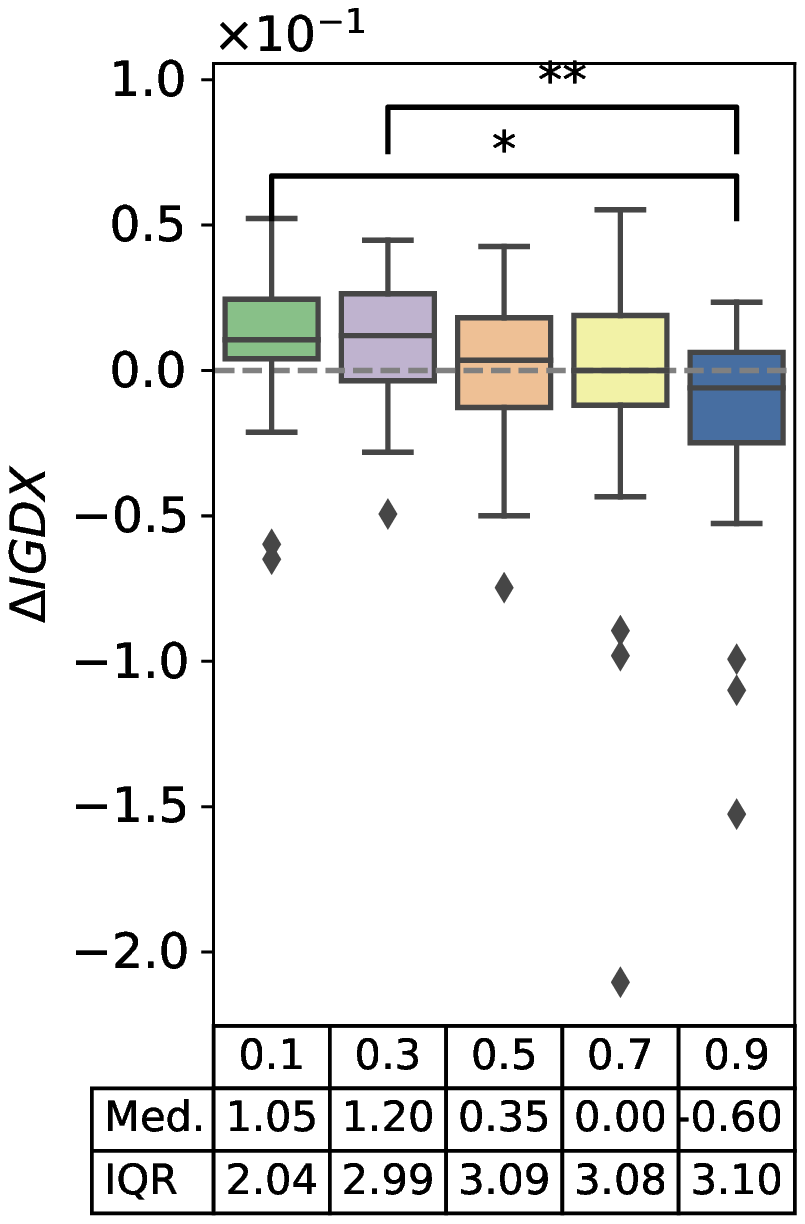}}
\end{minipage}
&
\begin{minipage}[b]{0.14\textwidth}
\subfloat[MMF6]{\includegraphics[scale=0.3]{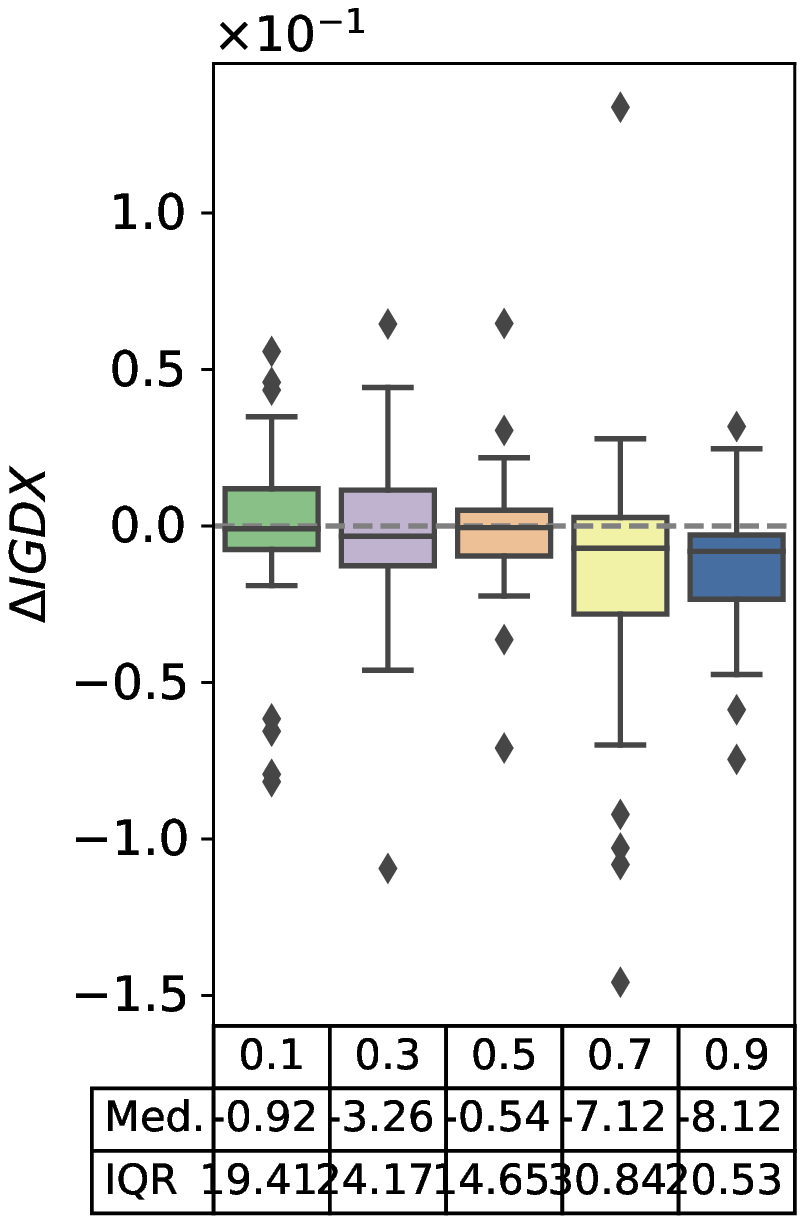}}
\end{minipage}
&
\begin{minipage}[b]{0.14\textwidth}
\subfloat[MMF8]{\includegraphics[scale=0.3]{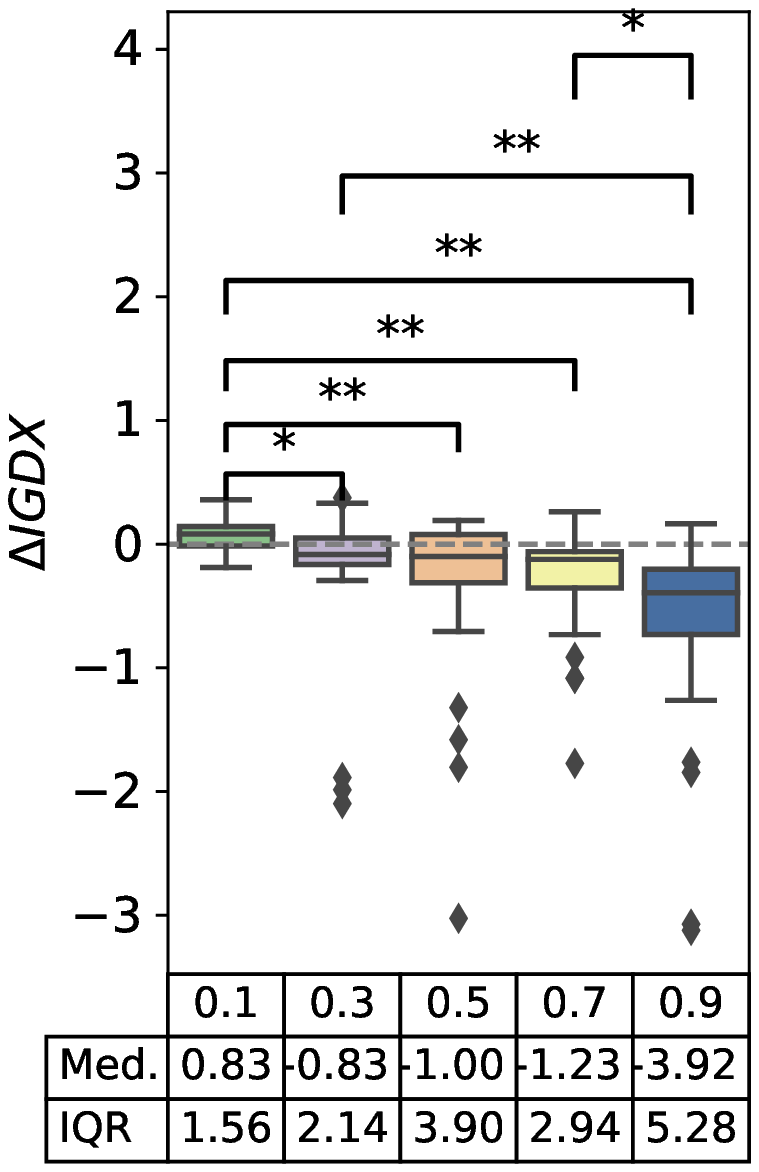}}
\end{minipage}
\end{tabular}
\caption{$\Delta IGDX$ with \textbf{Bias} after the maximum fitness evaluations (different $r_s$)}
\label{fig:delta_igdx_bias1_exp2}
\end{figure*}
These results show that $r_s=0.1$ obtains a significantly worse $IGD$ value in MMF4, MMF5, and MMF6. Moreover, the selection ratio of $r_s=0.1$ shows relatively worse $IGD$ values than the other settings in other problems. These results indicate that a small $r_s$ negatively affects the search capability in \textbf{No-bias} and \textbf{Bias}.
When using $r_s=0.1$, the proposed method selects parents from 10\% of solutions in the current population, which restricts the diversity of the parents and reduces the search capability.

On the other hand, the results with $r_s\ge 0.3$ show no significant difference in all problems and both evaluation times.
These results indicate that a small selection ratio should be avoided, but larger selection ratios do not essentially affect the search capability.

\subsection{Effect of evaluation time bias}
\label{sec:exp2_effect}
Figures~\ref{fig:delta_igdx_nobias_exp2} and \ref{fig:delta_igdx_bias1_exp2} show the boxplot of the $\Delta IGDX$ value for \textbf{No-bias} and \textbf{Bias}, respectively, and the bottom table summarizes the median and IQR values.
The horizontal axis shows the selection ratio $r_s$, while the vertical axis shows the $\Delta IGDX$ value.
Like figures~\ref{fig:igd_bp_nobias_exp2} and \ref{fig:igd_bp_bias1_exp2}, the two boxes with a significant difference are connected with the ``*'' or ``**'' symbols.

First, the results with \textbf{No-bias} show that there is no significant difference between different selection ratios.
On the other hand, Fig.~\ref{fig:delta_igdx_bias1_exp2} shows that the $\Delta IGDX$ value decreases as the selection ratio $r_s$ increases on \textbf{Bias}. Specifically, the selection ratio of $r_s=0.9$ is significantly biased toward the search region with a shorter evaluation time (a negative $\Delta IGDX$ value). 
Since a large selection ratio selects parent individuals from a large candidate pool containing solutions with a large search frequency, it gets close to the standard asynchronous method. Therefore, the larger the selection ratio $r_s$, the more susceptible the evaluation time bias.
These results indicate that a large selection ratio should be avoided to reduce the effect of the evaluation time bias.

\begin{figure*}[tb]
\begin{tabular}{cccccc}
\begin{minipage}[b]{0.14\textwidth}
\centering
\subfloat[MMF2]{\includegraphics[scale=0.3]{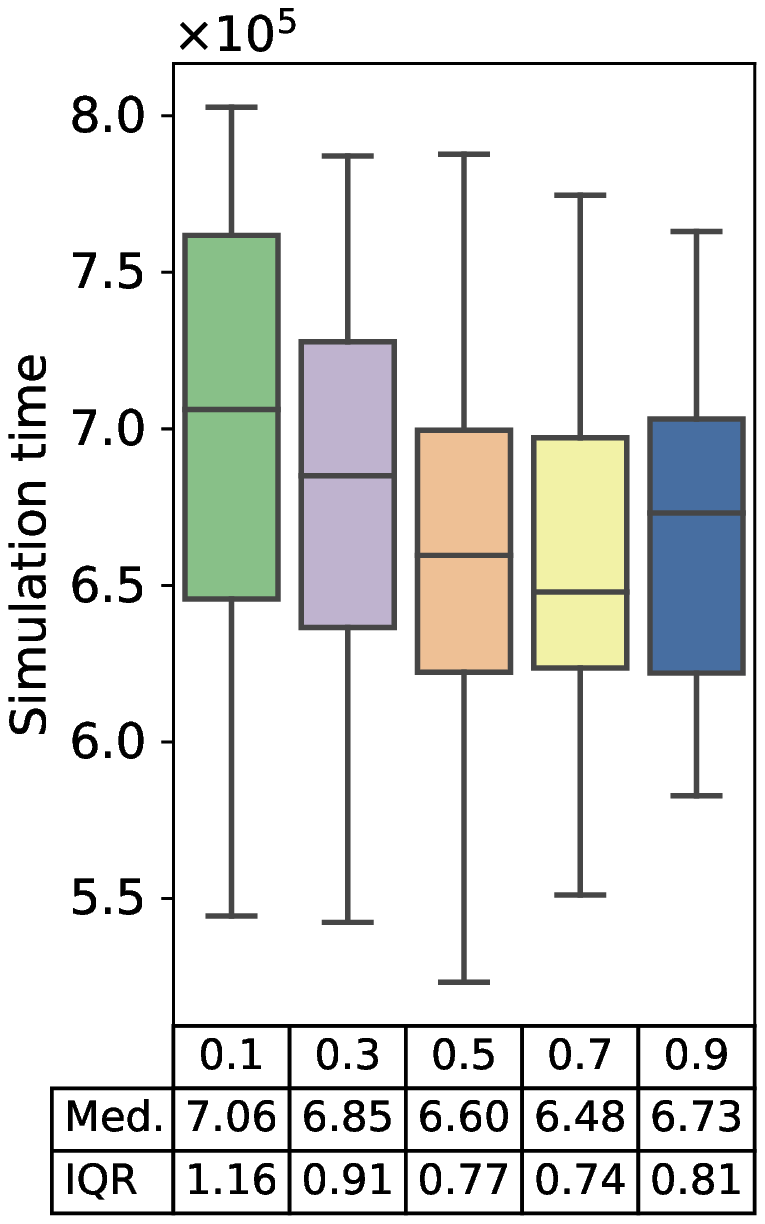}}
\end{minipage}
&
\begin{minipage}[b]{0.14\textwidth}
\centering
\subfloat[MMF3]{\includegraphics[scale=0.3]{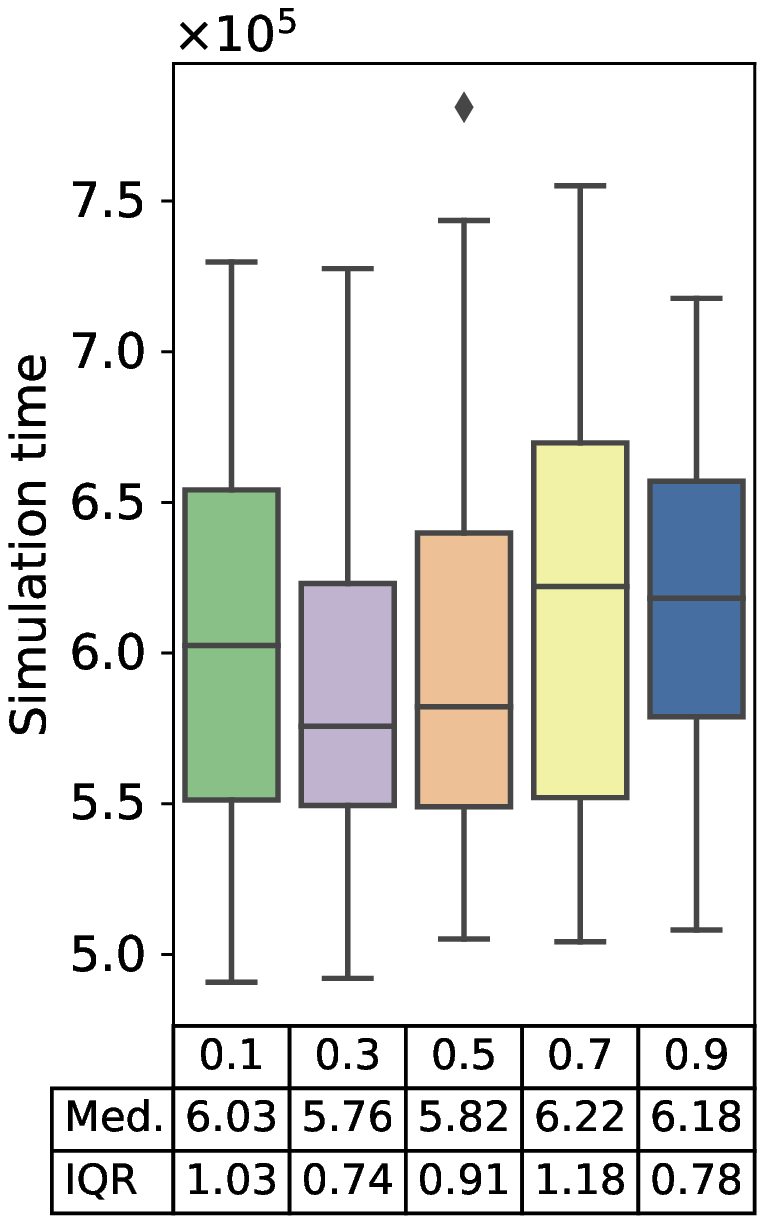}}
\end{minipage}
&
\begin{minipage}[b]{0.14\textwidth}
\centering
\subfloat[MMF4]{\includegraphics[scale=0.3]{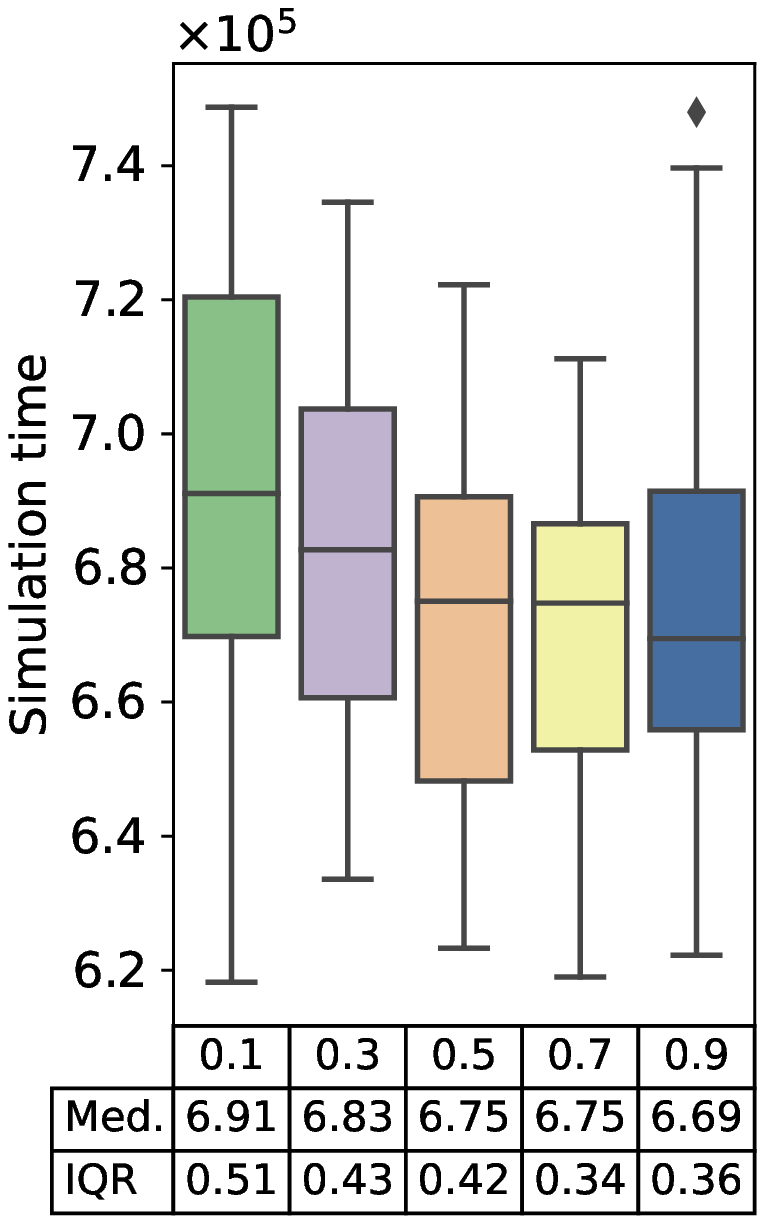}}
\end{minipage}
&
\begin{minipage}[b]{0.14\textwidth}
\centering
\subfloat[MMF5]{\includegraphics[scale=0.3]{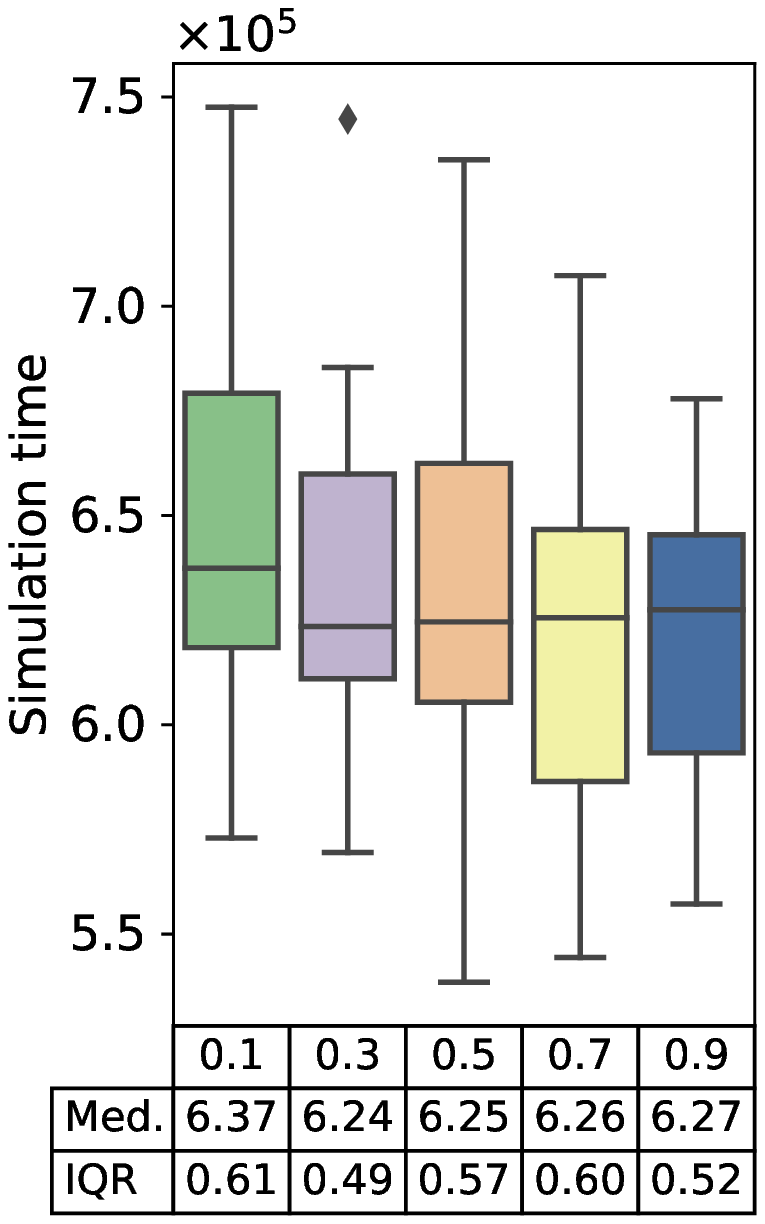}}
\end{minipage}
&
\begin{minipage}[b]{0.14\textwidth}
\centering
\subfloat[MMF6]{\includegraphics[scale=0.3]{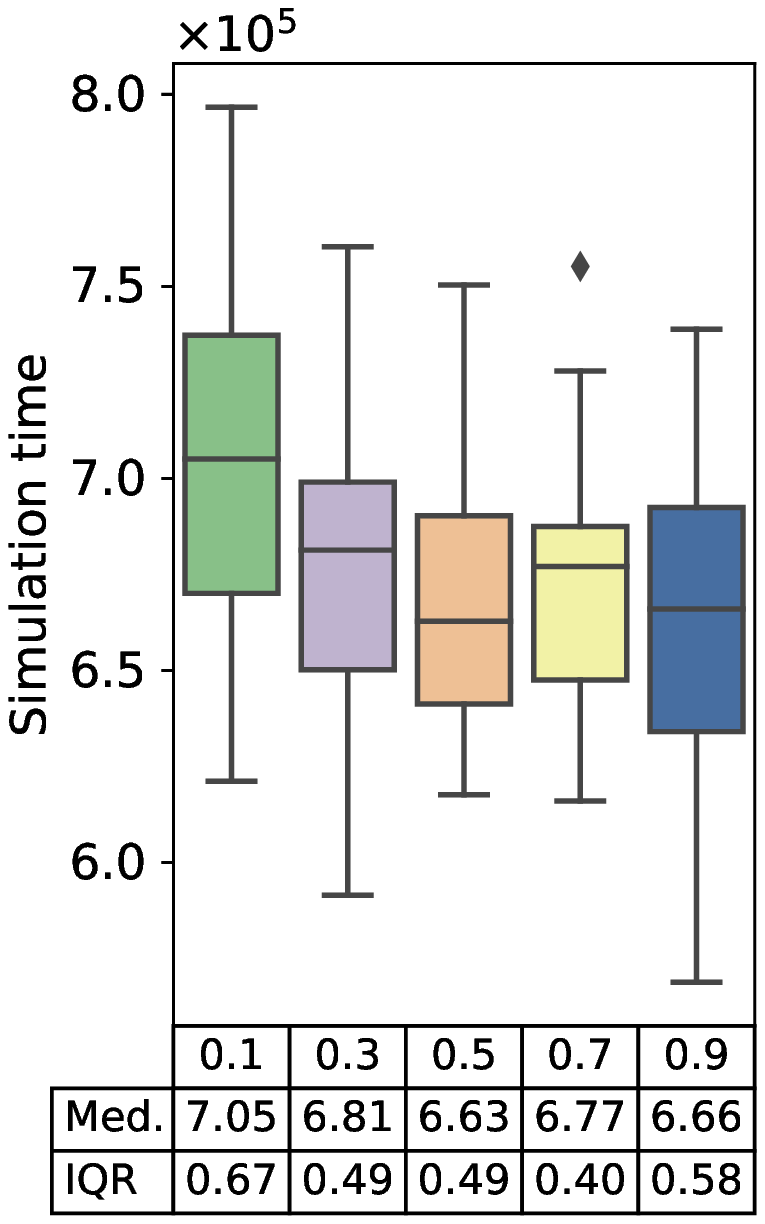}}
\end{minipage}
&
\begin{minipage}[b]{0.14\textwidth}
\centering
\subfloat[MMF8]{\includegraphics[scale=0.3]{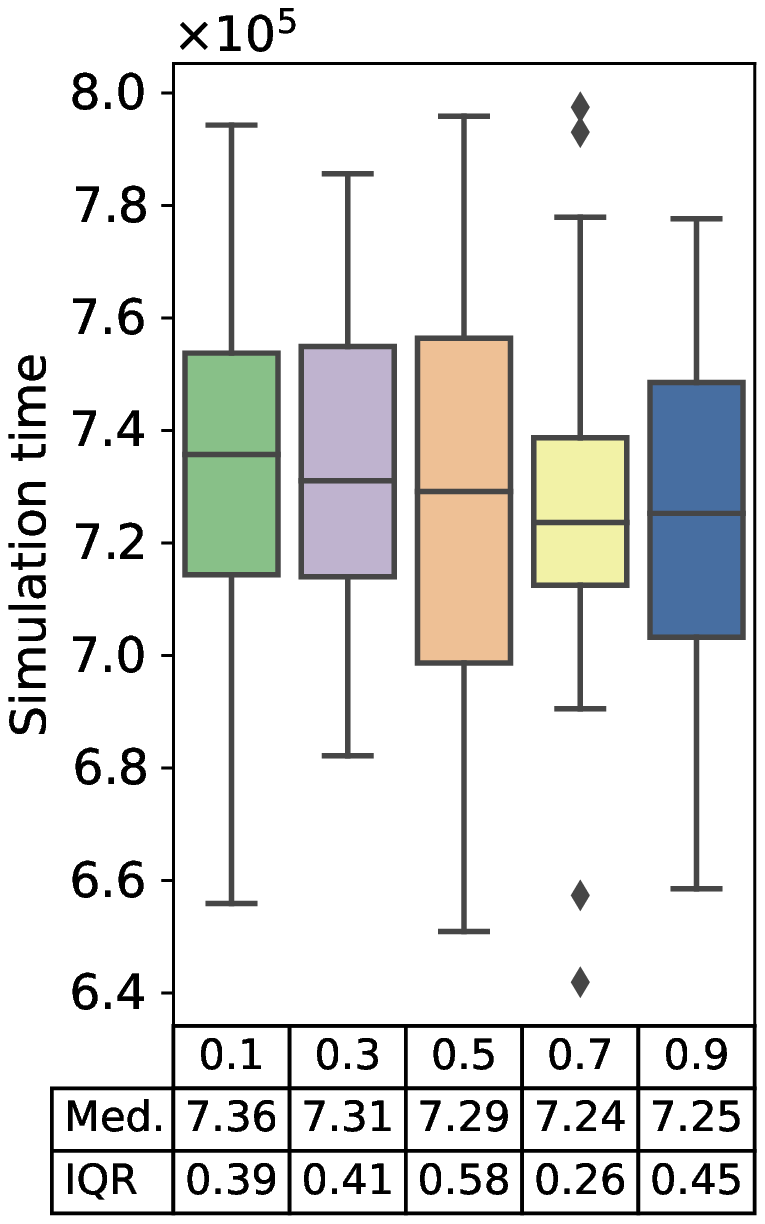}}
\end{minipage}
\end{tabular}
\caption{The simulation time until reaching a particular $IGD$ value with \textbf{No-bias} (different $r_s$)}
\label{fig:time_bp_nobias_exp2}
\begin{tabular}{cccccc}
\begin{minipage}[b]{0.14\textwidth}
\centering
\subfloat[MMF2]{\includegraphics[scale=0.3]{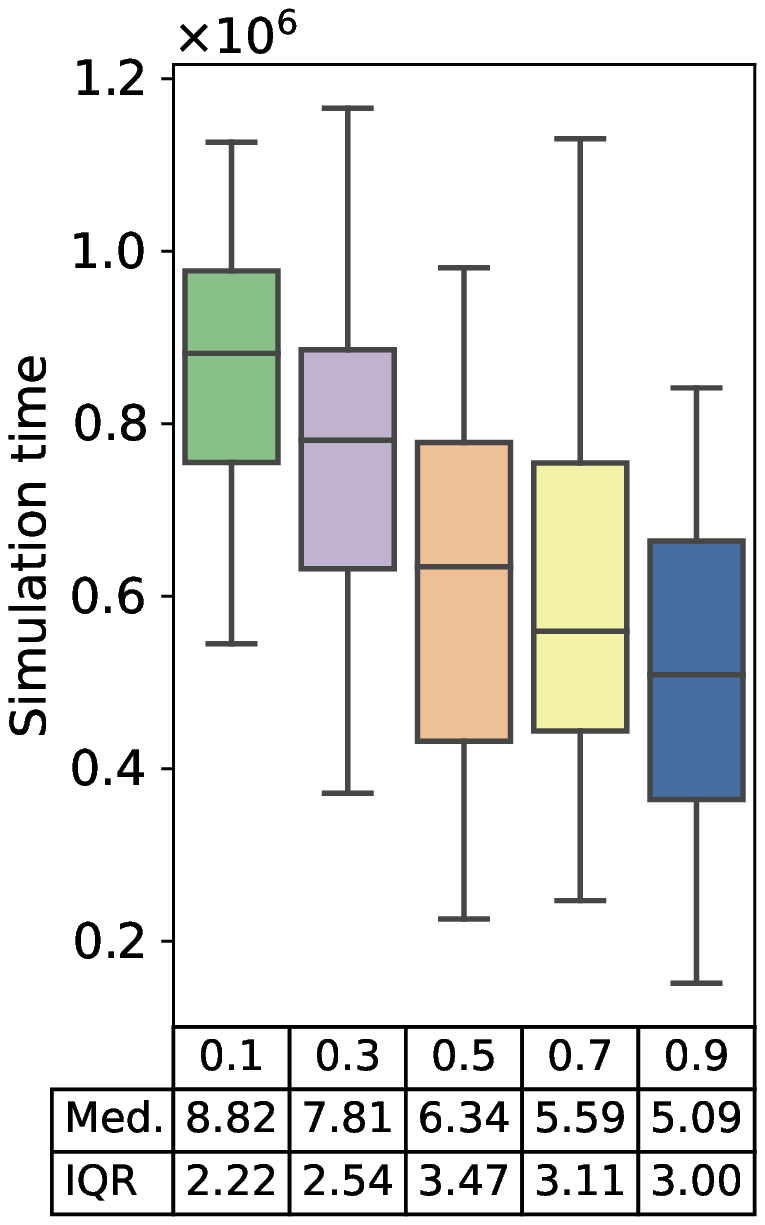}}
\end{minipage}
&
\begin{minipage}[b]{0.14\textwidth}
\centering
\subfloat[MMF3]{\includegraphics[scale=0.3]{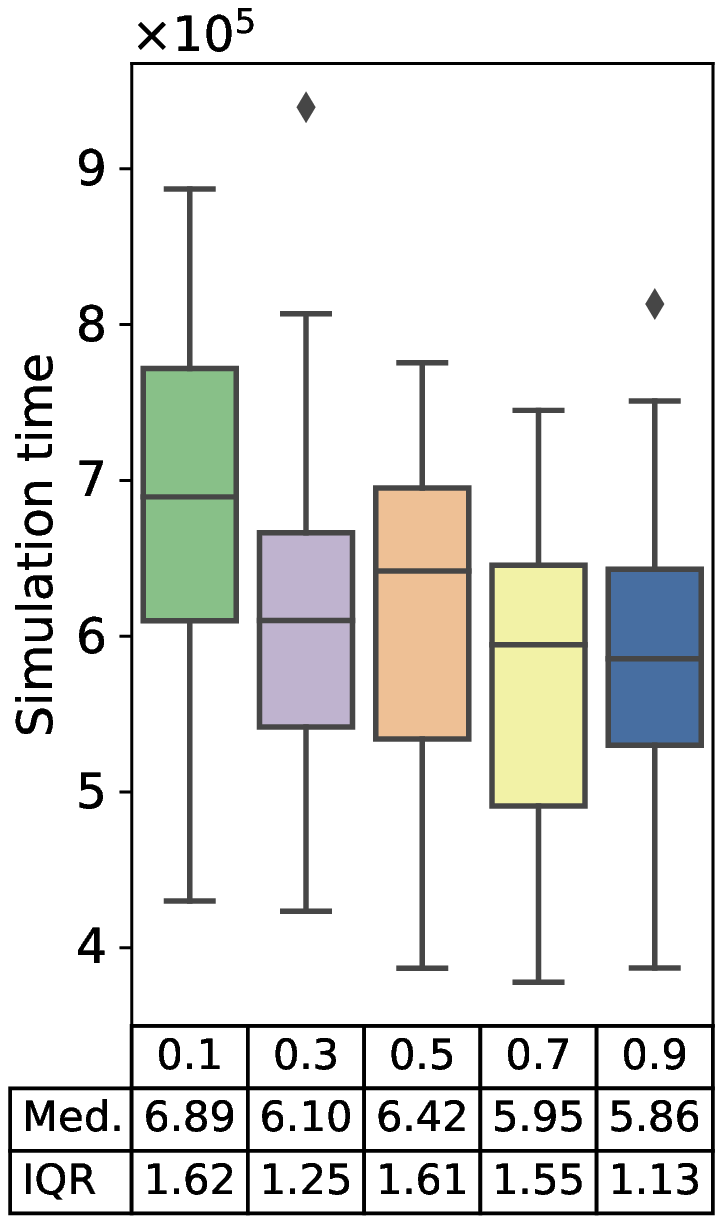}}
\end{minipage}
&
\begin{minipage}[b]{0.14\textwidth}
\centering
\subfloat[MMF4]{\includegraphics[scale=0.3]{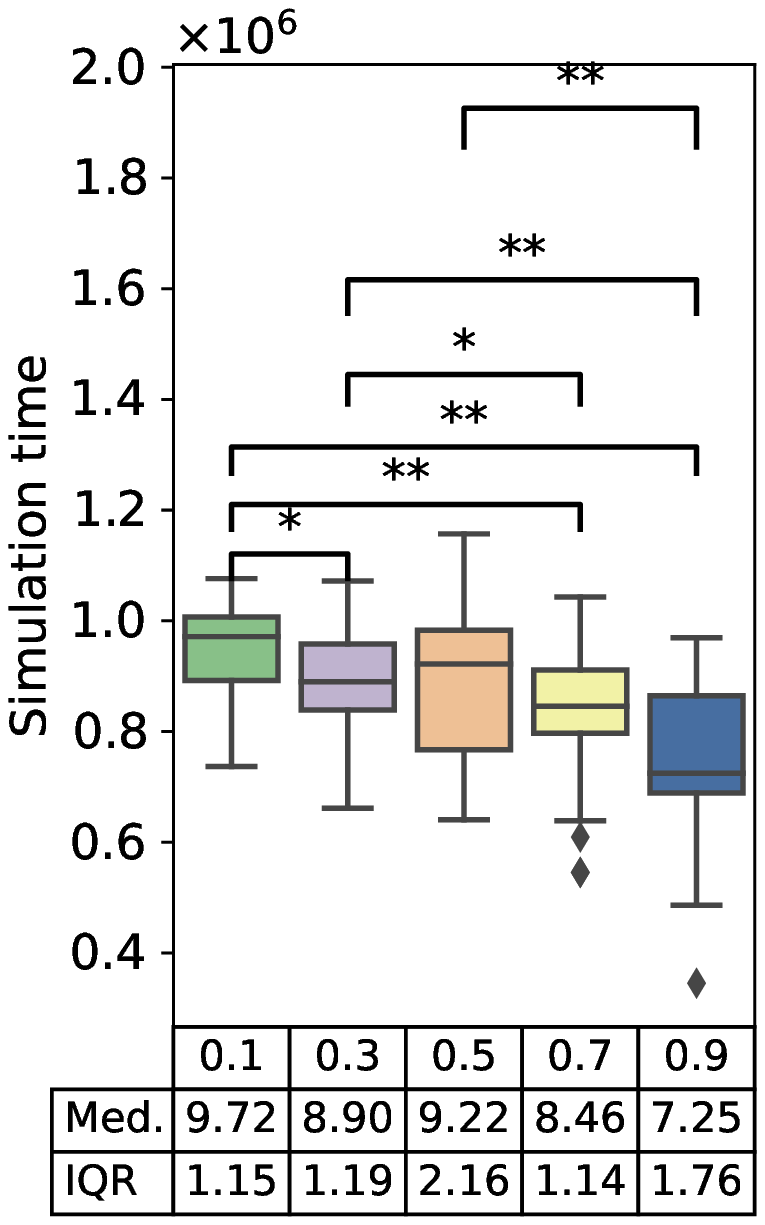}}
\end{minipage}
&
\begin{minipage}[b]{0.14\textwidth}
\centering
\subfloat[MMF5]{\includegraphics[scale=0.3]{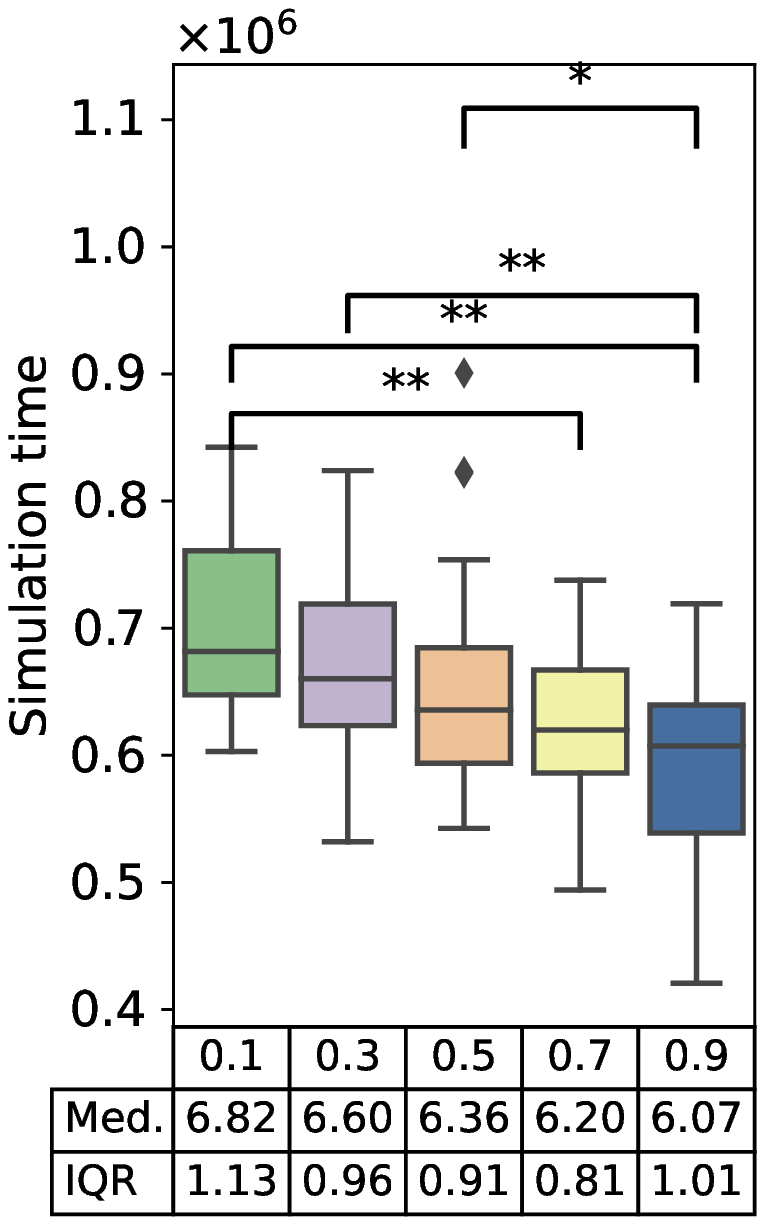}}
\end{minipage}
&
\begin{minipage}[b]{0.14\textwidth}
\centering
\subfloat[MMF6]{\includegraphics[scale=0.3]{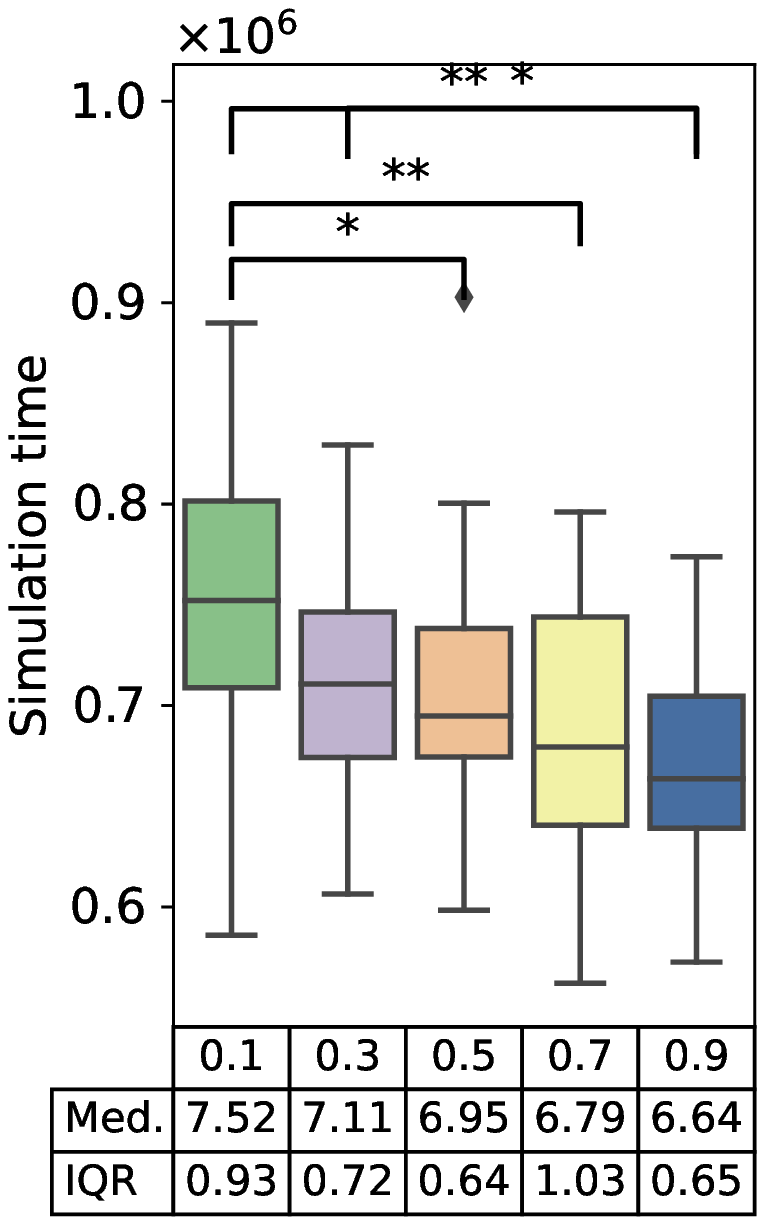}}
\end{minipage}
&
\begin{minipage}[b]{0.14\textwidth}
\centering
\subfloat[MMF8]{\includegraphics[scale=0.3]{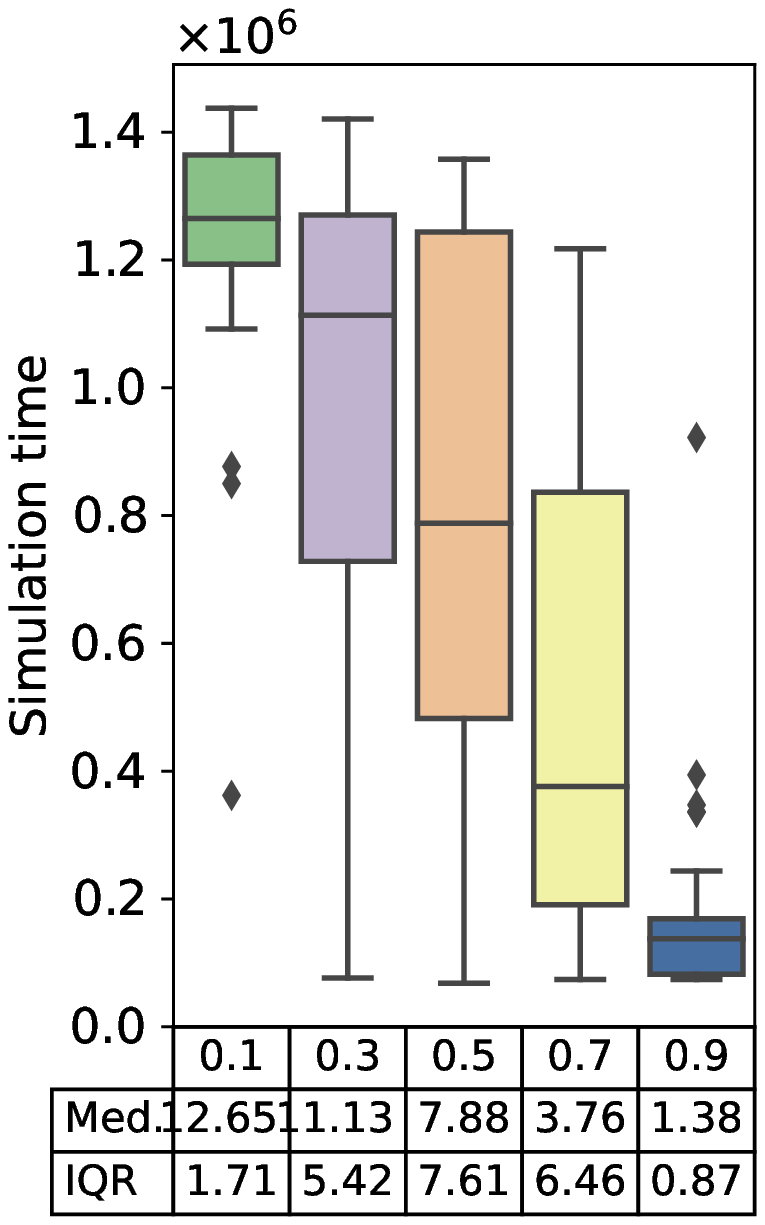}}
\end{minipage}
\end{tabular}
\caption{The simulation time until reaching a particular $IGD$ value with \textbf{Bias} (different $r_s$)}
\label{fig:time_bp_bias1_exp2}
\end{figure*}
\subsection{Computational efficiency}
\label{sec:exp2_time}
This subsection analyzes the computational efficiency of different selection ratios by comparing the simulation time until obtaining a particular quality of the Pareto front.
Concretely, the target $IGD$ value is decided for each problem from the previous results as follows:
$2.0\times 10^{-3}$ in MMF2 and MMF3,
$2.0\times 10^{-4}$ in MMF4,
$4.5\times 10^{-4}$ in MMF5,
$3.5\times 10^{-4}$ in MMF6,
and $1.5\times 10^{-4}$ in MMF8.

Figures~\ref{fig:time_bp_nobias_exp2} and \ref{fig:time_bp_bias1_exp2} show the simulation time until the target $IGD$ value is reached for \textbf{No-bias} and \textbf{Bias}, and the bottom tables summarize the median and IQR values of simulation time until reaching the target $IGD$ value. The horizontal axis shows the selection ratio, while the vertical axis shows the simulation time.
As with the previous results, two boxes with a significant difference are connected with the ``*'' symbols. 

First, no significant difference is found in all test problems when using \textbf{No-bias}. However, the selection ratio $r_s=0.1$ requires a relatively longer simulation time than the others in all problems. For the other selection ratios of $r_s\ge 0.3$, all selection ratios take almost equal simulation time in \textbf{No-bias}. These results can be explained because a small selection ratio of $r_s=0.1$ shows the lower search capability, as demonstrated in Section~\ref{sec:exp2_capability}.

From the results with \textbf{Bias}, on the other hand, the larger the selection ratio is used, the shorter the simulation time is obtained.
Here, it is necessary to consider the effect of the evaluation time bias on the execution time. Specifically, when using the selection ratios of $r_s=0.9$, the search direction is biased toward a region with a shorter evaluation time, as demonstrated in Section~\ref{sec:exp2_effect}, resulting in a shorter execution time. In fact, it can be seen that the evaluation time of the selection ratios of $r_s=0.9$ is the shortest in all problems, and some are significantly shorter than others.
For the other selection ratios, the selection ratio of $r_s=0.1$ obtains the significantly longest execution time in MMF4, MMF5, and MMF6. 
This is due to a combination of the fact that the small selection ratio decreases its search capability (shown in Section \ref{sec:exp2_capability}), and it obtains solutions with both longer and shorter evaluation times by reducing the effect of the evaluation time bias (shown in Section \ref{sec:exp2_effect}).

In contrast, the selection ratios of $r_s=0.5, 0.7$ acquire stably shorter execution times in both evaluation times.
These ratios acquire comparable search capability to the large selection ratios (e.g., $r_s=0.9$) and decrease the influence of the evaluation time bias by using a smaller ratio (e.g., $r_s=0.1$). For this fact, it is indicated that such moderate selection ratios are appropriate.

\section{Comparison of Different Parallelization Schemes}\label{sec:result}
This section compares the performance of different parallelization schemes, SP-NSGA-III, AP-NSGA-III, and FS-NSGA-III.
Since the results in the previous section suggested the selection ratio such as $r_s=0.5, 0.7$, the following experiments use $r_s=0.5$ in the proposed method.

\subsection{Search capability}
\label{sec:exp1_capability}
\begin{figure*}[tb]
\begin{tabular}{cccccc}
\begin{minipage}[b]{0.14\textwidth}
\centering
\subfloat[MMF2]{\includegraphics[scale=0.3]{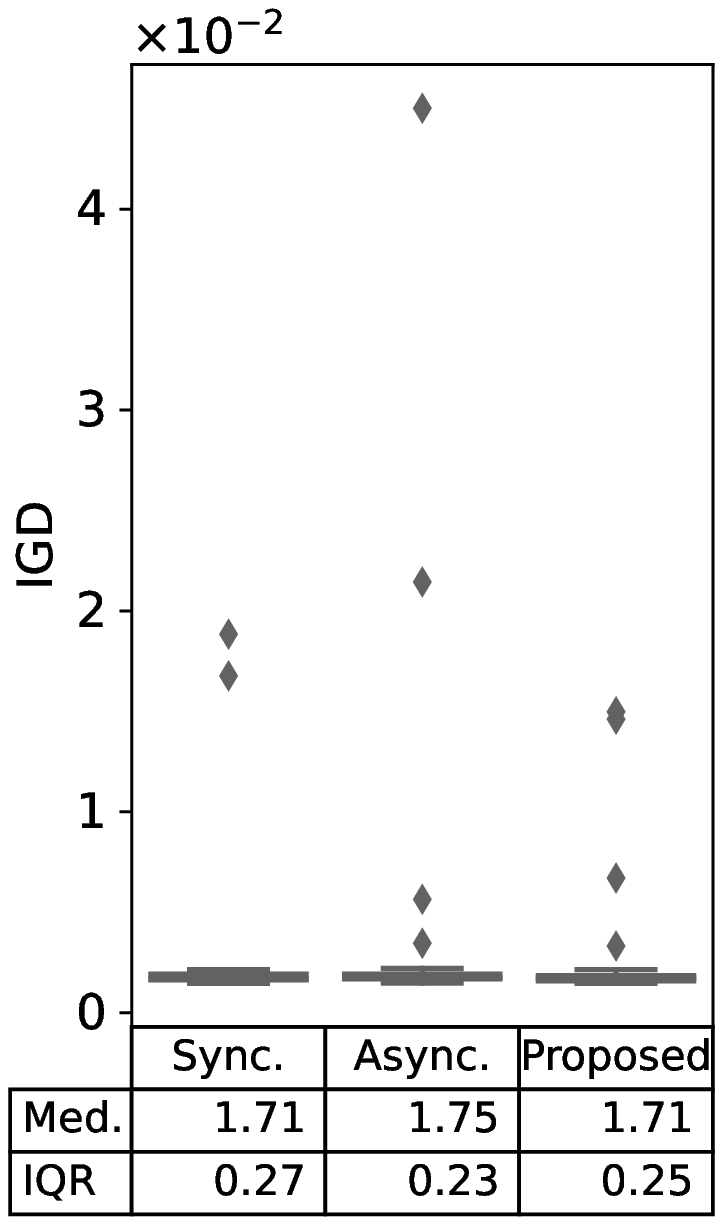}}
\end{minipage}
&
\begin{minipage}[b]{0.14\textwidth}
\centering
\subfloat[MMF3]{\includegraphics[scale=0.3]{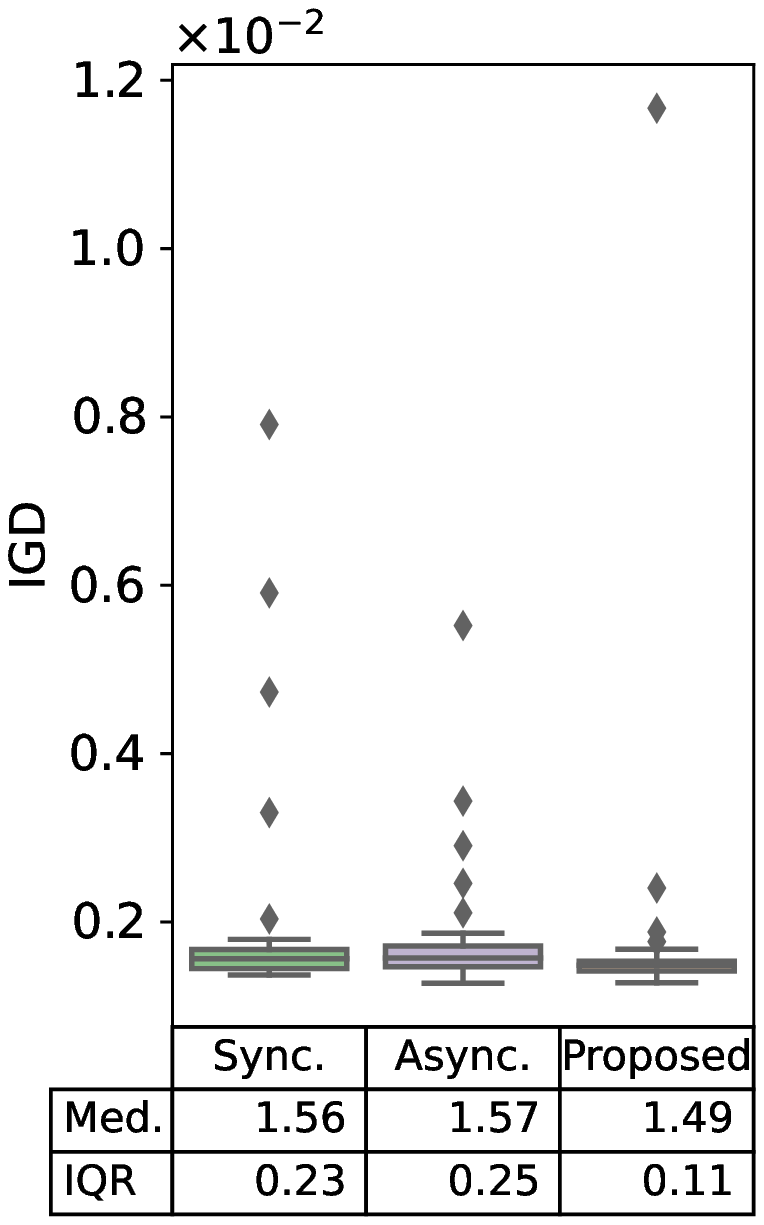}}
\end{minipage}
&
\begin{minipage}[b]{0.14\textwidth}
\centering
\subfloat[MMF4]{\includegraphics[scale=0.3]{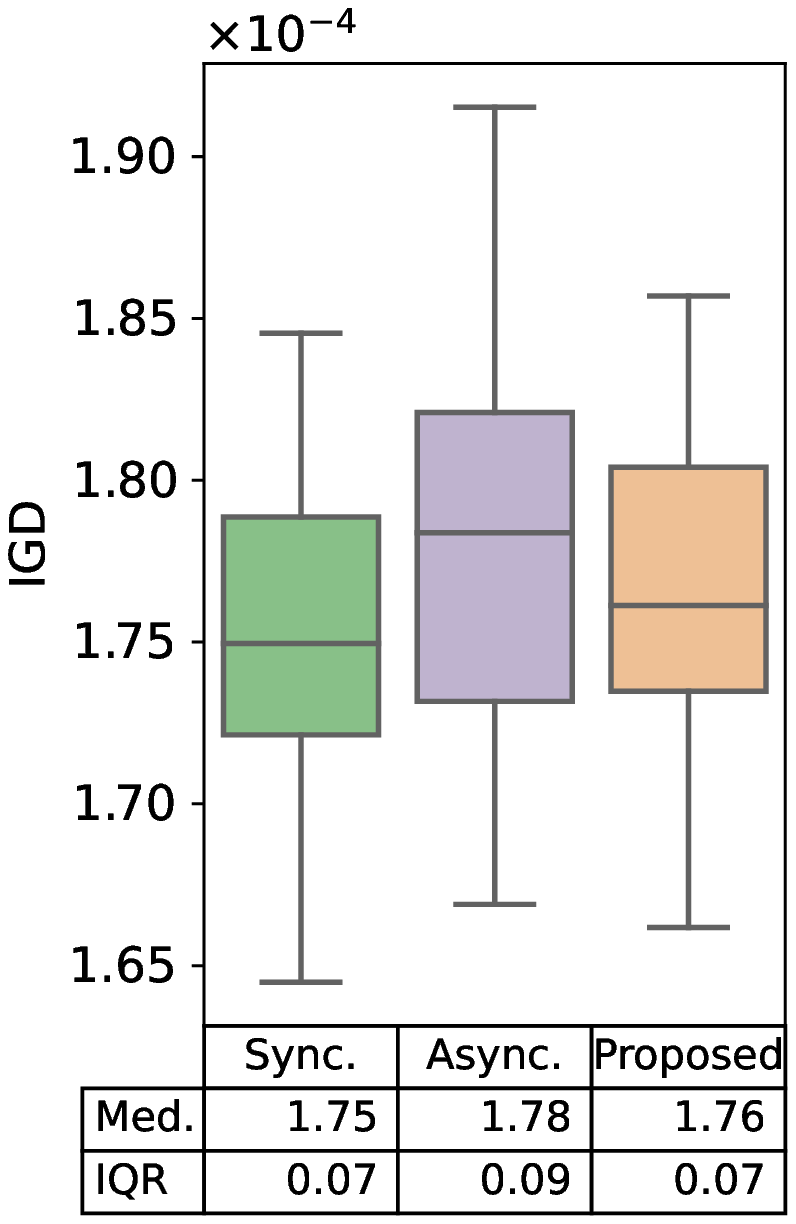}}
\end{minipage}
&
\begin{minipage}[b]{0.14\textwidth}
\centering
\subfloat[MMF5]{\includegraphics[scale=0.3]{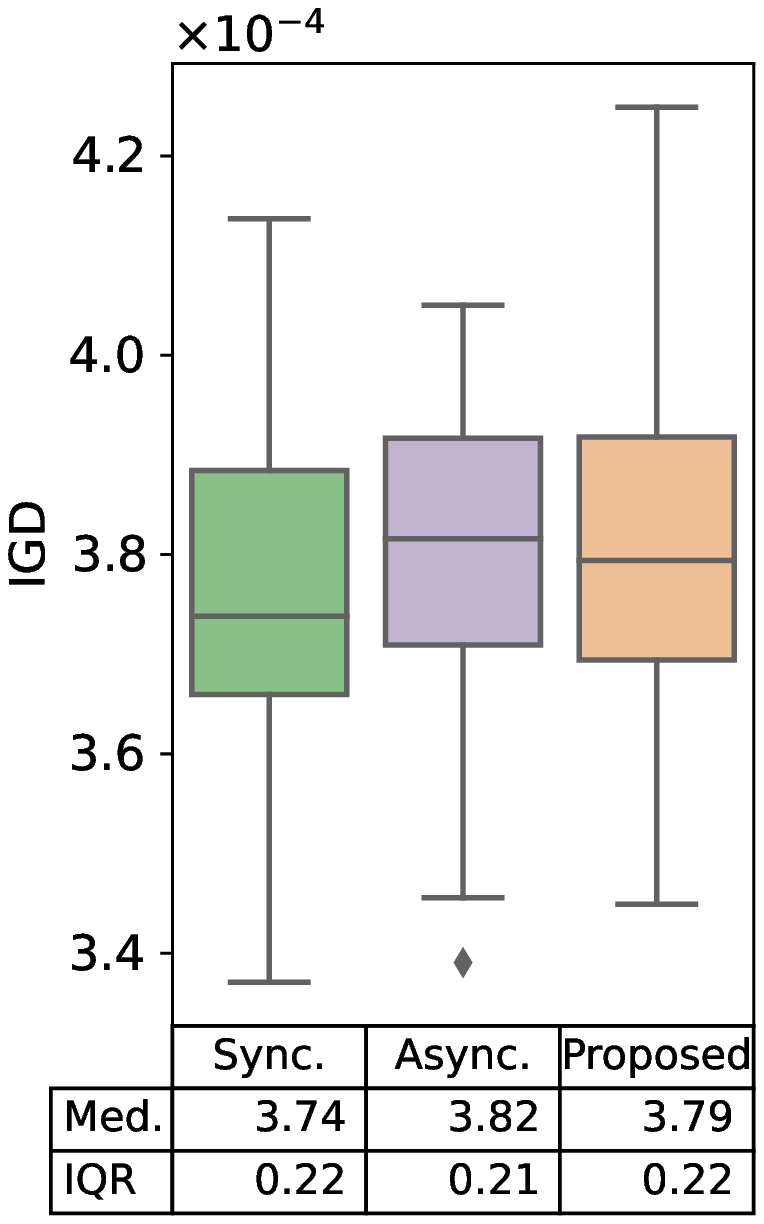}}
\end{minipage}
&
\begin{minipage}[b]{0.14\textwidth}
\centering
\subfloat[MMF6]{\includegraphics[scale=0.3]{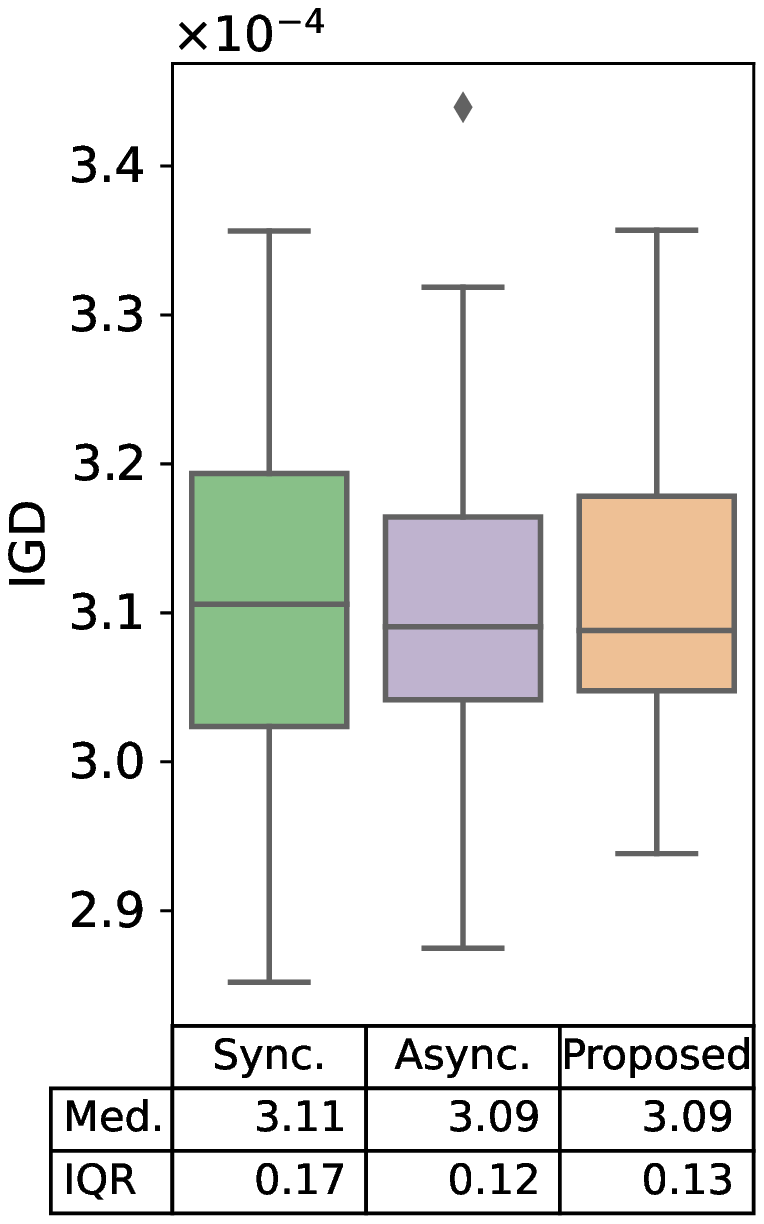}}
\end{minipage}
&
\begin{minipage}[b]{0.14\textwidth}
\centering
\subfloat[MMF8]{\includegraphics[scale=0.3]{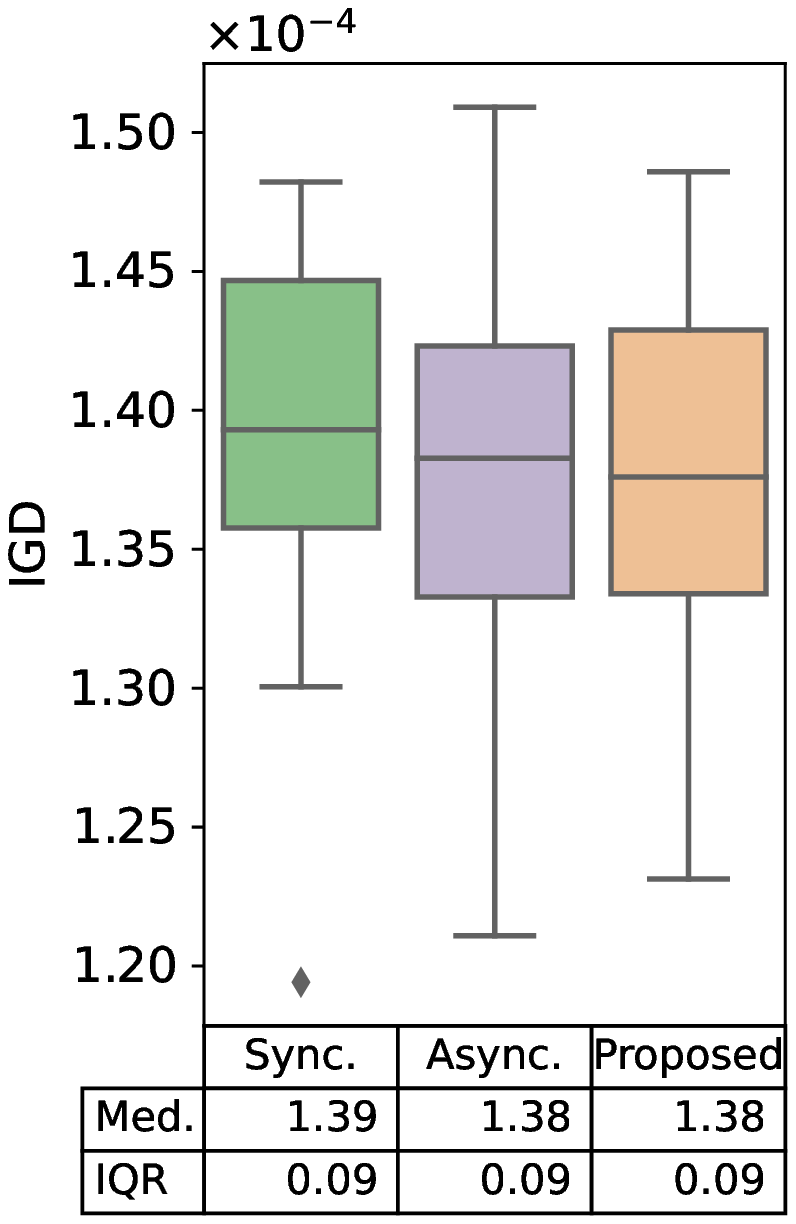}}
\end{minipage}
\end{tabular}
\caption{$IGD$ with \textbf{No-bias} after the maximum fitness evaluations (different parallelization schemes)}
\label{fig:igd_bp_nobias}
\begin{tabular}{cccccc}
\begin{minipage}[b]{0.14\textwidth}
\centering
\subfloat[MMF2]{\includegraphics[scale=0.3]{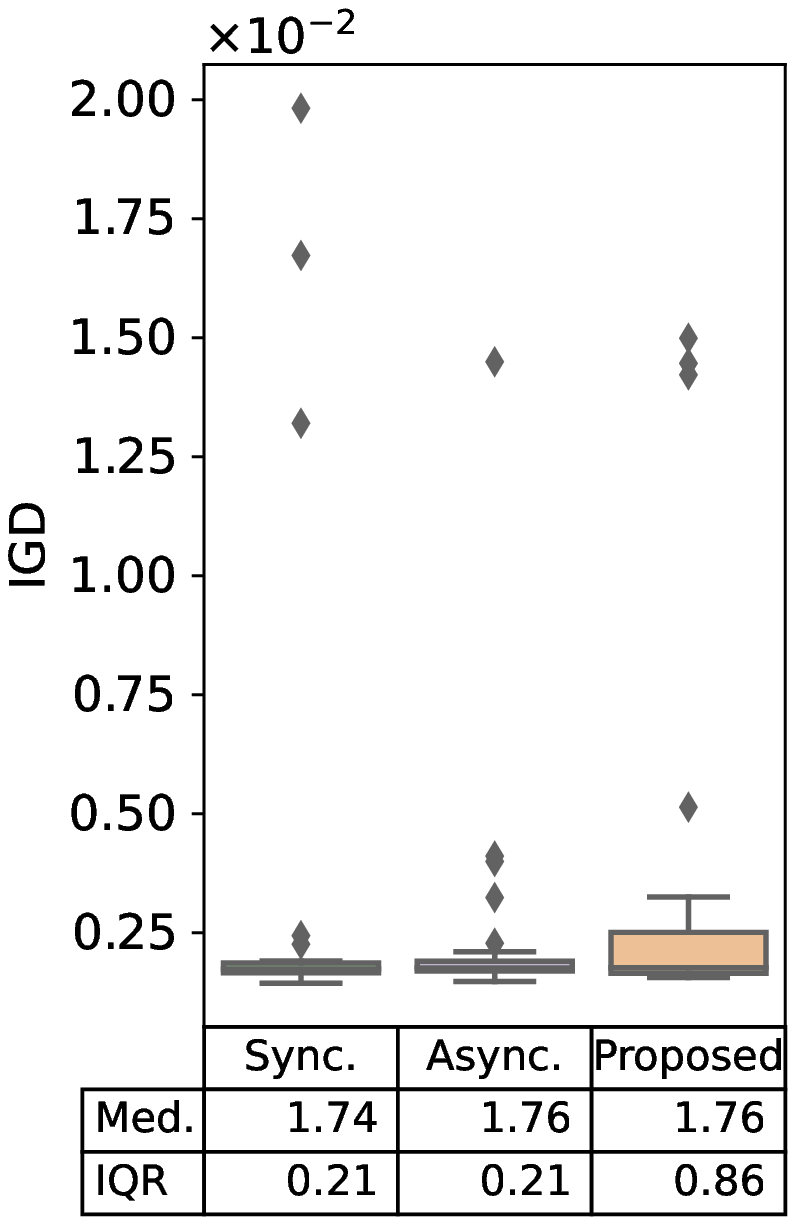}}
\end{minipage}
&
\begin{minipage}[b]{0.14\textwidth}
\centering
\subfloat[MMF3]{\includegraphics[scale=0.3]{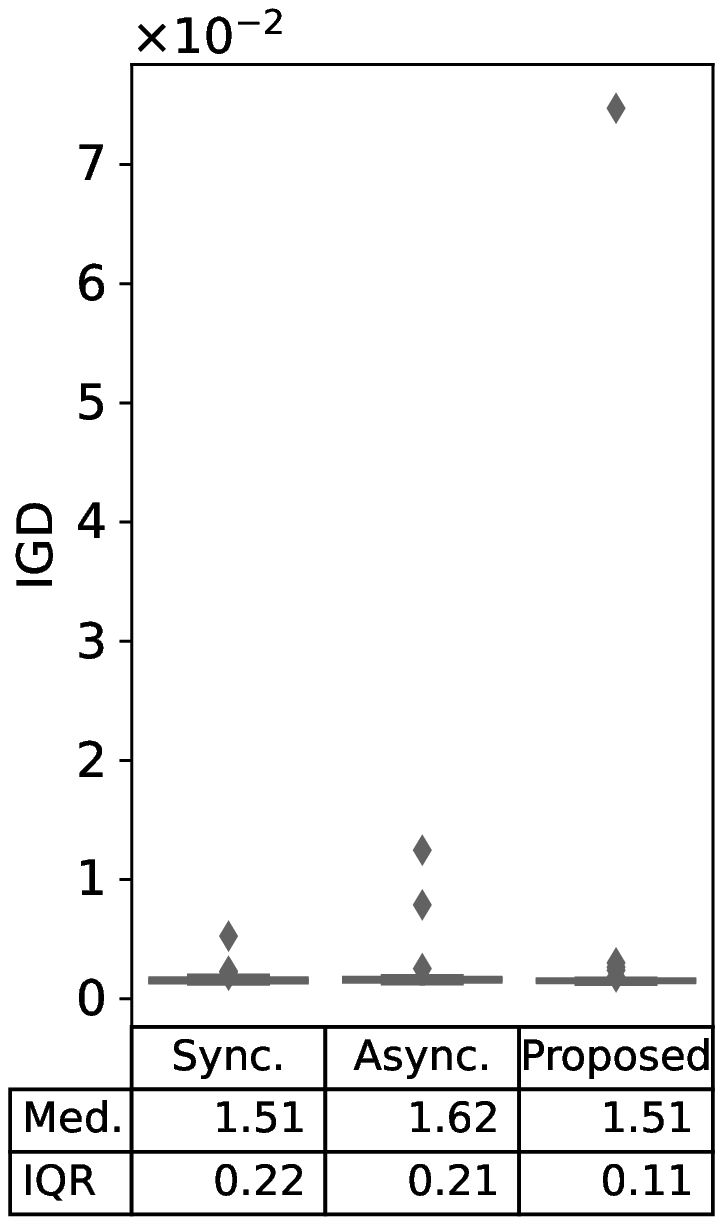}}
\end{minipage}
&
\begin{minipage}[b]{0.14\textwidth}
\centering
\subfloat[MMF4]{\includegraphics[scale=0.3]{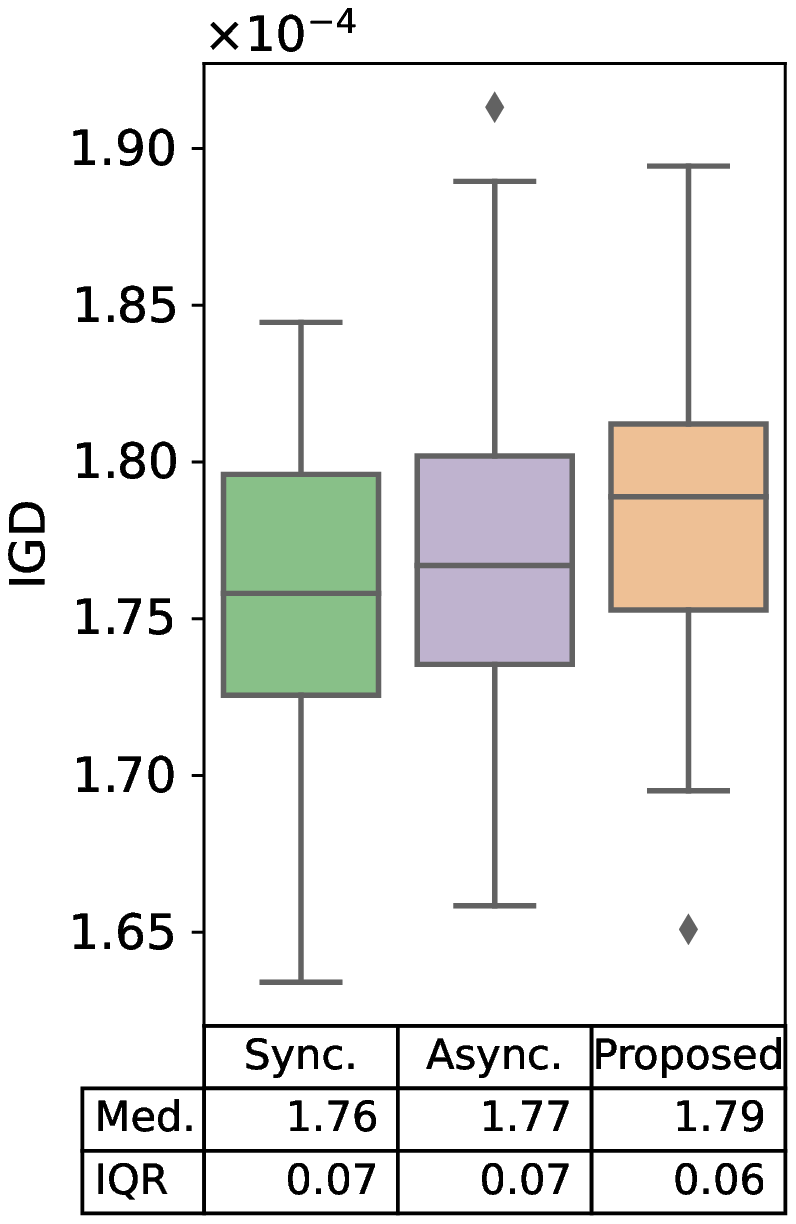}}
\end{minipage}
&
\begin{minipage}[b]{0.14\textwidth}
\centering
\subfloat[MMF5]{\includegraphics[scale=0.3]{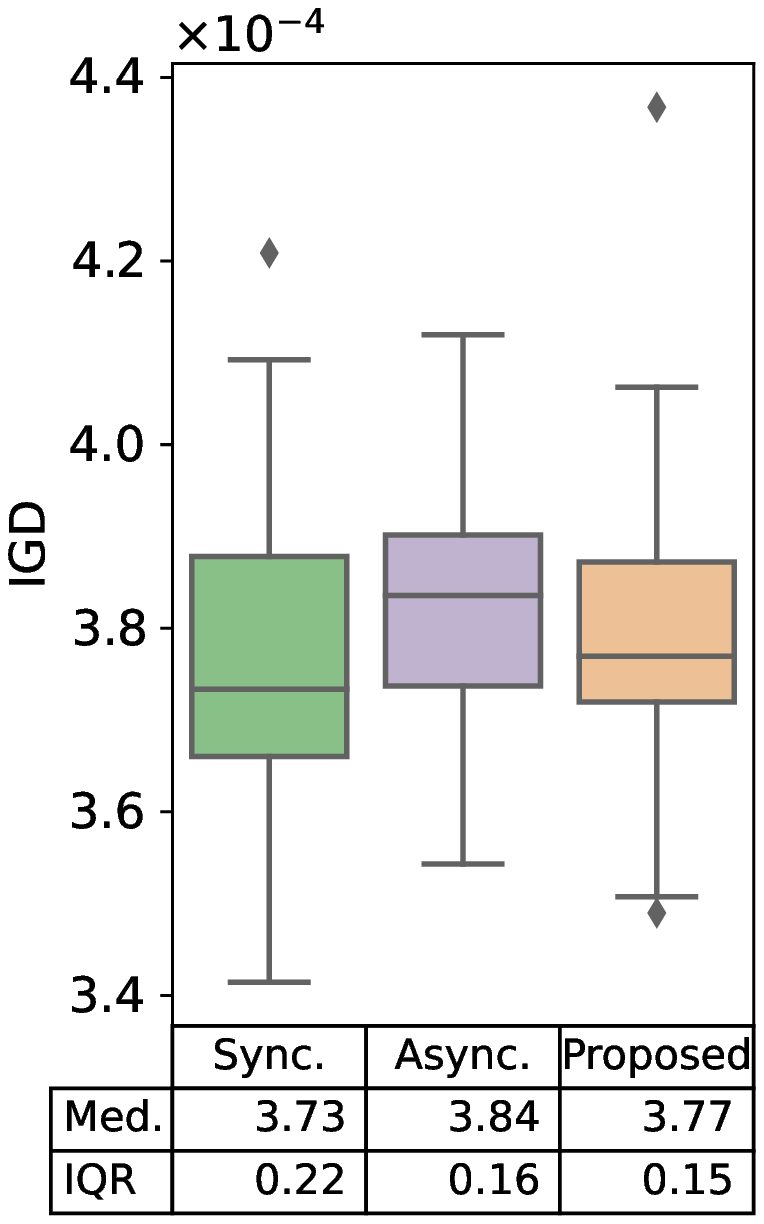}}
\end{minipage}
&
\begin{minipage}[b]{0.14\textwidth}
\centering
\subfloat[MMF6]{\includegraphics[scale=0.3]{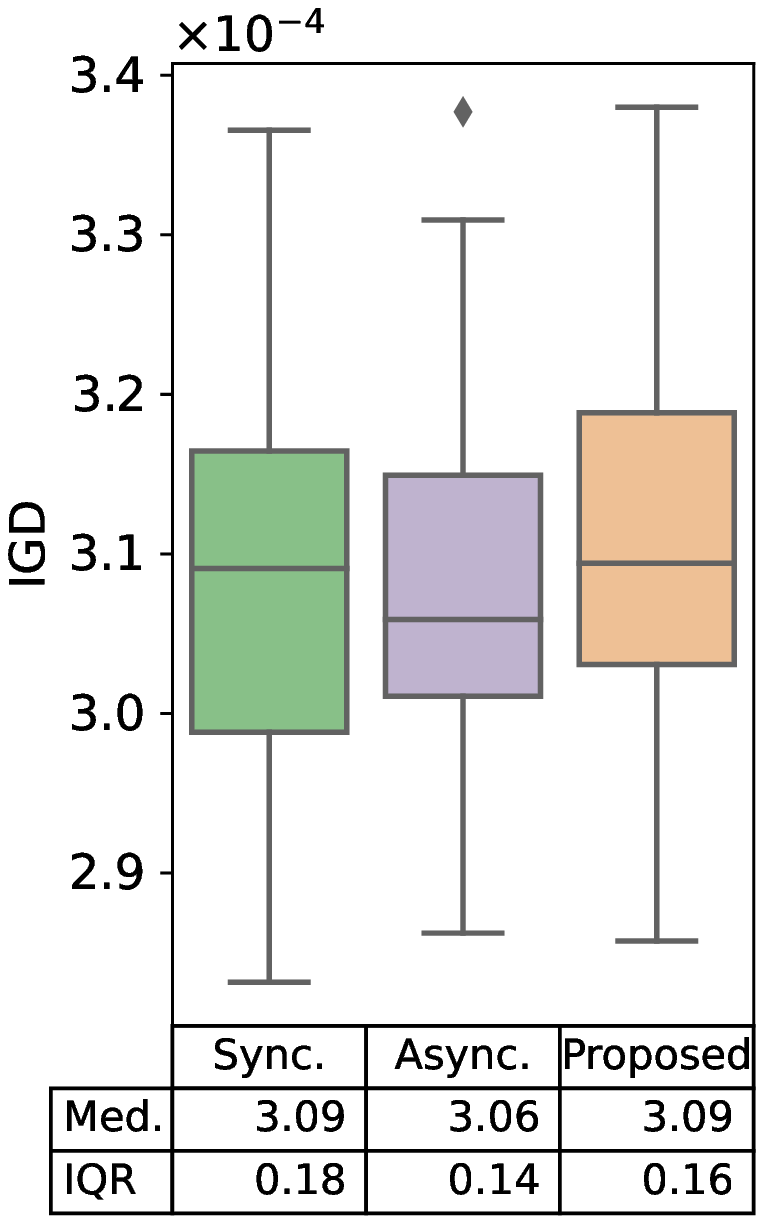}}
\end{minipage}
&
\begin{minipage}[b]{0.14\textwidth}
\centering
\subfloat[MMF8]{\includegraphics[scale=0.3]{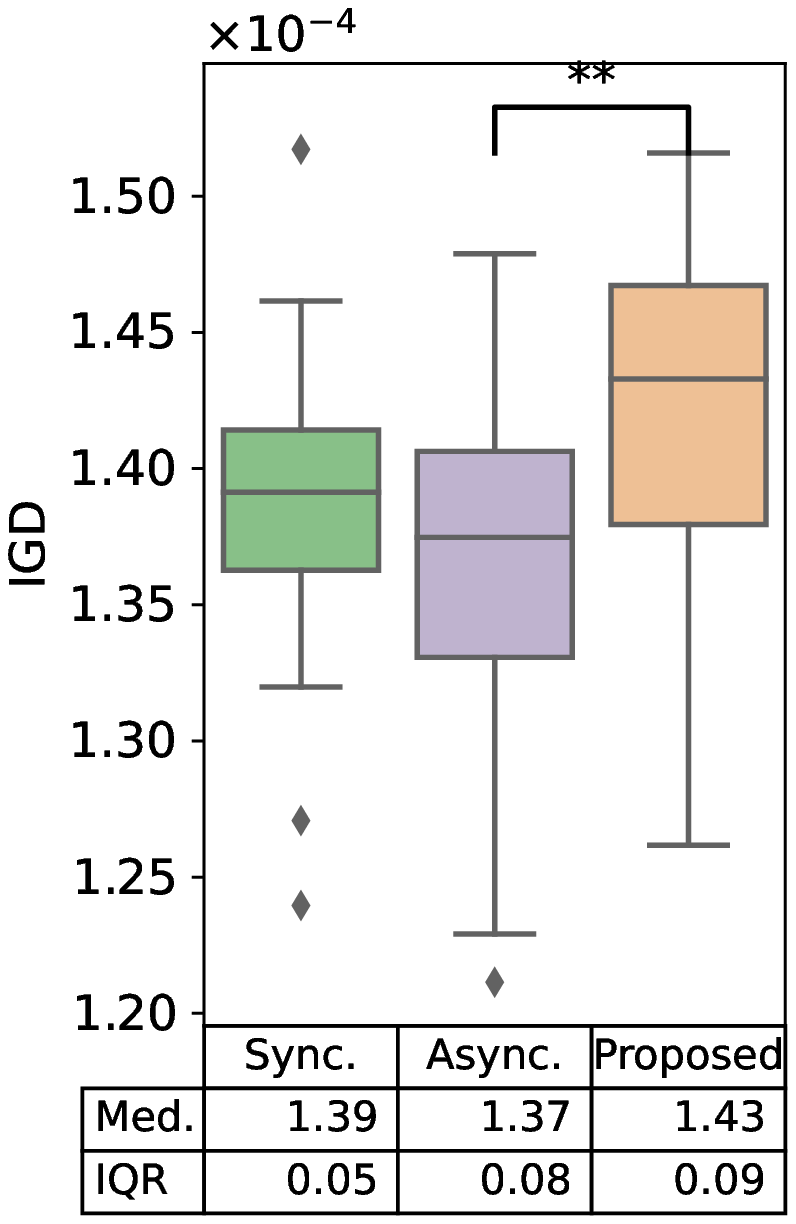}}
\end{minipage}
\end{tabular}
\caption{$IGD$ with \textbf{Bias} after the maximum fitness evaluations (different parallelization schemes)}
\label{fig:igd_bp_bias1}
\end{figure*}
Figures~\ref{fig:igd_bp_nobias} and \ref{fig:igd_bp_bias1} show the boxplot of the $IGD$ value after the maximum number of evaluations for \textbf{No-bias} and \textbf{Bias}, and the bottom tables summarize the median and IQR values.
The horizontal axis shows the different parallelization methods, while the vertical axis shows the $IGD$ value.
As in the previous section, a significant difference is depicted with ``*'' symbols.

From these results, when using \textbf{No-bias}, there is no significant difference between the three parallelization methods.
On the other hand, from Fig.~\ref{fig:igd_bp_bias1}, no significant difference is found except for MMF8 when using \textbf{Bias}.
FS-NSGA-III produces a comparable $IGD$ value in other problems compared with other methods. 
In MMF8, FS-NSGA-III obtains a significantly larger (worse) $IGD$ value than AP-NSGA-III.
This result can be explained in Fig.~\ref{fig:delta_igdx_bias1}.
Since AP-NSGA-III is biased toward the search region with a short evaluation time, it precisely obtains the Pareto front by only solutions in PS1.
In contrast, since FS-NSGA-III and SP-NSGA-III obtain both Pareto sets equally, the $IGD$ values are inferior to those of AP-NSGA-III that approximates the Pareto front elaborated by PS1 only.

This result indicates that FS-NSGA-III does not negatively affect the search capability of AP-NSGA-III, even though selecting the parents from the limited candidate pool.

\subsection{Effect of evaluation time bias}
\label{sec:exp1_effect}
\begin{figure*}[!tb]
\begin{tabular}{cccccc}
\begin{minipage}[b]{0.14\textwidth}
\subfloat[MMF2]{\includegraphics[scale=0.3]{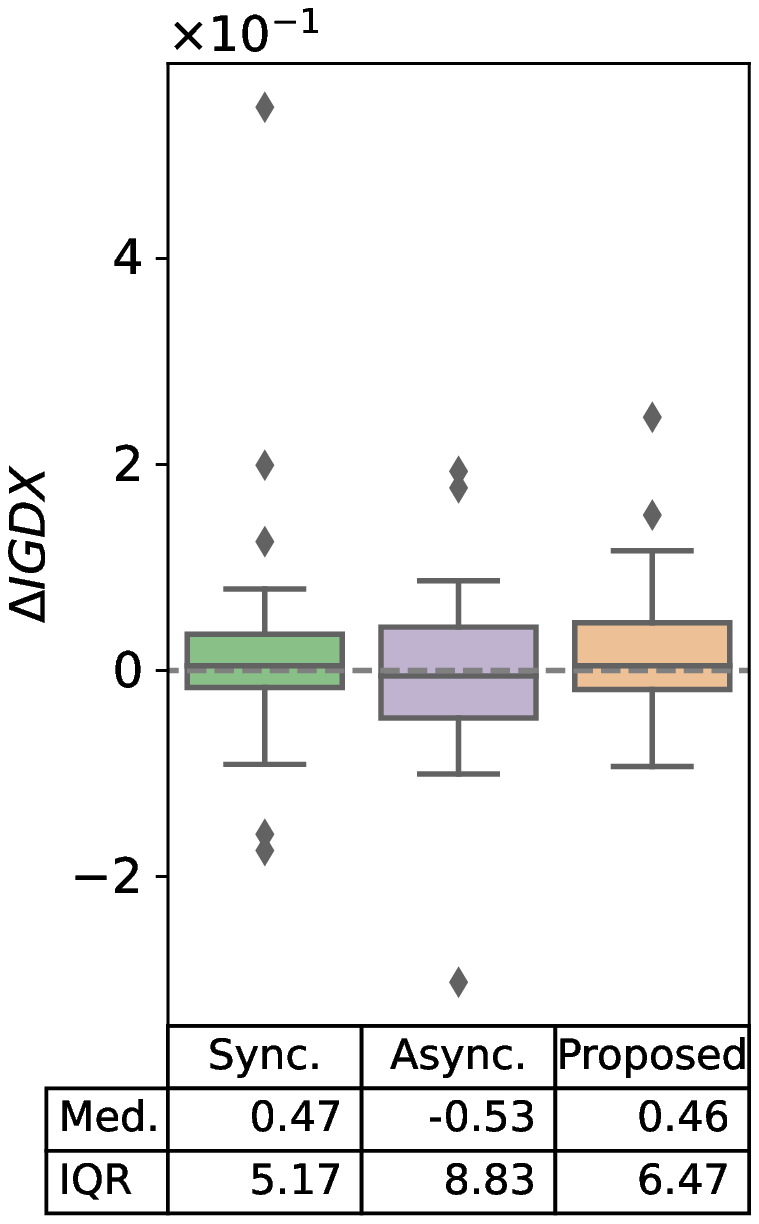}}
\end{minipage}
&
\begin{minipage}[b]{0.14\textwidth}
\subfloat[MMF3]{\includegraphics[scale=0.3]{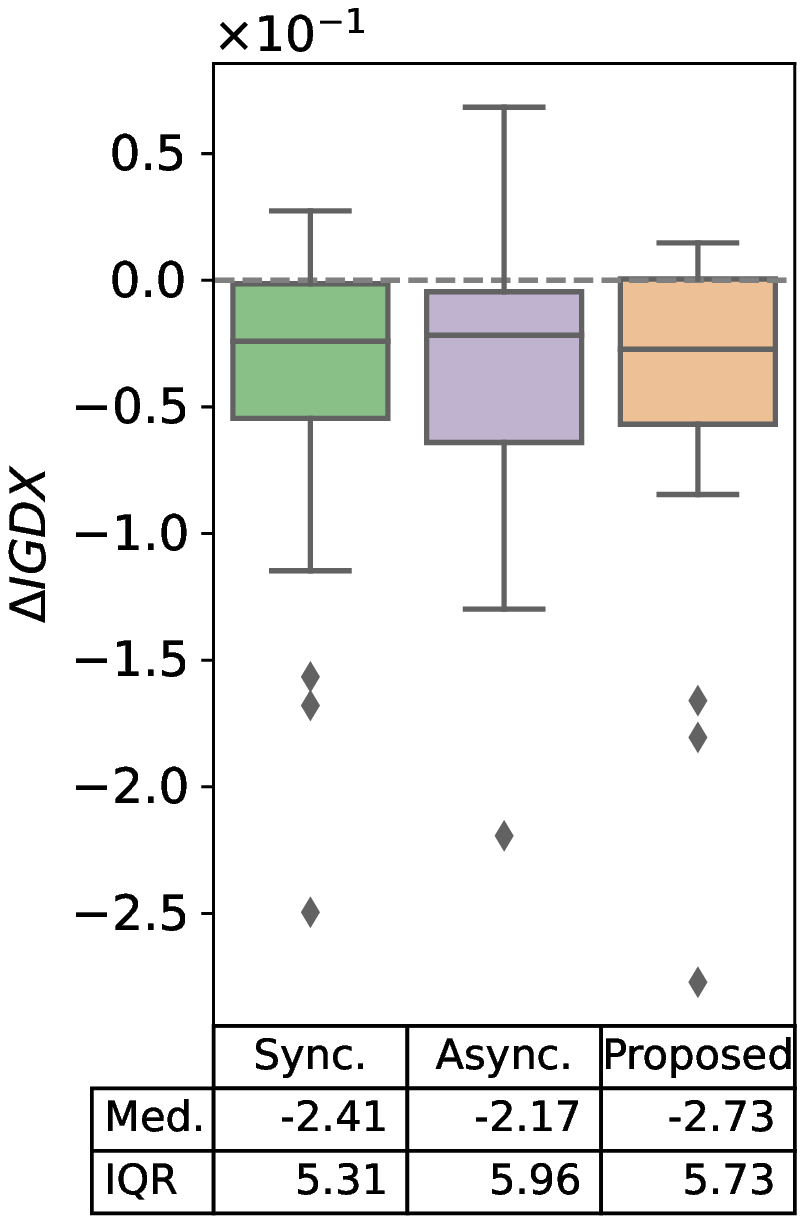}}
\end{minipage}
&
\begin{minipage}[b]{0.14\textwidth}
\subfloat[MMF4]{\includegraphics[scale=0.3]{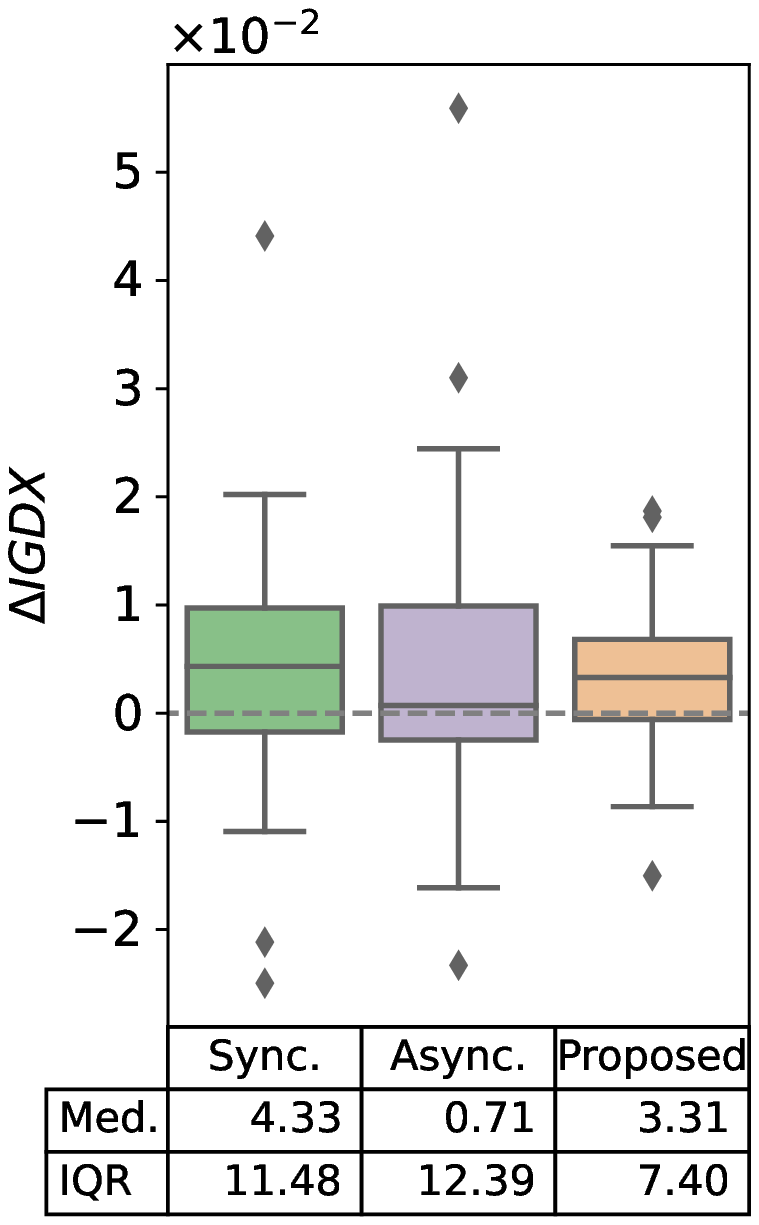}}
\end{minipage}
&
\begin{minipage}[b]{0.14\textwidth}
\subfloat[MMF5]{\includegraphics[scale=0.3]{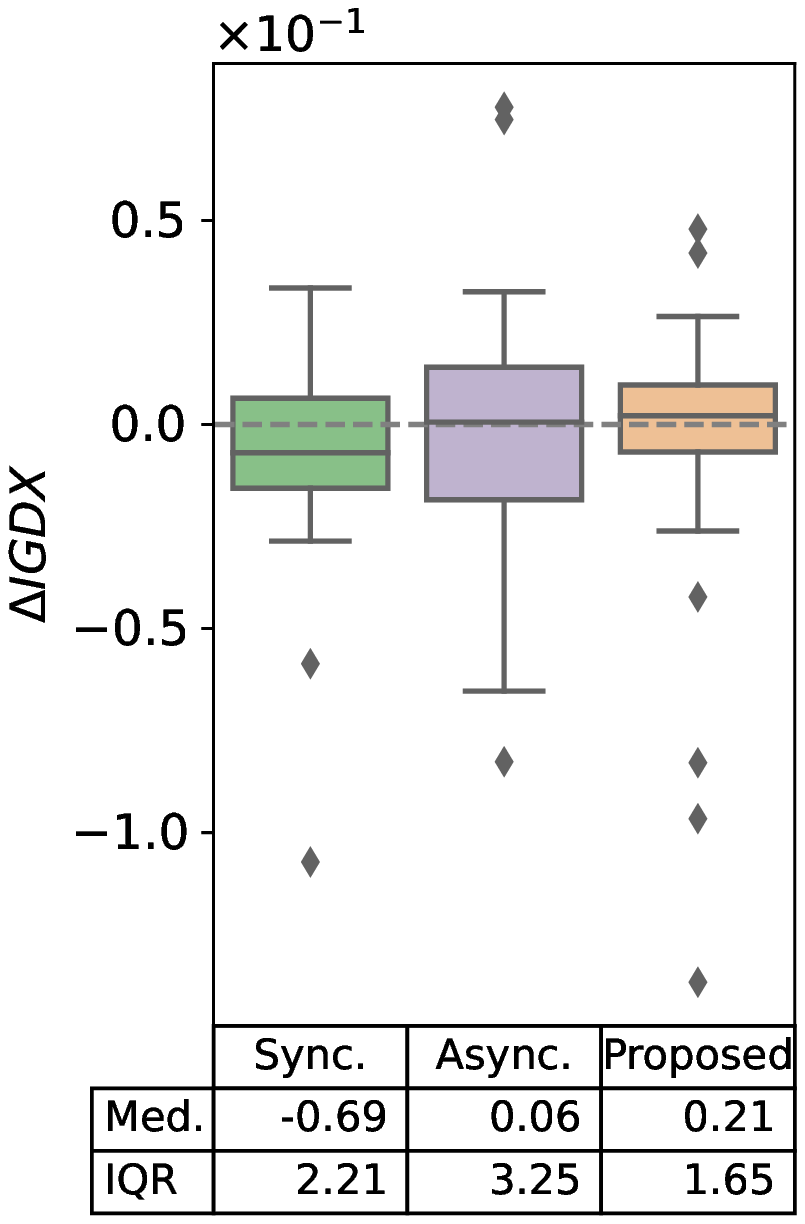}}
\end{minipage}
&
\begin{minipage}[b]{0.14\textwidth}
\subfloat[MMF6]{\includegraphics[scale=0.3]{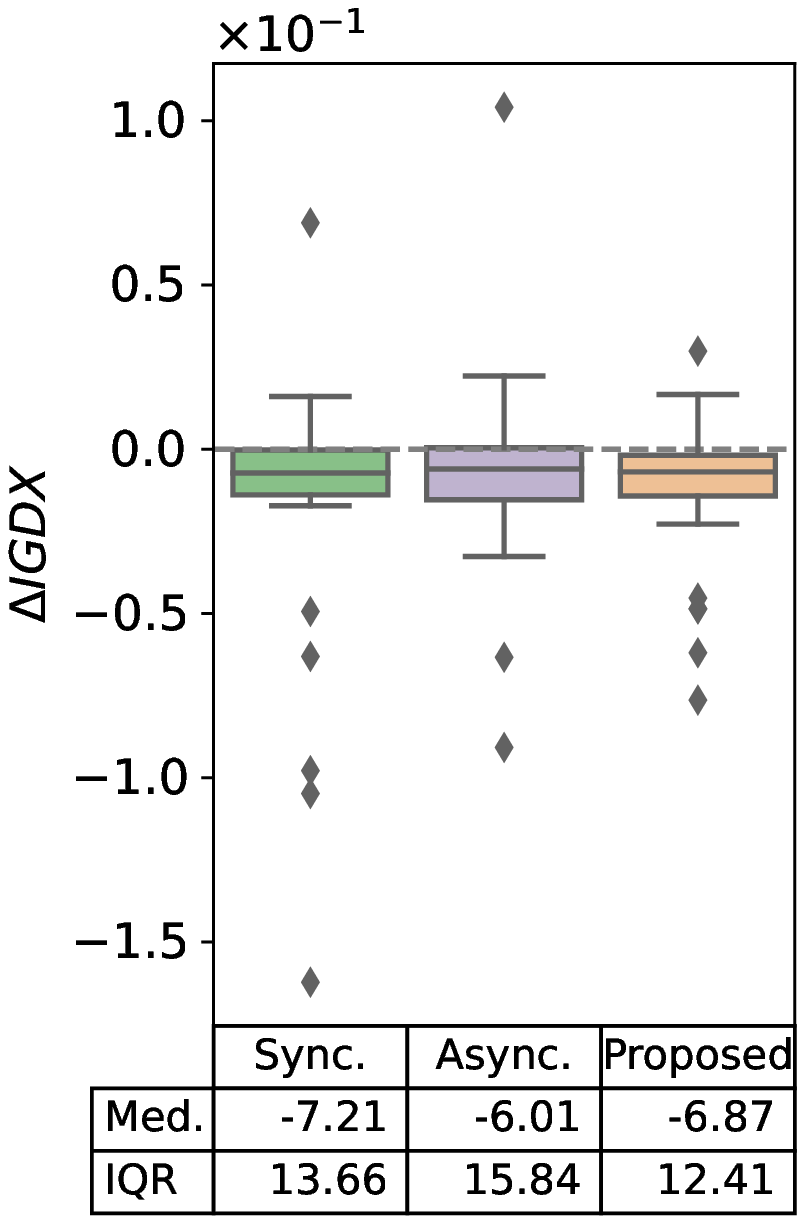}}
\end{minipage}
&
\begin{minipage}[b]{0.14\textwidth}
\subfloat[MMF8]{\includegraphics[scale=0.3]{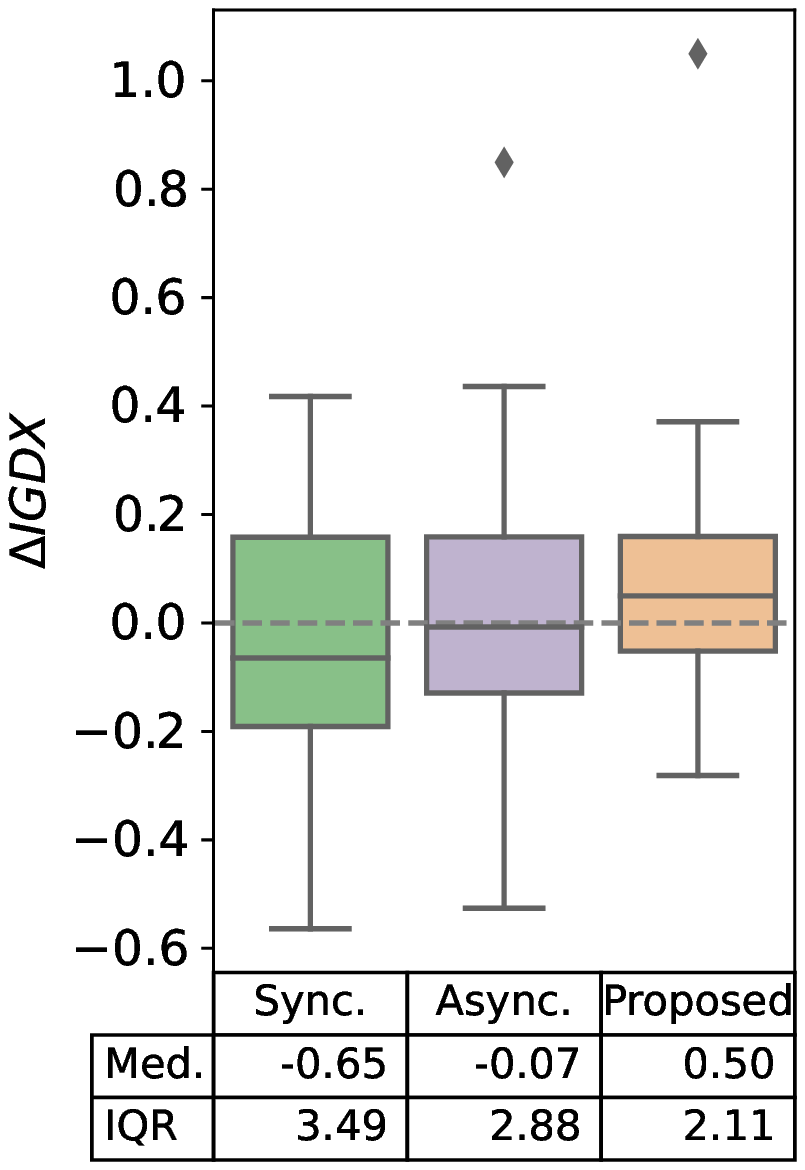}}
\end{minipage}
\end{tabular}
\caption{$\Delta IGDX$ with \textbf{No-bias} after the maximum fitness evaluations (different parallelization schemes)}
\label{fig:delta_igdx_nobias}
\begin{tabular}{cccccc}
\begin{minipage}[b]{0.14\textwidth}
\subfloat[MMF2]{\includegraphics[scale=0.3]{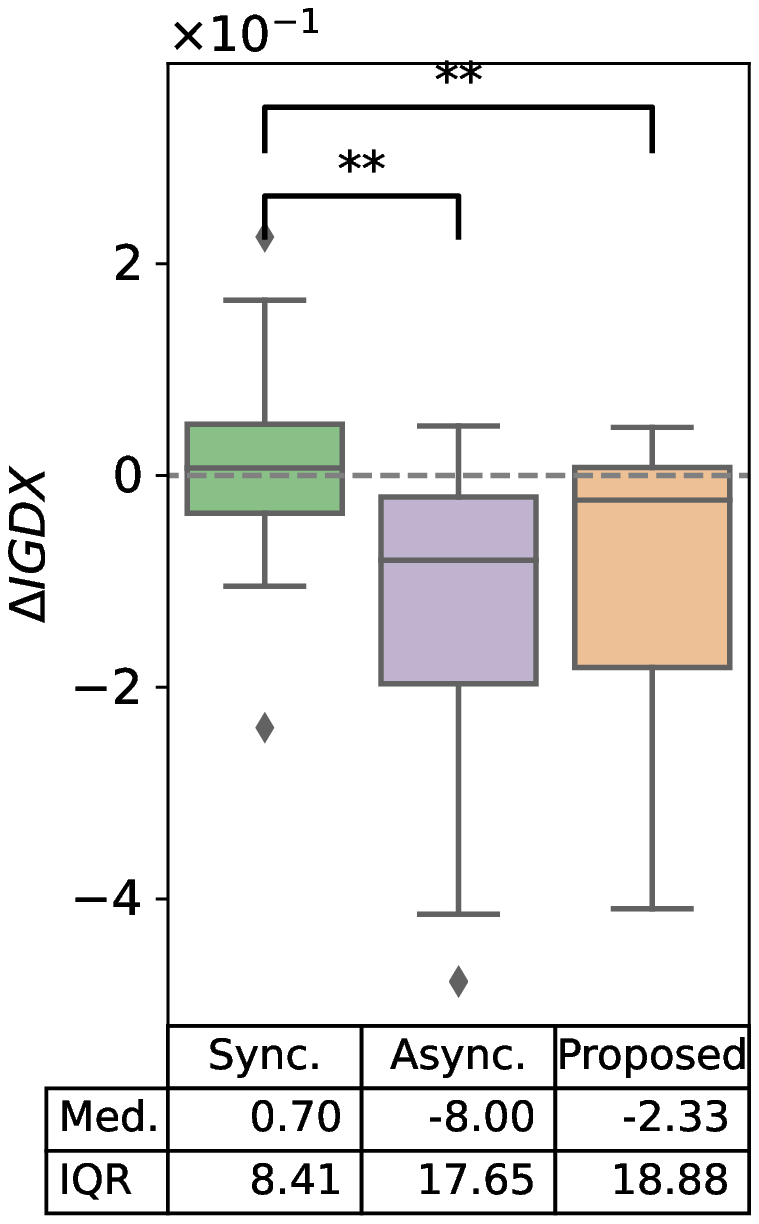}}
\end{minipage}
&
\begin{minipage}[b]{0.14\textwidth}
\subfloat[MMF3]{\includegraphics[scale=0.3]{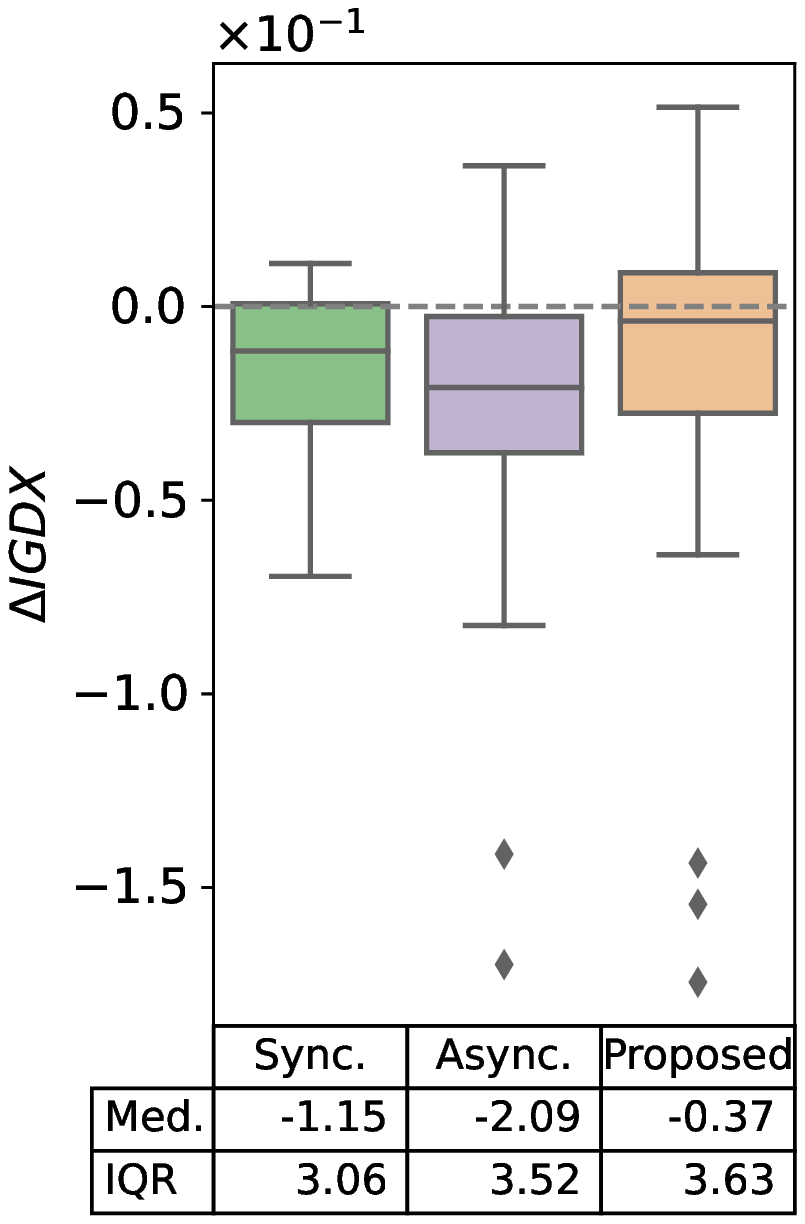}}
\end{minipage}
&
\begin{minipage}[b]{0.14\textwidth}
\subfloat[MMF4]{\includegraphics[scale=0.3]{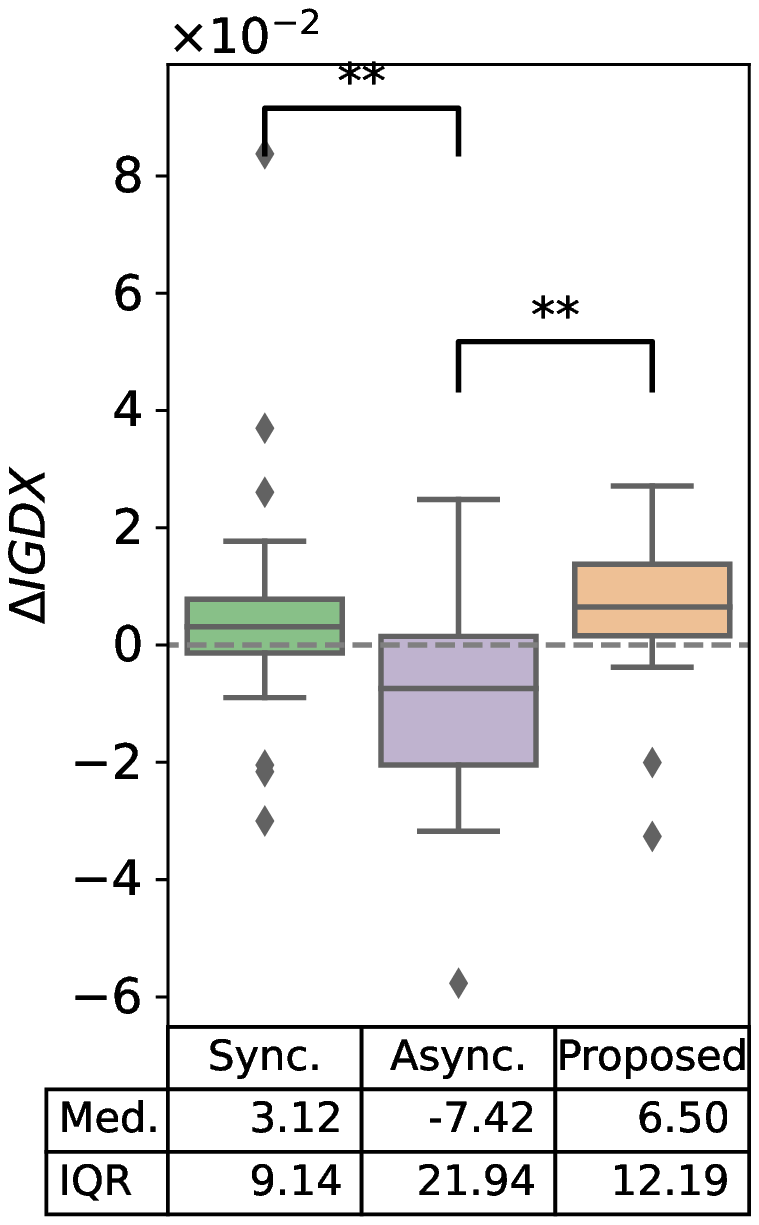}}
\end{minipage}
&
\begin{minipage}[b]{0.14\textwidth}
\subfloat[MMF5]{\includegraphics[scale=0.3]{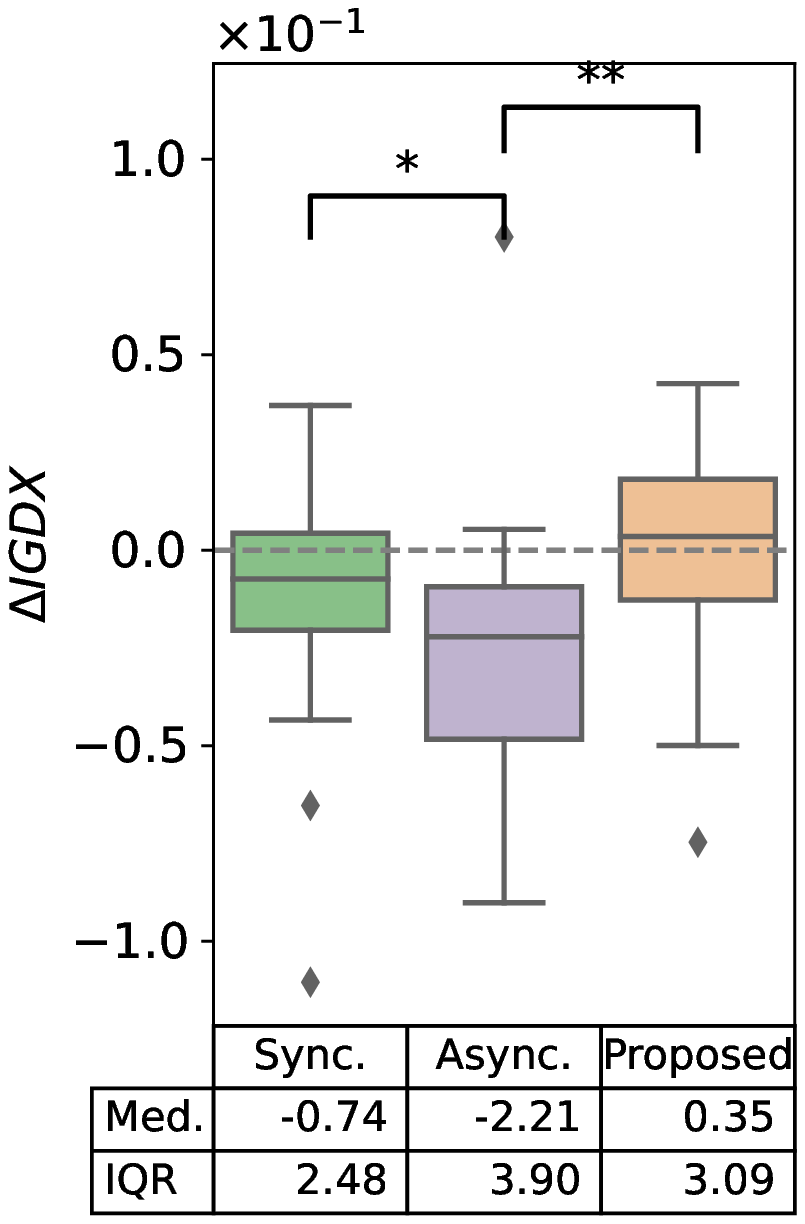}}
\end{minipage}
&
\begin{minipage}[b]{0.14\textwidth}
\subfloat[MMF6]{\includegraphics[scale=0.3]{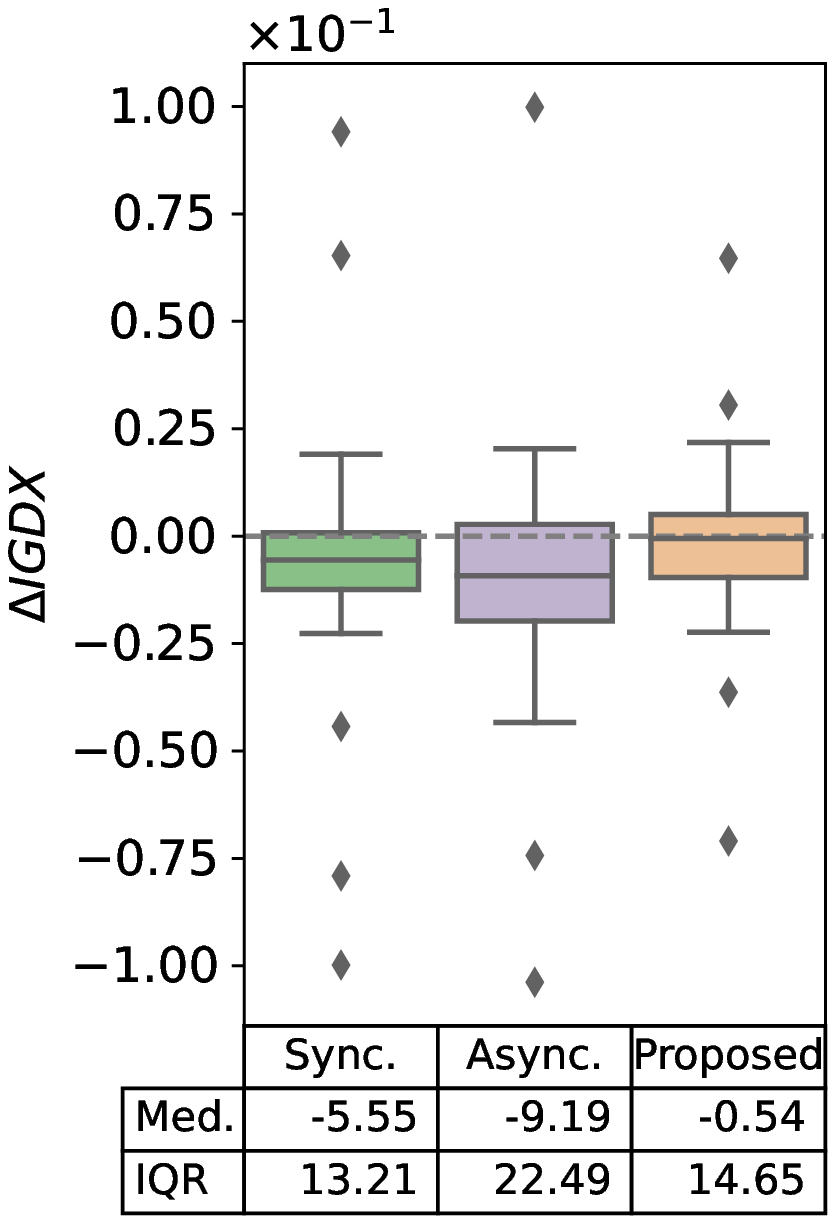}}
\end{minipage}
&
\begin{minipage}[b]{0.14\textwidth}
\subfloat[MMF8]{\includegraphics[scale=0.3]{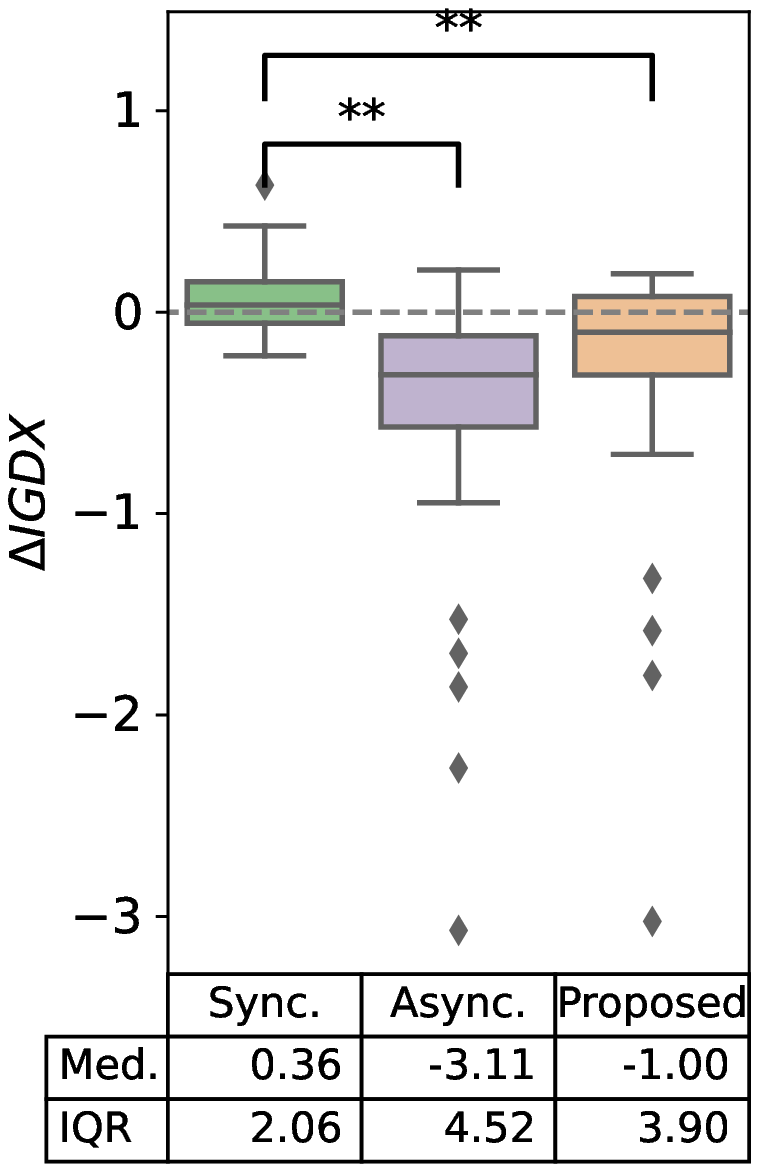}}
\end{minipage}
\end{tabular}
\caption{$\Delta IGDX$ with \textbf{Bias} after the maximum fitness evaluations (different parallelization schemes)}
\label{fig:delta_igdx_bias1}
\end{figure*}
Figures~\ref{fig:delta_igdx_nobias} and \ref{fig:delta_igdx_bias1} show the boxplot of the $\Delta IGDX$ value after the maximum number of fitness evaluations for \textbf{No-bias} and \textbf{Bias}, and the bottom tables summarize the median and IQR values.
The horizontal axis shows the different methods, while the vertical axis shows the $\Delta IGDX$ value.
As in the previous section, a significant difference is depicted with ``*'' symbols.

First, the \textbf{No-bias} results show no significant difference in the $\Delta IGDX$ value between the three parallelization methods. In addition, since the $\Delta IGDX$ value is almost zero in all problems, it is revealed that all parallelization schemes can obtain the separated Pareto sets equally if the evaluation time is not biased.

On the other hand, when using \textbf{Bias}, significant differences are found in MMF2, MMF4, MMF5, and MMF8, while no significant difference is found in MMF3 and MMF6. In particular, AP-NSGA-III obtains a significantly smaller (negative) $\Delta IGDX$ value than SP-NSGA-III in MMF2, MMF4, MMF5, and MMF8. This brings out the effect of the evaluation time bias in the asynchronous method.

From the results of the proposed method, a significant difference in MMF4 and MMF5 can be found. In these problems, the $\Delta IGDX$ value of FS-NSGA-III is not significantly different from that of SP-NSGA-III. In contrast, AP-NSGA-III is significantly biased toward PS1 (shorter evaluation time) than FS-NSGA-III and SP-NSGA-III.
On the other hand, in MMF2 and MMF8, FS-NSGA-III shows a significantly smaller (negative) $\Delta IGDX$ value than SP-NSGA-III. Moreover, no difference between FS-NSGA-III and AP-NSGA-III is found, though the distribution of the $\Delta IGDX$ value of FS-NSGA-III is close to zero compared with AP-NSGA-III. 

These results can be classified into three categories that are; (1) MMF3 and MMF6, where the $\Delta IGDX$ values are almost equal between the three methods ($\text{Proposed}\approx\text{Sync.}\approx\text{Async.}$); (2) MMF4 and MMF5, where the asynchronous method is significantly biased toward the region with a shorter evaluation time than the others ($\text{Proposed}\approx\text{Sync.}\gg\text{Async.}$); and (3) MMF2 and MMF8, where the proposed method is also biased ($\text{Sync.}\gg\text{Proposed}\approx\text{Async.}$). 

The difference between these categories can be explained from the perspective of the distribution of the Pareto set shown in Fig.~\ref{fig:PS}.
In MMF3 and MMF6, two Pareto sets are overlapped in the $x_2$ dimension, and a solution in one Pareto set is easily generated from a solution in another Pareto set. Thus, the evaluation time bias effect is small, and all methods are not biased.
On the other hand, two Pareto sets are completely separated in the $x_2$ dimension in the other benchmarks, but they are close in MMF4 and MMF5 compared with MMF2 and MMF8. When the regions of Pareto sets are separated and their evaluation times are biased, the asynchronous method results in a biased search toward regions with short evaluation times. On the other hand, the proposed method can reduce the effect of evaluation time bias even when optimal solutions exist in separate regions with the biased evaluation time.

These results indicate that the proposed method can reduce the effect of the evaluation time bias despite being asynchronous. In contrast, the asynchronous method without the proposed method easily converges to a Pareto set with a shorter evaluation time. 

\subsection{Computational efficiency}
\label{sec:exp1_efficiency}
This subsection analyzes the computational efficiency of three methods by comparing the simulation time until obtaining the target $IGD$ values defined in Section~\ref{sec:exp2_time}.

\begin{figure*}[!tb]
\begin{tabular}{cccccc}
\begin{minipage}[b]{0.14\textwidth}
\centering
\subfloat[MMF2]{\includegraphics[scale=0.3]{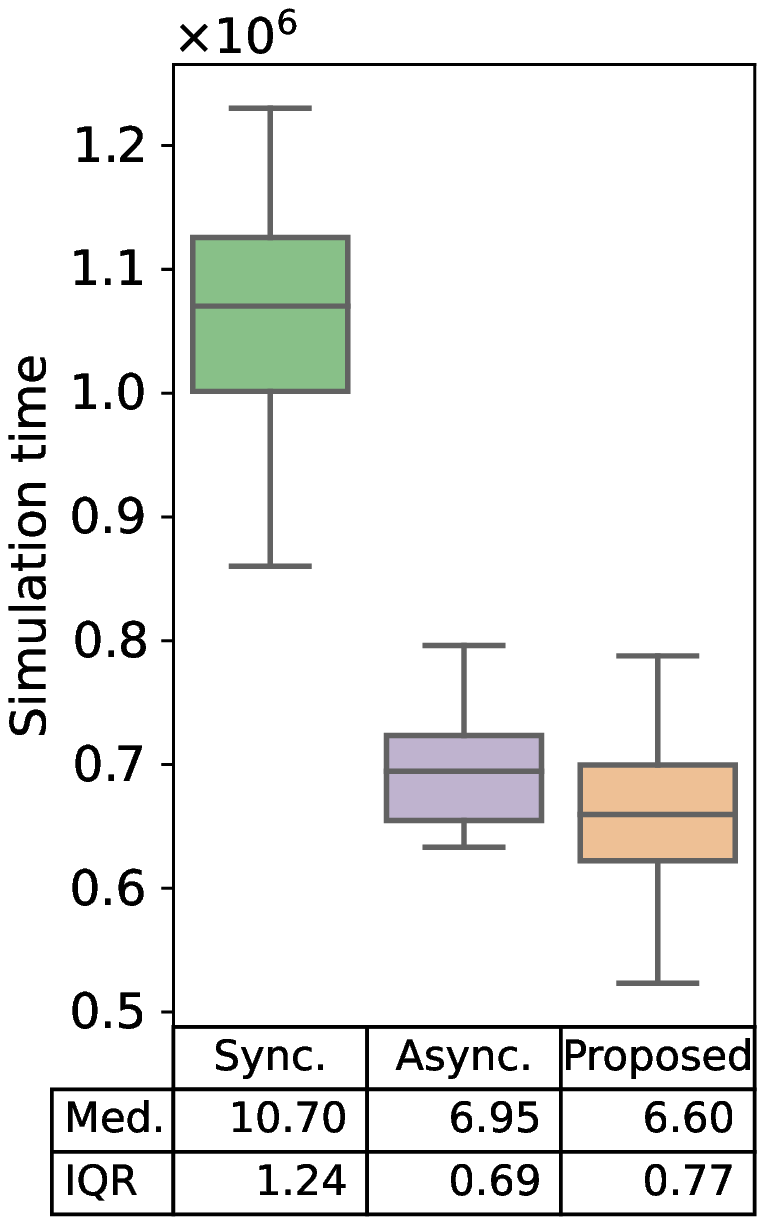}}
\end{minipage}
&
\begin{minipage}[b]{0.14\textwidth}
\centering
\subfloat[MMF3]{\includegraphics[scale=0.3]{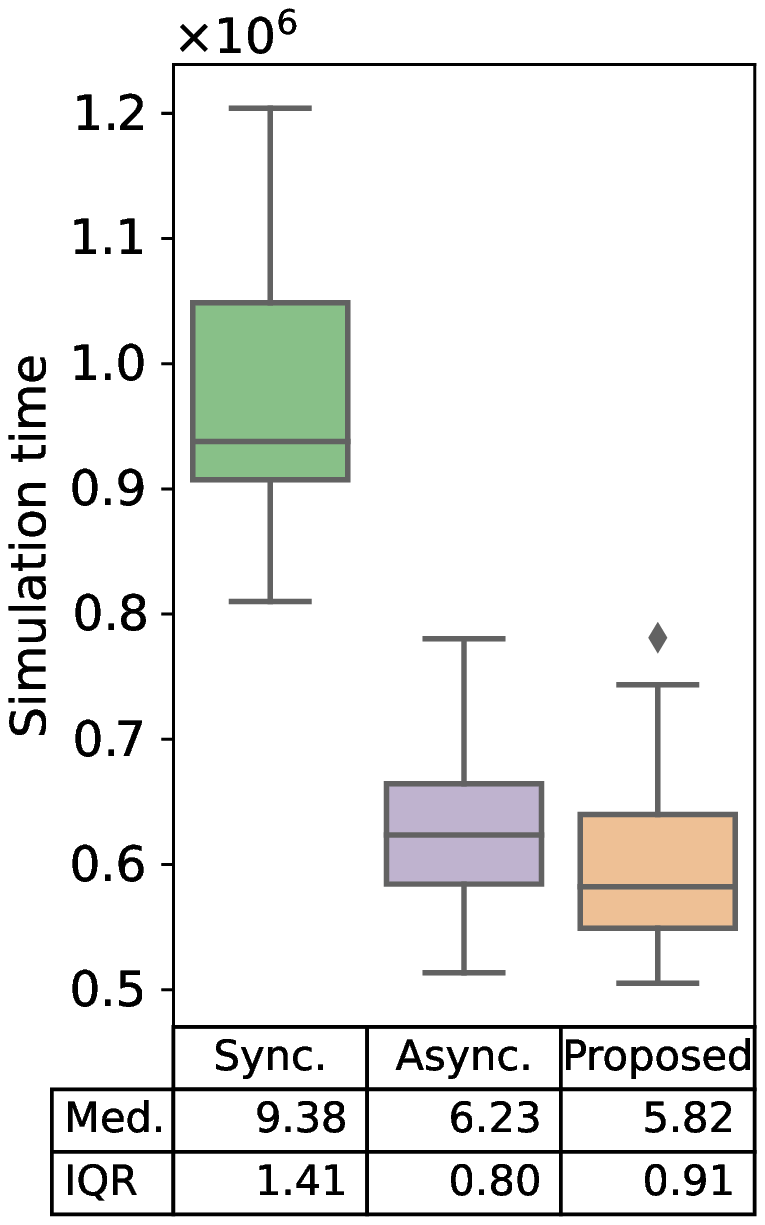}}
\end{minipage}
&
\begin{minipage}[b]{0.14\textwidth}
\centering
\subfloat[MMF4]{\includegraphics[scale=0.3]{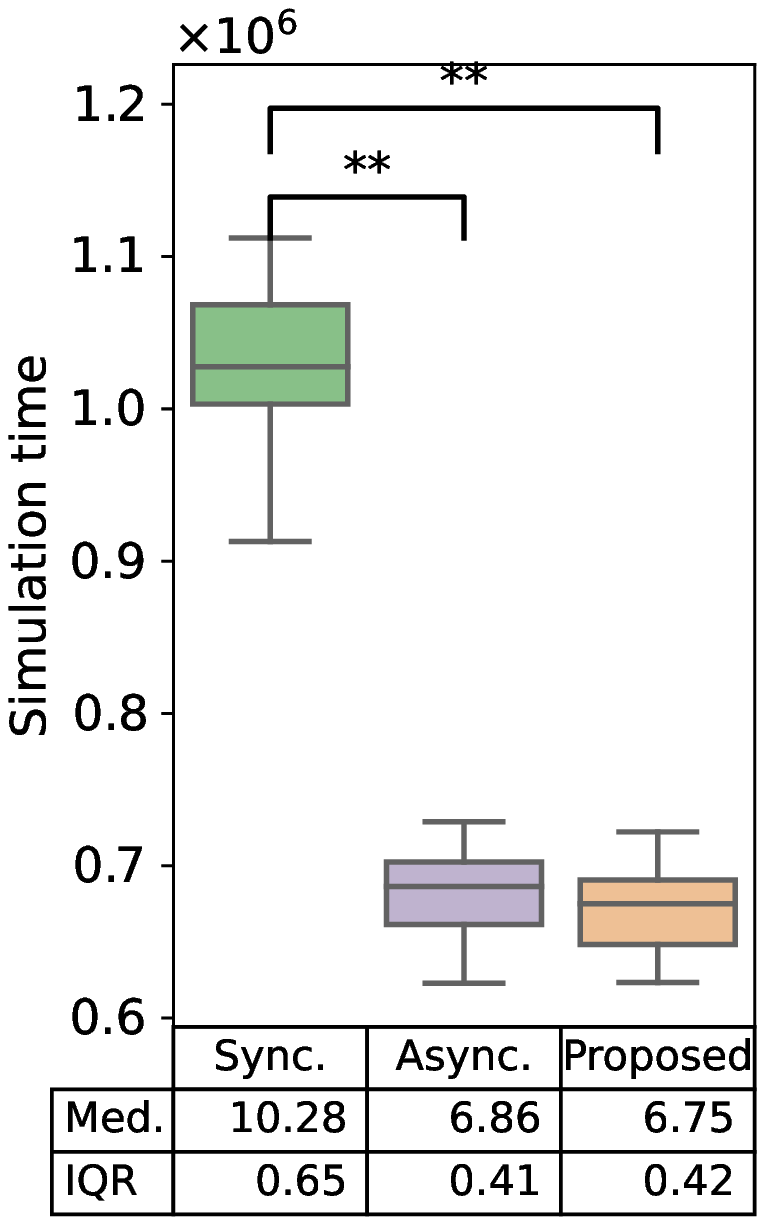}}
\end{minipage}
&
\begin{minipage}[b]{0.14\textwidth}
\centering
\subfloat[MMF5]{\includegraphics[scale=0.3]{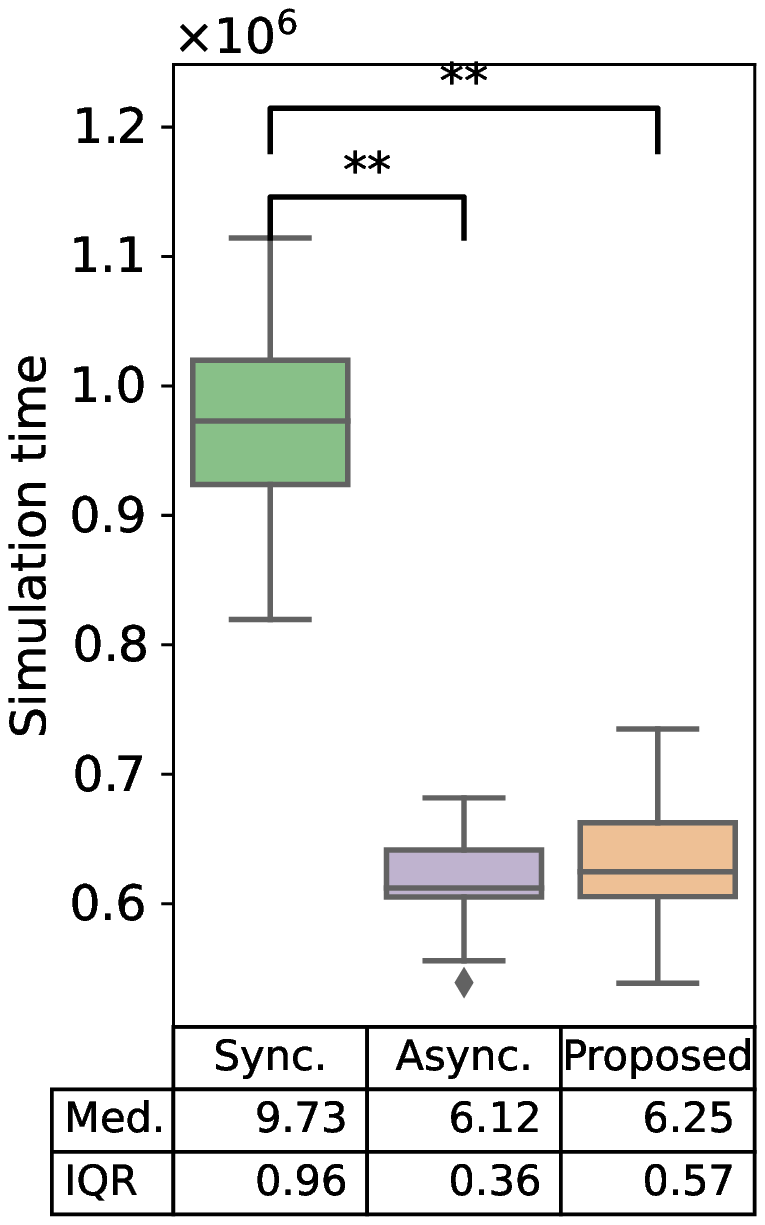}}
\end{minipage}
&
\begin{minipage}[b]{0.14\textwidth}
\centering
\subfloat[MMF6]{\includegraphics[scale=0.3]{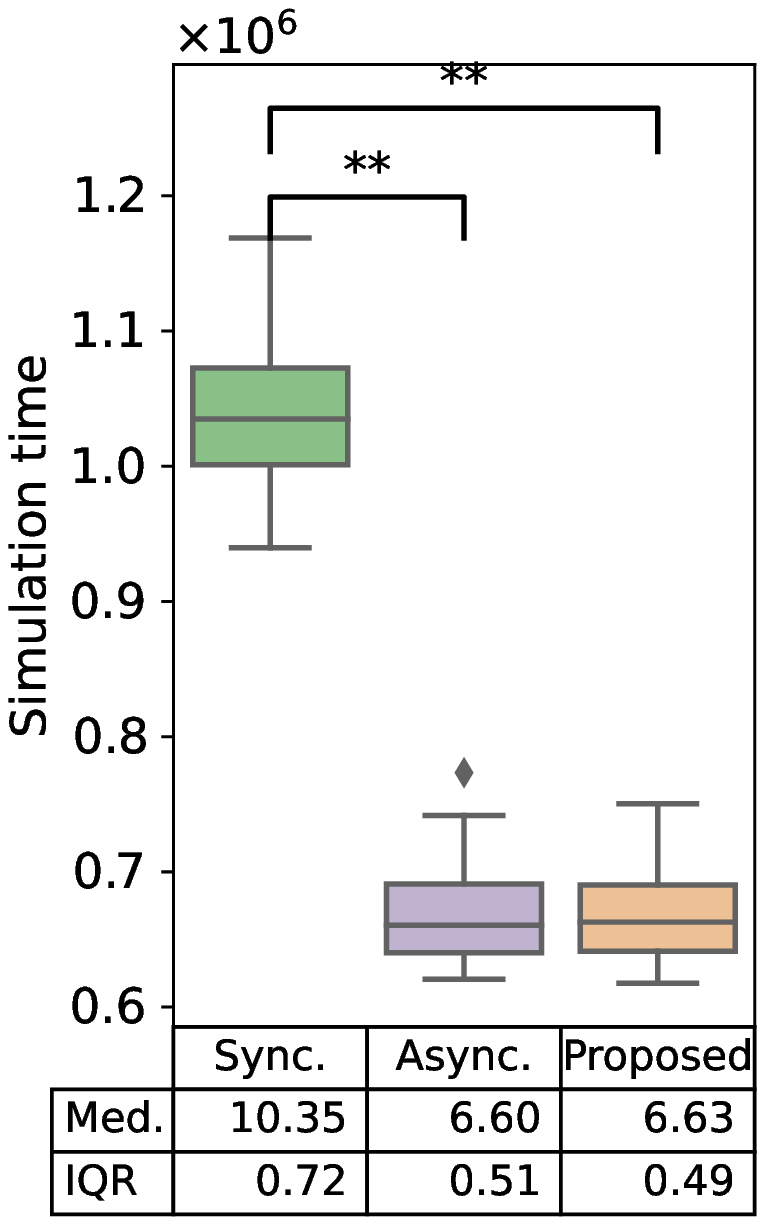}}
\end{minipage}
&
\begin{minipage}[b]{0.14\textwidth}
\centering
\subfloat[MMF8]{\includegraphics[scale=0.3]{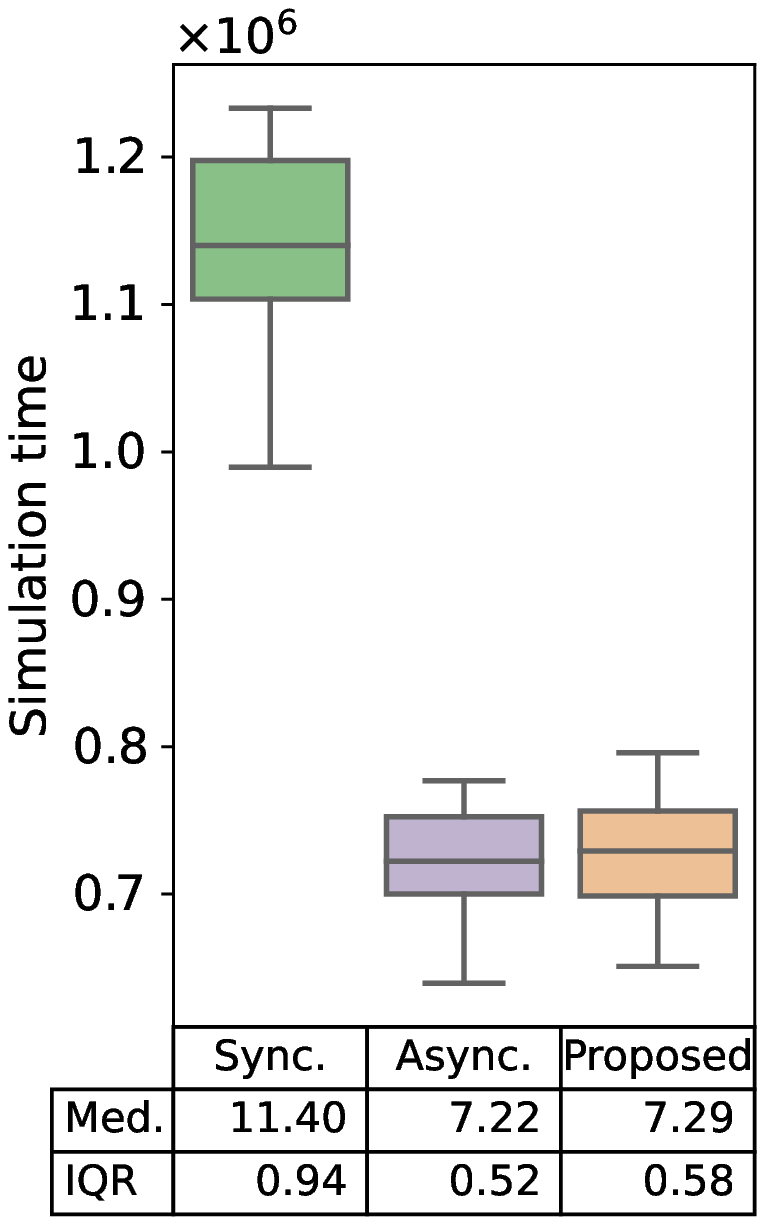}}
\end{minipage}
\end{tabular}
\caption{The simulation time until reaching a particular $IGD$ value with \textbf{No-bias} (different parallelization schemes)}
\label{fig:time_bp_nobias}
\begin{tabular}{cccccc}
\begin{minipage}[b]{0.14\textwidth}
\centering
\subfloat[MMF2]{\includegraphics[scale=0.3]{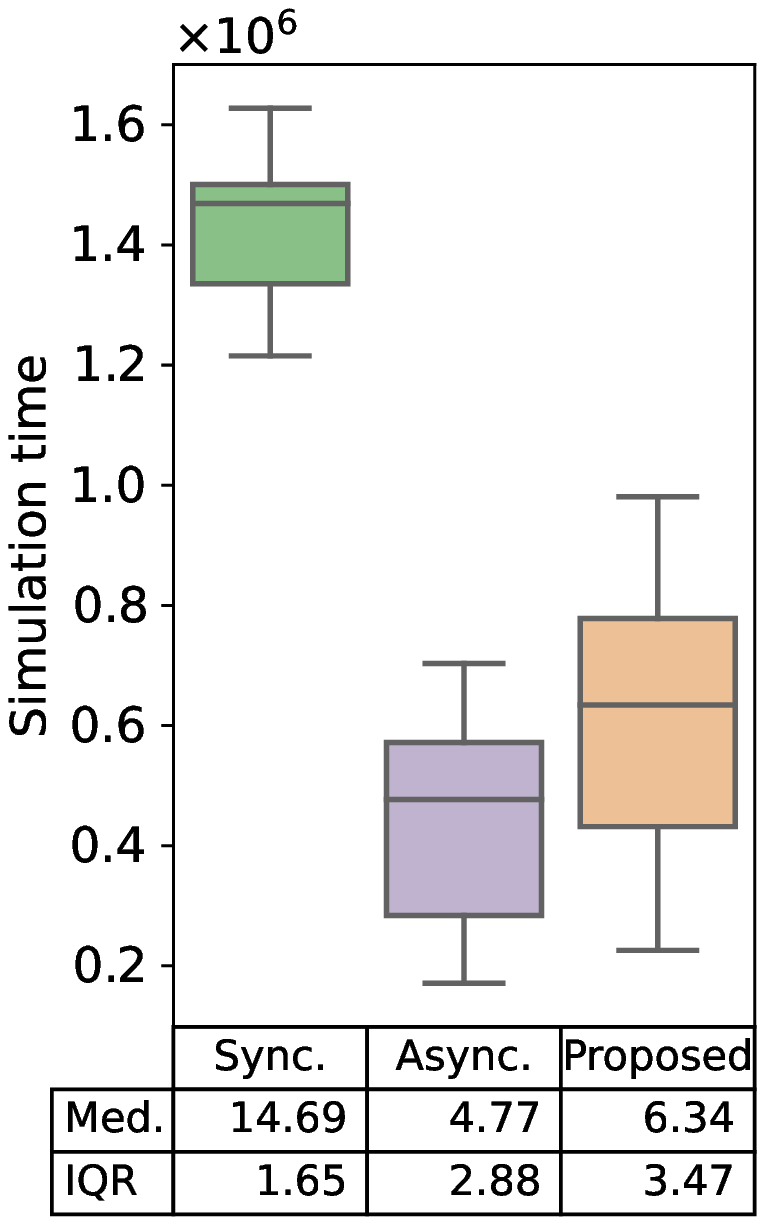}}
\end{minipage}
&
\begin{minipage}[b]{0.14\textwidth}
\centering
\subfloat[MMF3]{\includegraphics[scale=0.3]{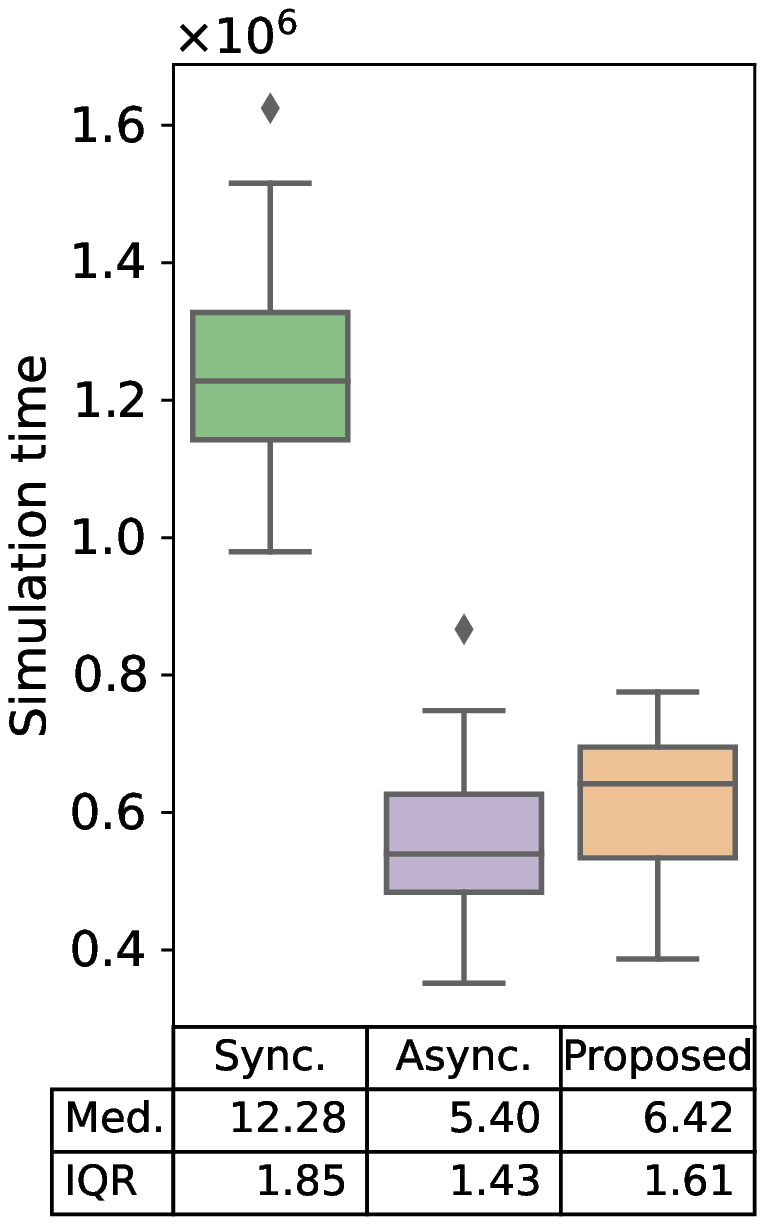}}
\end{minipage}
&
\begin{minipage}[b]{0.14\textwidth}
\centering
\subfloat[MMF4]{\includegraphics[scale=0.3]{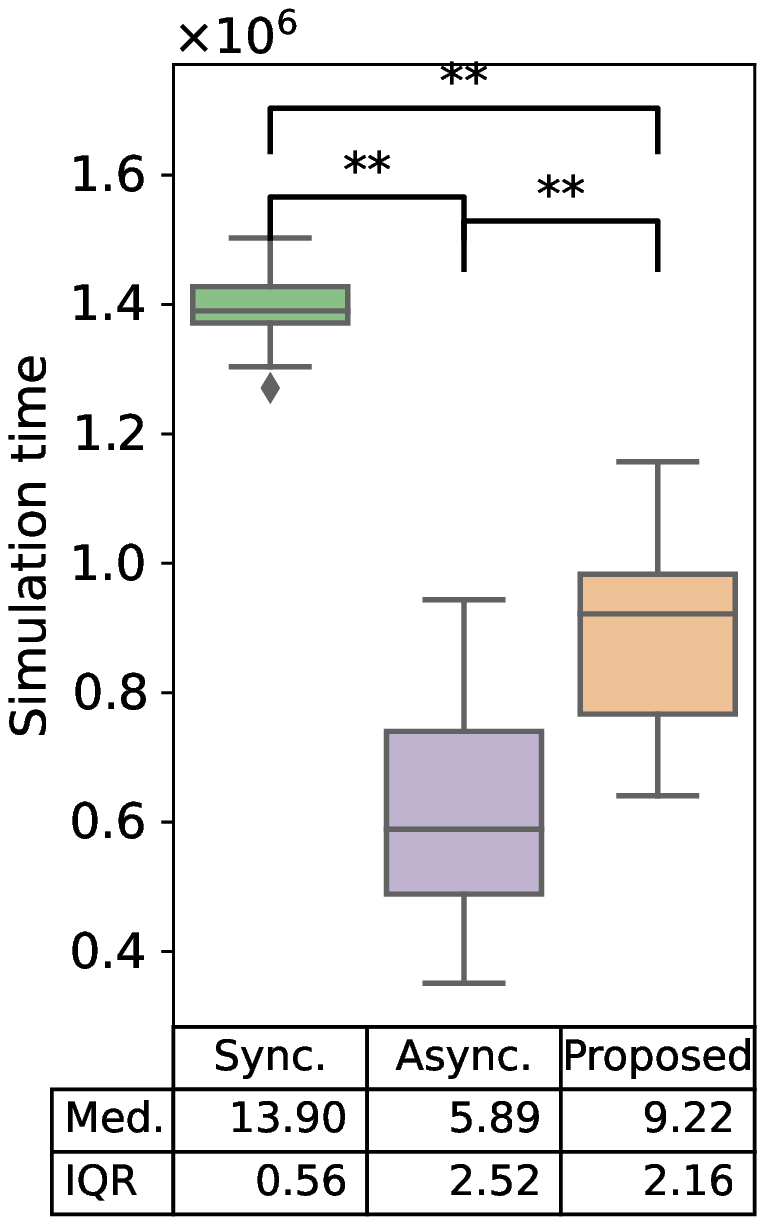}}
\end{minipage}
&
\begin{minipage}[b]{0.14\textwidth}
\centering
\subfloat[MMF5]{\includegraphics[scale=0.3]{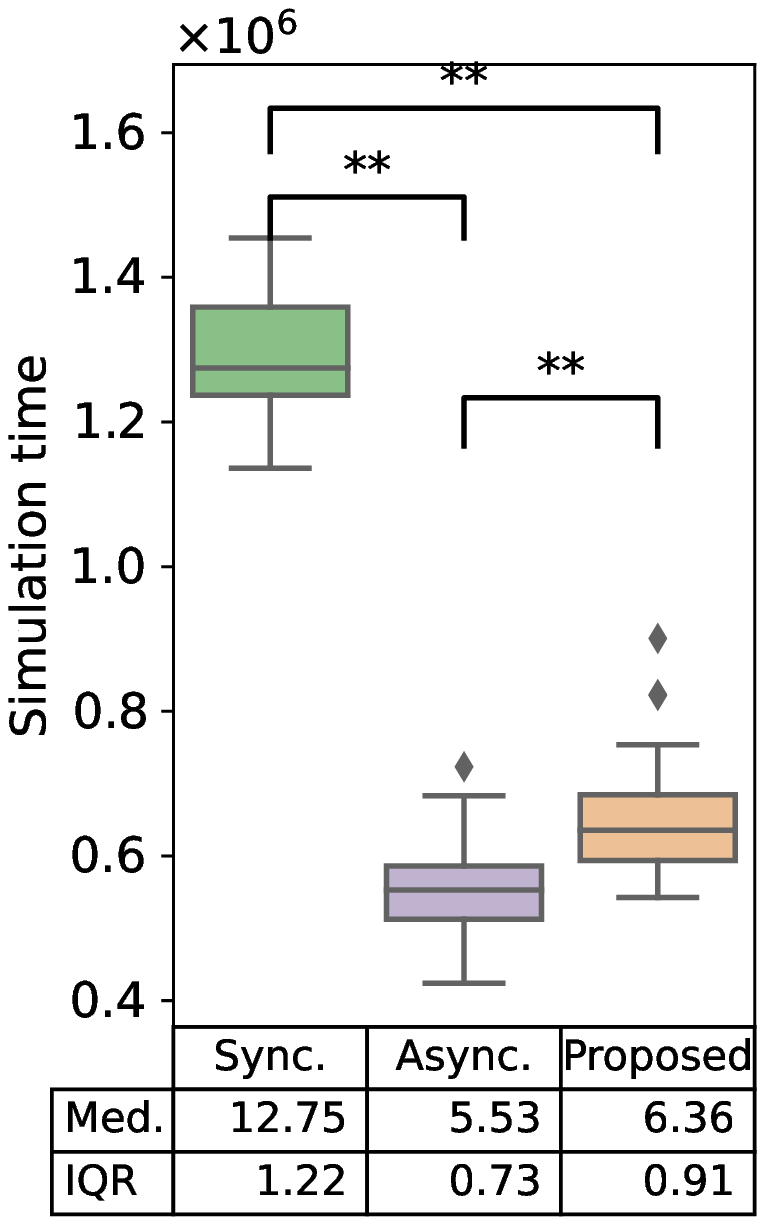}}
\end{minipage}
&
\begin{minipage}[b]{0.14\textwidth}
\centering
\subfloat[MMF6]{\includegraphics[scale=0.3]{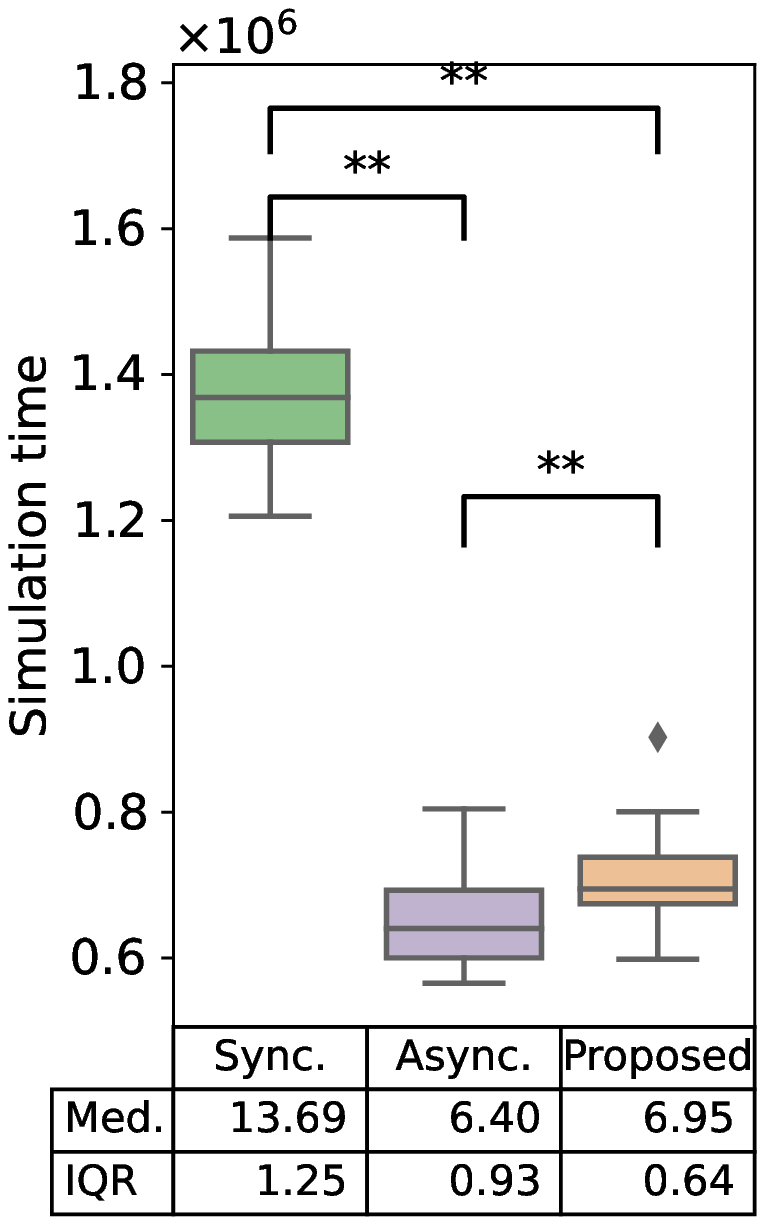}}
\end{minipage}
&
\begin{minipage}[b]{0.14\textwidth}
\centering
\subfloat[MMF8]{\includegraphics[scale=0.3]{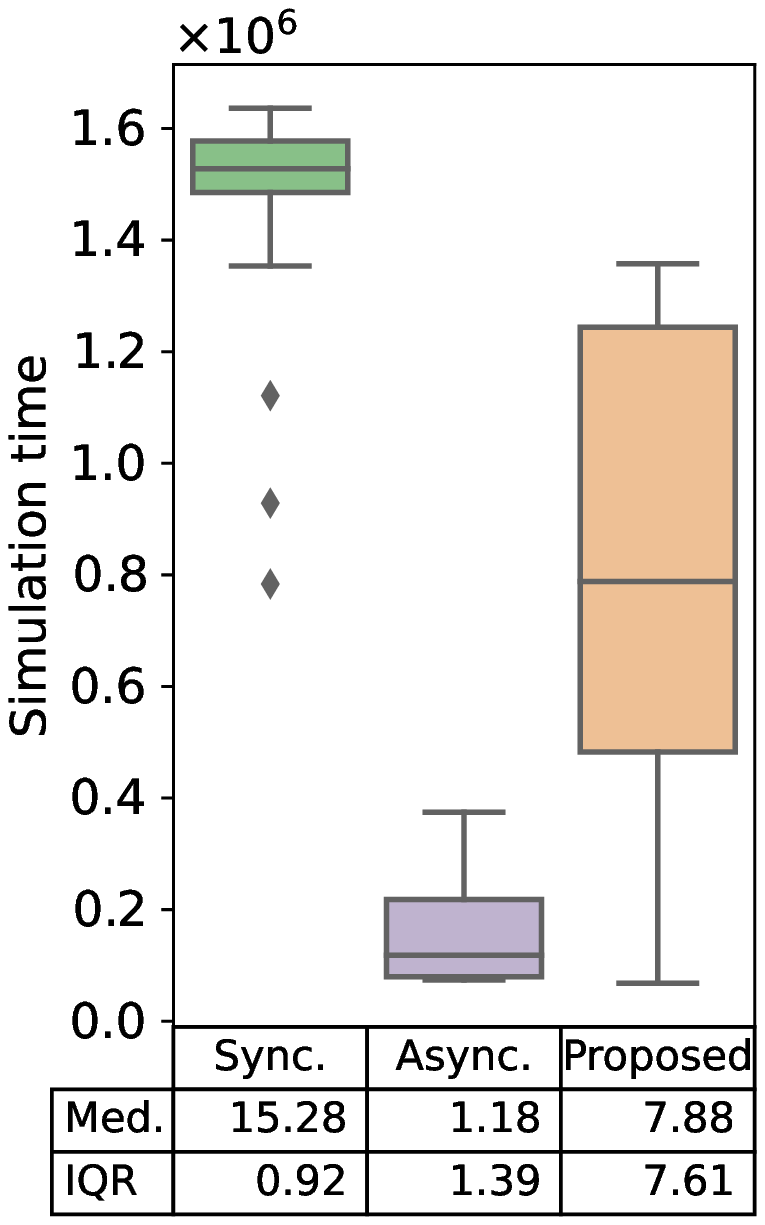}}
\end{minipage}
\end{tabular}
\caption{The simulation time until reaching a particular $IGD$ value with \textbf{Bias} (different parallelization schemes)}
\label{fig:time_bp_bias1}
\end{figure*}
Figures~\ref{fig:time_bp_nobias} and \ref{fig:time_bp_bias1} show the simulation time until obtaining the target $IGD$ value for \textbf{No-bias} and \textbf{Bias}, and the median and IQR values are summarized at the bottom of the figures. The horizontal axis shows the different parallelization methods, while the vertical axis shows the simulation time.
As in the previous section, a significant difference is depicted with ``*'' symbols.

From these results, FS-NSGA-III significantly reduces the execution time compared with the SP-NSGA-III when using \textbf{Bias}. Meanwhile, there is no significant difference between FS-NSGA-III and AP-NSGA-III. This indicates that the proposed method retains the computational efficiency of the asynchronous one.

On the other hand, from the results with the evaluation time of \textbf{Bias}, FS-NSGA-III also acquires a shorter execution time than SP-NSGA-III. In particular, the proposed method significantly reduces the execution time in MMF4, MMF5, and MMF6. Although no significant difference is found in MMF2, MMF3, and MMF8, the proposed method obtains enough better performance in half execution time than SP-NSGA-III. Since the result in Fig.~\ref{fig:delta_igdx_bias1} showed that FS-NSGA-III and SP-NSGA-III equally obtain two Pareto sets with different evaluation times, it can be said that the proposed method can reduce the execution time while reducing the effect of evaluation time bias.

The comparison of FS-NSGA-III and AP-NSGA-III indicates that the proposed method requires a significantly longer execution time when using \textbf{Bias}. However, this behavior can be explained because AP-NSGA-III is biased toward searching for solutions with shorter evaluation times (PS1), as indicated in Fig.~\ref{fig:delta_igdx_bias1}. On the other hand, since FS-NSGA-III obtains solutions with longer evaluation times, its execution time increases compared with AP-NSGA-III.

This result indicates that the proposed method obtains the computational efficiency of the asynchronous method while avoiding the effect of the evaluation time bias.

\section{Conclusion}\label{sec:conclusion}
This paper proposed a new parent selection method for reducing the effect of evaluation time bias in APEAs.
In particular, the proposed method considers the search frequency of each solution and selects parents from the pre-selected candidate pool.
This paper conducted experiments on multi-objective optimization test problems based on MMFs to analyze the effect of the evaluation time bias deeply. 
The proposed method was applied to the parallel NSGA-III and was compared with the synchronous and the asynchronous parallelizations.

The experiments first analyzed the impact of the selection ratio in the proposed method using the same test problems. This analysis indicated that the selection ratio in $0.5\le r_s\le 0.7$ acquires an appropriate balance between the search capability and the computational efficiency while reducing the effect of the evaluation time bias.
Then, the experimental results indicated that the proposed method could reduce the negative influence of the evaluation time bias.
The proposed method also does not adversely affect the search capability of APEAs while reducing the execution time significantly from SPEAs.
These results revealed that the proposed method possesses high search capability and high computational efficiency for problems with heterogeneous evaluation time.

It should be addressed to further analyze the proposed method on other benchmarks and with other EA methods shortly. In addition, although this paper only compared the proposed method with the synchronous and the asynchronous method, it should be compared with or integrated into a semi-asynchronous method~\citep{Harada2020} to adapt any characteristics of the evaluation time.


\section*{Declarations}
\bmhead{Funding}

This work was supported by Japan Society for the Promotion of Science Grant-in-Aid for Young Scientists Grant Number JP19K20362.

\bmhead{Conflict of interest}
The author declares that he has no conflict of interest.

\bibliography{bib_harada}


\end{document}